%% file: output.tex
\documentclass[10pt,twocolumn,letterpaper]{article}

\usepackage{cvpr}      

\input{preamble}

\usepackage{caption}
\usepackage{bm}
\usepackage{bbm}
\usepackage{dsfont}
\usepackage{graphicx}
\usepackage{tcolorbox}
\usepackage{multirow}

\input{macros}

\definecolor{cvprblue}{rgb}{0.21,0.49,0.74}
\usepackage[pagebackref,breaklinks,colorlinks,allcolors=cvprblue]{hyperref}

\def\paperID{*****} 
\def\confName{CVPR}
\def\confYear{2025}

\title{ConceptMix++: Leveling the Playing Field in Text-to-Image \\Benchmarking via Iterative Prompt Optimization} 
\author{Haosheng Gan, Berk Tinaz, Mohammad Shahab Sepehri, Zalan Fabian, Mahdi Soltanolkotabi\\
University of Southern California\\
Los Angeles, CA 90089\\
{\tt\small \{woodygan, tinaz, sepehri, zfabian, soltanol\}@usc.edu}
}

\begin{document}
\maketitle
\begin{abstract}
Current text-to-image (T2I) benchmarks evaluate models on rigid prompts, potentially underestimating true generative capabilities due to prompt sensitivity and creating biases that favor certain models while disadvantaging others. We introduce ConceptMix++, a framework that disentangles prompt phrasing from visual generation capabilities by applying iterative prompt optimization. Building on ConceptMix, our approach incorporates a multimodal optimization pipeline that leverages vision-language model feedback to refine prompts systematically. Through extensive experiments across multiple diffusion models, we show that optimized prompts significantly improve compositional generation performance, revealing previously hidden model capabilities and enabling fairer comparisons across T2I models. Our analysis reveals that certain visual concepts -- such as spatial relationships and shapes -- benefit more from optimization than others, suggesting that existing benchmarks systematically underestimate model performance in these categories. Additionally, we find strong cross-model transferability of optimized prompts, indicating shared preferences for effective prompt phrasing across models. These findings demonstrate that rigid benchmarking approaches may significantly underrepresent true model capabilities, while our framework provides more accurate assessment and insights for future development.
\vspace{-0.3cm}
\end{abstract}

\section{Introduction}
Text-to-image (T2I) generation aims to synthesize images based on user-specified textual descriptions. Diffusion models (DM) \citep{ho_denoising_2020, song_generative_2020} have established state-of-the-art in image \citep{dhariwal_diffusion_2021, nichol_improved_2021, saharia2022photorealistic, rombach2022high, ho2022cascaded}, audio \citep{kong2020diffwave}, and video generation \citep{ho2022video}. In the specific domain of T2I generation, DM-based approaches have become the dominant paradigm \cite{rombach2022high, saharia_image_2021}, achieving remarkable progress in both visual quality and semantic fidelity.

To enable systematic evaluation and comparison of DM-based T2I models, several benchmarks have been proposed \citep{wu2024conceptmix, huang2023t2i, bakr2023hrs}. These benchmarks typically employ fixed prompt formulations to assess model capabilities across various visual concepts and compositional complexities. However, evaluating and comparing these models remains challenging due to their inherent prompt sensitivity \citep{witteveen2022investigatingpromptengineeringdiffusion}. This sensitivity creates a fundamental disconnect between a model's true generative capabilities and what current benchmarks measure.
\begin{figure}[ht]
\centering
\includegraphics[width=0.9\linewidth]{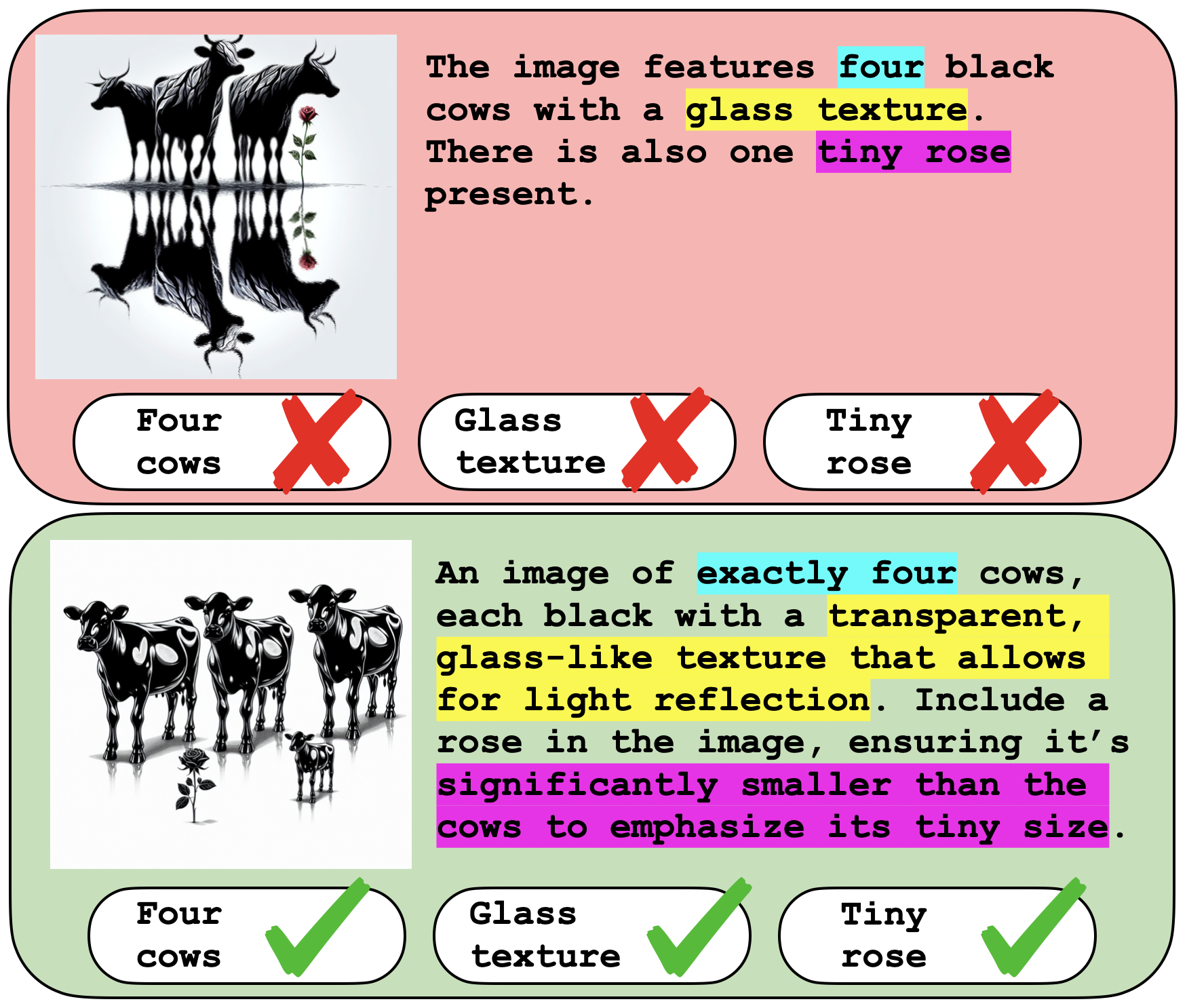}
\caption{Text-to-image models are sensitive to the specific phrasing of the input prompt, thus using a rigid prompt format may underestimate generation capabilities.
}
\label{fig:frontpage}
\end{figure} 

For example, as illustrated in Figure \ref{fig:frontpage}, two prompts expressing the same scene ("four cows", "glass texture", "tiny rose") yield drastically different results depending on how they are phrased. One prompt produces a successful generation (bottom), while the other fails entirely (top). This phenomenon suggests that existing evaluation methodologies may systematically underestimate model performance, creating biases that favor certain models while disadvantaging others based on their sensitivity to specific prompt formulations rather than their actual visual synthesis abilities.

To address this limitation, we propose ConceptMix++, a framework that disentangles prompt understanding from visual generation capabilities via systematic prompt optimization. Building on ConceptMix \citep{wu2024conceptmix}, our approach incorporates a multimodal optimization module that leverages feedback from vision-language models to iteratively refine prompts. This enables us to assess models under conditions that better reflect their maximum potential rather than their sensitivity to arbitrary prompt formulations.

Through extensive experiments with multiple state-of-the-art diffusion models, we demonstrate that:
\begin{itemize}
    \item Optimized prompts substantially improve compositional generation performance across architectures, revealing hidden capabilities that standard benchmarks miss.
    \item Visual concept categories (e.g., spatial relations, shapes) benefit unevenly from optimization, highlighting category-specific bottlenecks and suggesting areas for targeted architectural improvements.
    \item Optimized prompts exhibit remarkable cross-model transferability, suggesting shared preferences in effective prompting and common underlying representations across different architectures.
\end{itemize}

ConceptMix++ offers a more nuanced lens for evaluating T2I models, surfacing capabilities that conventional benchmarks may overlook while simultaneously exposing fundamental limitations in current evaluation methodologies. By providing both a fairer comparison framework and insights into model-specific strengths and weaknesses, our approach contributes to a more accurate understanding of the current state and future directions of text-to-image generation research.

\section{Related Work}

\textbf{Diffusion Models for Text-to-Image Generation.} Denoising diffusion probabilistic models \cite{ho_denoising_2020, song_generative_2020} have revolutionized text-to-image synthesis. Early works like DALL·E \cite{ramesh2021zero} and GLIDE \cite{nichol2021glide} demonstrated large-scale generative potential, while latent diffusion models \cite{rombach2022high} improved efficiency by operating in compressed latent spaces. Recent advances include transformer-based architectures \cite{peebles2023scalable} and specialized models like DALL·E 3 \cite{dalle3}.

\noindent\textbf{Text-to-Image Benchmarking.} Comprehensive benchmarks assess both visual quality and semantic alignment in T2I models. While early benchmarks like MS-COCO \cite{lin2014microsoft} focused on basic object recognition, recent specialized frameworks evaluate specific capabilities: DrawBench \cite{saharia2022photorealistic} uses human evaluation for compositional generation, T2I-CompBench \cite{huang2023t2i} measures compositional abilities across multiple dimensions, and ConceptMix \cite{wu2024conceptmix} targets compositional generation with varying complexity levels. However, current benchmarks use fixed prompt formulations, potentially underestimating model capabilities due to inherent prompt sensitivity.

\noindent\textbf{Vision-Language Models.} Vision-language models have emerged as powerful multimodal understanding tools. CLIP \cite{radford2021learning} marked a breakthrough with zero-shot capabilities through contrastive learning, followed by developments like BLIP \cite{li2022blip} and large-scale models like GPT-4V \cite{openai2023gpt4v}. In text-to-image evaluation, VLMs serve as automated evaluators for assessing semantic alignment between generated images and descriptions \cite{hessel2021clipscore}, making them essential for scalable benchmarking.

\noindent\textbf{Prompt Optimization and Engineering.} Prompt design significantly impacts language model effectiveness, spurring research in optimization techniques. While manual engineering \cite{wei2022chain} and automated methods like AutoPrompt \cite{shin2020autoprompt} have shown improvements, TextGrad \cite{yuksekgonul2024textgrad} introduced gradient-like feedback optimization. In the text-to-image domain, prompt engineering has been explored through manual techniques \cite{witteveen2022investigatingpromptengineeringdiffusion} and style-specific optimization methods. However, systematic automated prompt optimization using feedback-based approaches like TextGrad remains underexplored for T2I generation, which we address in this work.
\section{Method}
\subsection{Overview of our framework} ConceptMix++ is a framework designed to disentangle prompt phrasing from visual synthesis abilities of T2I models. Our key insight is that by adapting and optimizing prompts for each model individually, we can more accurately assess their true visual generation potential.

Our framework operates in three-stages:
\begin{enumerate}
    \item \underline{Baseline prompt evaluation:} Evaluate model performance using standard benchmark prompts.
    \item \underline{Prompt optimization:} Use our prompt optimization module to iteratively refine prompts tailored to each model.
    \item \underline{Capability analysis:} Assess each model's full visual synthesis ability under optimized prompting.
\end{enumerate}

This methodology allows for a more nuanced analysis of model capabilities, revealing: (1) performance gains (2) persistent limitations that remain even with optimized prompts.


\subsection{ConceptMix Baseline} For baseline evaluation, we adopt the ConceptMix \cite{wu2024conceptmix} benchmark, which measures compositional generation capabilities across a broad range of visual concept categories. This benchmark is particularly well-suited for capability analysis because:

\begin{itemize}
    \item It provides fine-grained evaluations across 8 diverse visual concept categories.
    \item It supports scalable compositional complexity denoted with complexity level $k$ ($k$ ranges between $1$ and $7$).
\end{itemize}

At each complexity level $k$, ConceptMix specifies $k+1$ criteria, each corresponding to one of the 8 visual concept categories. The initial prompt $p_0$ is generated by GPT-4o by providing it with full list of criteria. The diffusion model then synthesizes an image $\mathcal{I}$ from this prompt. To evaluate the generated image, ConceptMix utilizes GPT-4o acting as a verifier $\mathcal{V}$ which answers a yes/no question for each criterion to determine whether it has been satisfied. The image receives a score of 1 only if all criteria are met:
\begin{equation*}
s(\mathcal{I}) = \prod_{i=1}^{k+1} \1(\mathcal{V}(\text{``Is criteria $i$ satisfied ?"}|\mathcal{I}) = \texttt{"Yes"}) 
\end{equation*}
where $\1$ is the indicator function that equals 1 if the verifier thinks the criteria is satisfied and 0 otherwise.

\subsection{Multimodal prompt optimization}  
\begin{figure*}[ht]
\centering
\includegraphics[width=0.9\linewidth]{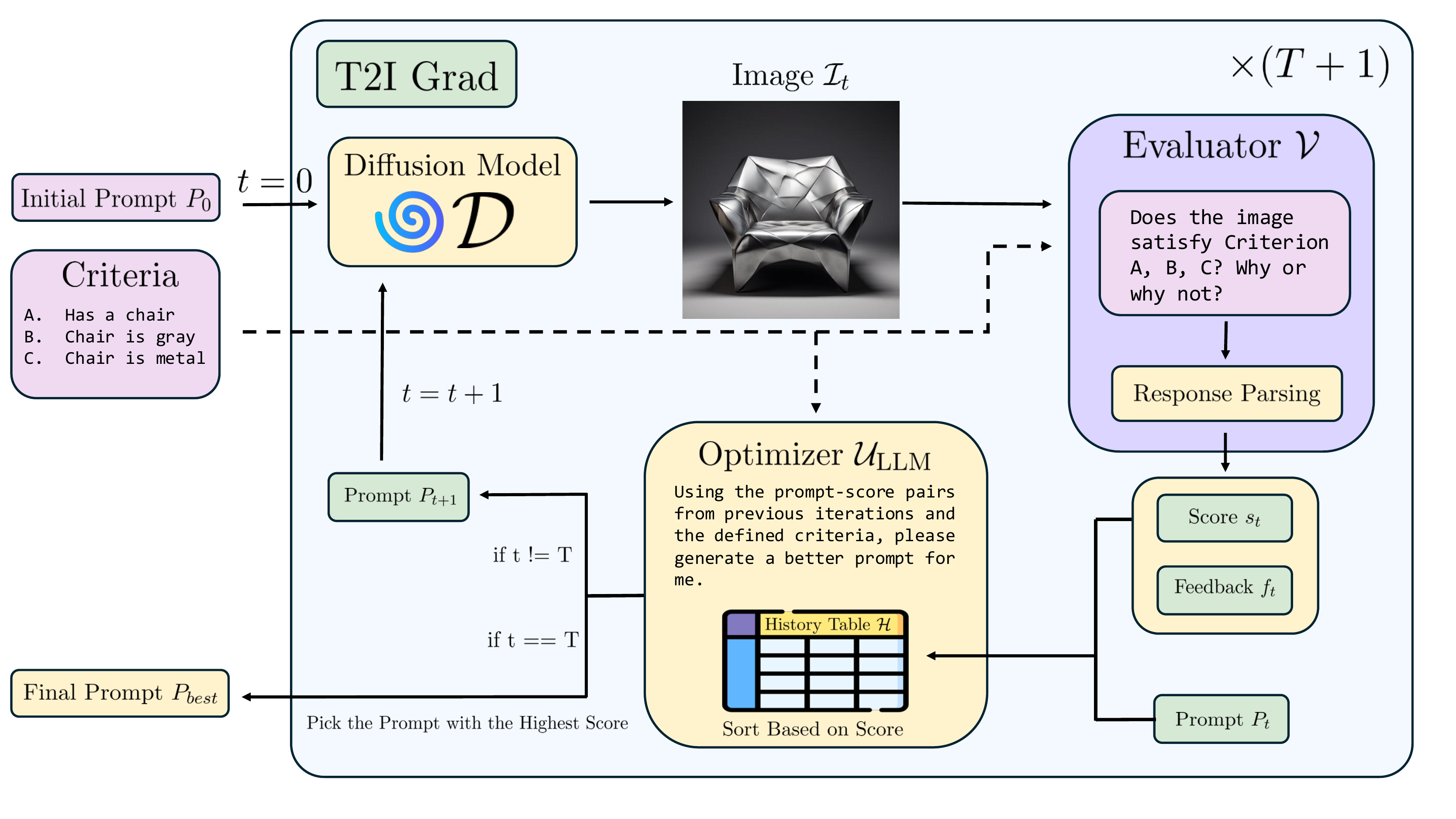}
\caption{Overview of our prompt optimization framework. Starting with an initial prompt, our iterative process first generates images using a diffusion model. Next, an evaluator scores and provides feedback. Finally, LLM optimizer proposes improved prompts based on past prompt-score pairs. 
}
\label{fig:t2i_grad}
\end{figure*}

To systematically enhance prompts, we introduce a prompt optimization module inspired by TextGrad \cite{yuksekgonul2024textgrad}, tailored specifically for the text-to-image (T2I) domain. As illustrated in Figure \ref{fig:t2i_grad}, our module operates end-to-end across the T2I generation and evaluation pipeline, refining prompts based on visual outcomes.

Given a T2I diffusion model $\mathcal{D}$ and a vision-language evaluator $\mathcal{V}$, the goal is to find an optimal prompt $p^*$ that maximizes the evaluation score:
\begin{equation*}
p^* = \arg\max_p \mathcal{V}(\mathcal{D}(p)) 
\end{equation*}
Here, $\mathcal{D}(p)$ denotes the image generated by the diffusion model from prompt $p$, and $\mathcal{V}(\mathcal{I})$ returns a scalar score assessing how well image $\mathcal{I}$ satisfies the specified visual concepts.

The optimization process (Figure \ref{fig:t2i_grad}) follows an iterative loop:
\begin{enumerate}
    \item Generate an image $\mathcal{I}_t = \mathcal{D}(p_t)$ from the current prompt.
    \item Evaluate $\mathcal{I}_t$ using the VLM to obtain a score $s_t = \mathcal{V}(\mathcal{I}_t)$ and feedback $f_t$.
    \item Store the prompt, score, and feedback tuple $(p_t, s_t, f_t)$ in a history buffer $H$, sorted by score.
    \item Update the prompt $p_{t+1} = \mathcal{U}_{\text{LLM}}(p_{best}, H)$, where $\mathcal{U}_{\text{LLM}}(\cdot)$ is an LLM-based update function.
\end{enumerate}

Unlike gradient-based optimization in continuous spaces, our method exploits the LLM's capability to learn from qualitative feedback and historical patterns, generating refined prompts in natural language.

To enable more fine-grained and stable updates, we extend ConceptMix's binary evaluation to a probabilistic one. Specifically, we use the likelihood of the affirmative answer ("Yes") predicted by the vision-language model for each criterion:
\begin{equation*}
\mathbbm{E}[s(\mathcal{I})] = \prod_{i=1}^{k+1} P(\mathcal{V}(\text{``Is criteria $i$ satisfied ?"}|\mathcal{I}) = \texttt{"Yes"}) 
\end{equation*}

We provide further implementation details and analysis of the optimization loop in Appendix~\ref{apx:prompt_opt}.
\section{Experiments}

\subsection{Setup}

We apply our ConceptMix++ framework to benchmark three state-of-the-art diffusion models: DALL·E 3 \cite{dalle3}, stable-diffusion-3.5-medium \cite{stablediffusion3.5}, and playground-v2.5-1024px-aesthetic \cite{playgroundv2.5}. In this setup, we use GPT-4o (2024-08-06) \cite{hurst2024gpt} as both the evaluator and the optimizer.

For each complexity level $k = 1$ to $k = 7$, we select 300 datapoints from the ConceptMix benchmark. To reduce the effect of randomness, we generate 5 images per prompt and evaluate performance based on two metrics. The \textit{average} score is the average across the 5 generations, while \textit{best-of-5} score is the highest score among them.

\subsection{Performance Analysis}

\label{sec:performance_gap}

\input{Table/main_results_table}

Table \ref{tab:main_results} presents the performance comparison between original and optimized prompts across all diffusion models and complexity levels. We observe substantial improvements following optimization -- up to $\approx 20\%$ absolute gains in both \textit{average} and \textit{best-of-5} scores for mid-range complexity levels ($k=3$ to $k=5$) across all models. These gains reflect the significant value of prompt optimization in unlocking true capabilities, particularly for compositions that require understanding and generating multiple visual concepts simultaneously. 

Interestingly, the performance gap between original and optimized prompts widens with increasing $k$ in the low-to-mid range, suggesting that prompt refinement becomes increasingly beneficial as compositional complexity grows. However, at the highest complexity levels ($k=6$, $k=7$), the gains taper off. This diminishing return likely stems from an upper bound imposed by the model's capacity itself: the prompts may become too detailed or domain-shifted compared to the model's training distribution, limiting the effectiveness of optimization.

While the numerical results are summarized in Table \ref{tab:main_results}, we present additional model-wise performance visualizations in Appendix~\ref{sec:model_charts} (Figure~\ref{fig:model_comparison}) that show both the \textit{mean} score and \textit{best-of-5} score metrics for each model before and after prompt optimization across varying complexity levels ($k=1$ to $k=7$). These plots clearly illustrate the performance gaps and reveal model-specific improvement patterns discussed above.

These results highlight the importance of more flexible evaluation frameworks in benchmarking text-to-image models. The substantial performance differences between standard and optimized prompts across all models demonstrate that conventional rigid benchmarking approaches may significantly underrepresent actual model capabilities, particularly for certain architectures like SD 3.5. Conversely, the persistent decline at higher complexity levels reveals fundamental limitations that require architectural innovations beyond prompt engineering.
\begin{figure}[ht]
    \centering
    \begin{subfigure}[t]{\linewidth}
        \centering
        \includegraphics[width=\linewidth]{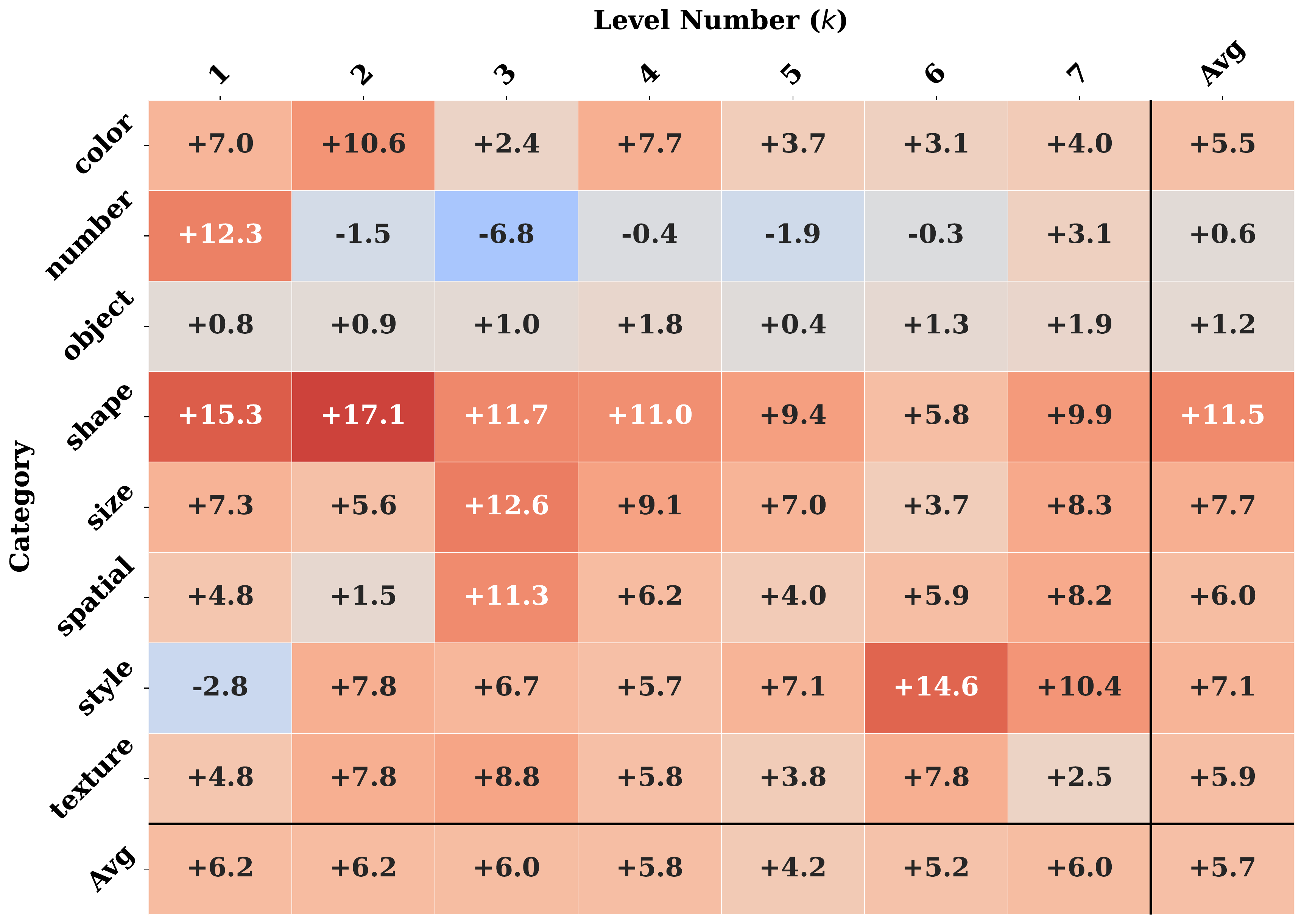}
        \caption{\dalle{}}
        \label{fig:category_heatmap_dalle3}
    \end{subfigure}

    
    \begin{subfigure}[t]{\linewidth}
        \centering
        \includegraphics[width=\linewidth]{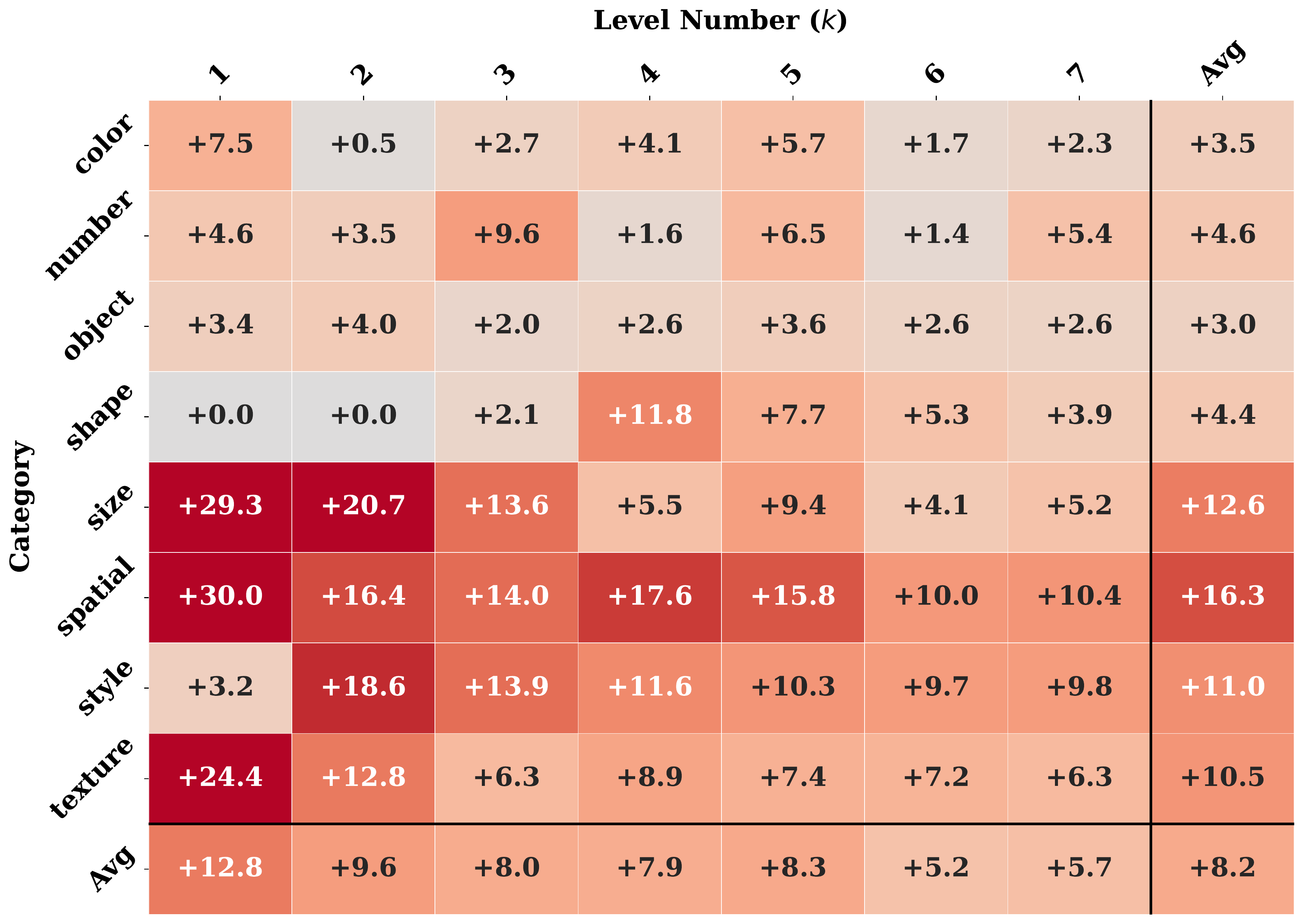}
        \caption{\sd{}}
        \label{fig:category_heatmap_sd}
    \end{subfigure}
    \begin{subfigure}[t]{\linewidth}
        \centering
        \includegraphics[width=\linewidth]{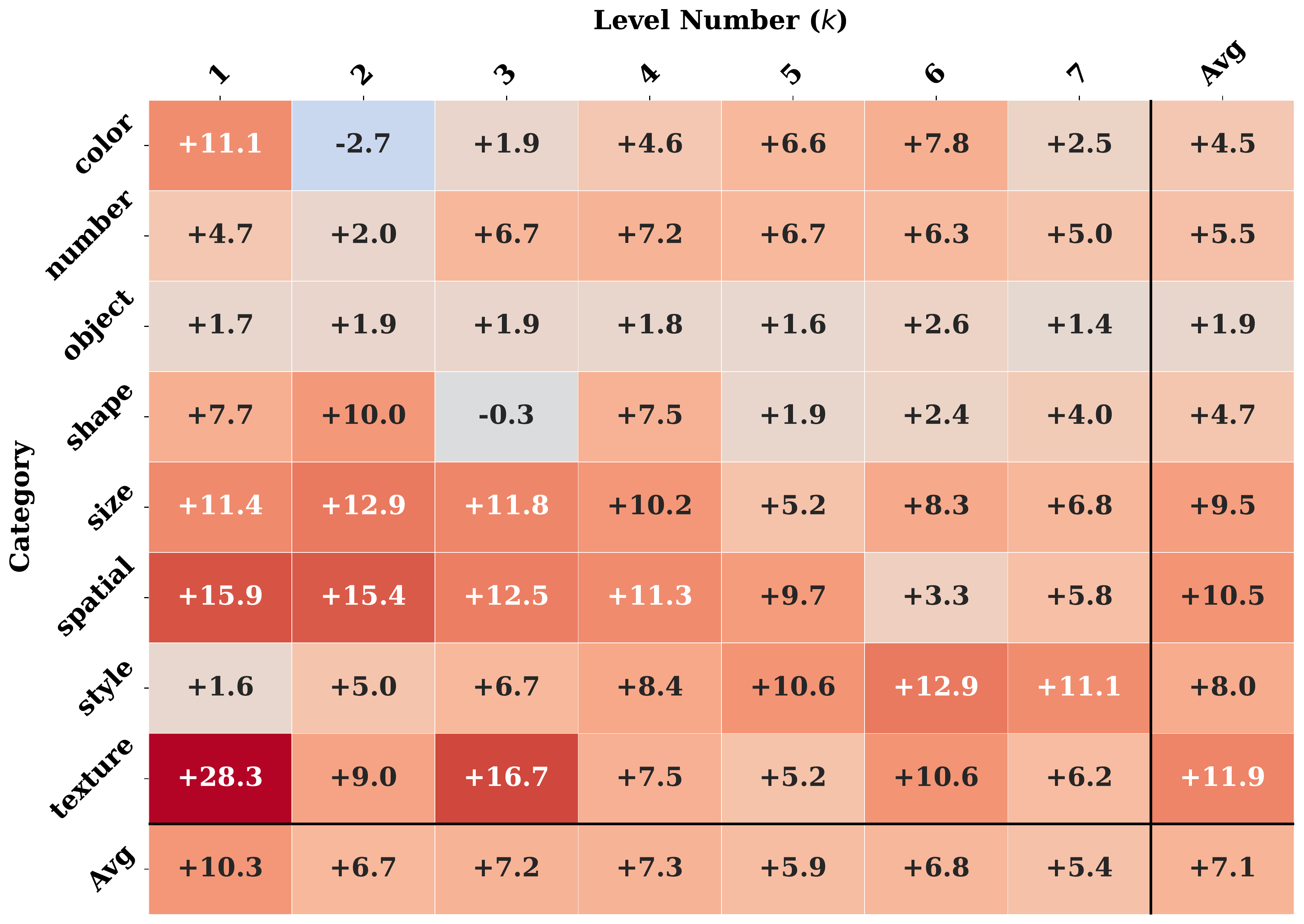}
        \caption{\pg{}}
        \label{fig:category_heatmap_pg}
    \end{subfigure}
    \caption{Heatmaps showing improvement magnitude across visual concept categories and complexity levels. Darker red indicates higher improvements, while blue indicates a decline in performance.
    }
    \label{fig:category_heatmap}
\end{figure}


\subsection{Category-wise Analysis}
\label{sec:category_analysis_main}
To gain deeper insight into the impact of our optimization framework, we conduct a category-wise analysis across eight visual concept categories defined in the ConceptMix benchmark: \textit{color}, \textit{number}, \textit{object}, \textit{shape}, \textit{size}, \textit{spatial}, \textit{style}, and \textit{texture}. 
This analysis allows us to examine how prompt optimization influences specific dimensions of compositional generation and to identify which aspects of diffusion model capabilities benefit most from our framework.

For each diffusion model and complexity level, we compute category-specific scores by evaluating performance on questions associated with each visual concept category. Formally, for a model $M$, complexity level $k$, and category $c$ with its corresponding set of questions $Q_c$, the score $S_{M,k,c}$ for an image $\mathcal{I}$ is defined as:
\begin{equation*}
S_{M,k,c} (\mathcal{I}) = \frac{1}{|Q_c|} \sum_{q \in Q_c} \mathbbm{1}(\mathcal{V} (q |\mathcal{I}) = \texttt{"yes"})
\end{equation*}

Figure~\ref{fig:category_heatmap} presents heatmap visualizations of category-wise improvement achieved by our optimization framework across all complexity levels for the three tested models.

For \dalle{}, the most improvement happens for \textit{shape} with an average improvement of $11.5\%$. Meanwhile, for \sd{}, the most significant improvements are observed in \textit{spatial} and \textit{size} categories, each showing an average gain exceeding $12\%$. For \pg{}, \textit{texture} shows the strongest consistent gains ($+11.9\%$ average). These observations highlight that these specific categories require precise prompt formulations for optimal performance.

On the other hand, the \textit{object} category exhibits the least improvement for all models, with an average gain of $1.2\%$ for \dalle{}, $3\%$ for \sd{} and, and $1.9\%$ for \pg{}. This is mainly because the original prompts already perform well in this category, leaving limited room for further gains. 

Additionally, the \textit{number} category shows relatively marginal improvements across all models. This can be attributed to the well-known challenge that generative models have with accurately representing specific quantities \cite{huang2023t2i, petsiuk2022human, saharia2022photorealistic}, compounded by the fact that specifying a number offers little flexibility for refinement. As a result, the performance in this category remains low both before and after optimization, suggesting that prompt refinement alone is insufficient to overcome this limitation.

The detailed radar chart analysis for each complexity level is provided in Appendix~\ref{sec:category_analysis}.
\subsection{Cross-Model Prompt Transferability}

A key question about our framework is whether the prompts optimized for one model can effectively generalize to another. To investigate this, we design a comprehensive transferability experiment across all model pairs. For each pair of models $(M_{src}, M_{tgt})$, we:
\begin{enumerate}
    \item Run our prompt optimization framework using $M_{src}$ as the backbone diffusion model to obtain optimized prompts $p^*_{src}$
    \item Evaluate these transferred prompts $p^*_{src}$ on target model $M_{tgt}$
    \item Compare the performance against both the original prompt $p_0$ and $M_{tgt}$'s self-optimized prompts $p^*_{tgt}$
\end{enumerate}

This experimental design allows us to systematically assess whether optimized prompts capture universal phrasing patterns that are effective across different model architectures, or if they are model-specific optimizations.

\begin{figure}[ht]
        \centering
        \includegraphics[width=\linewidth]{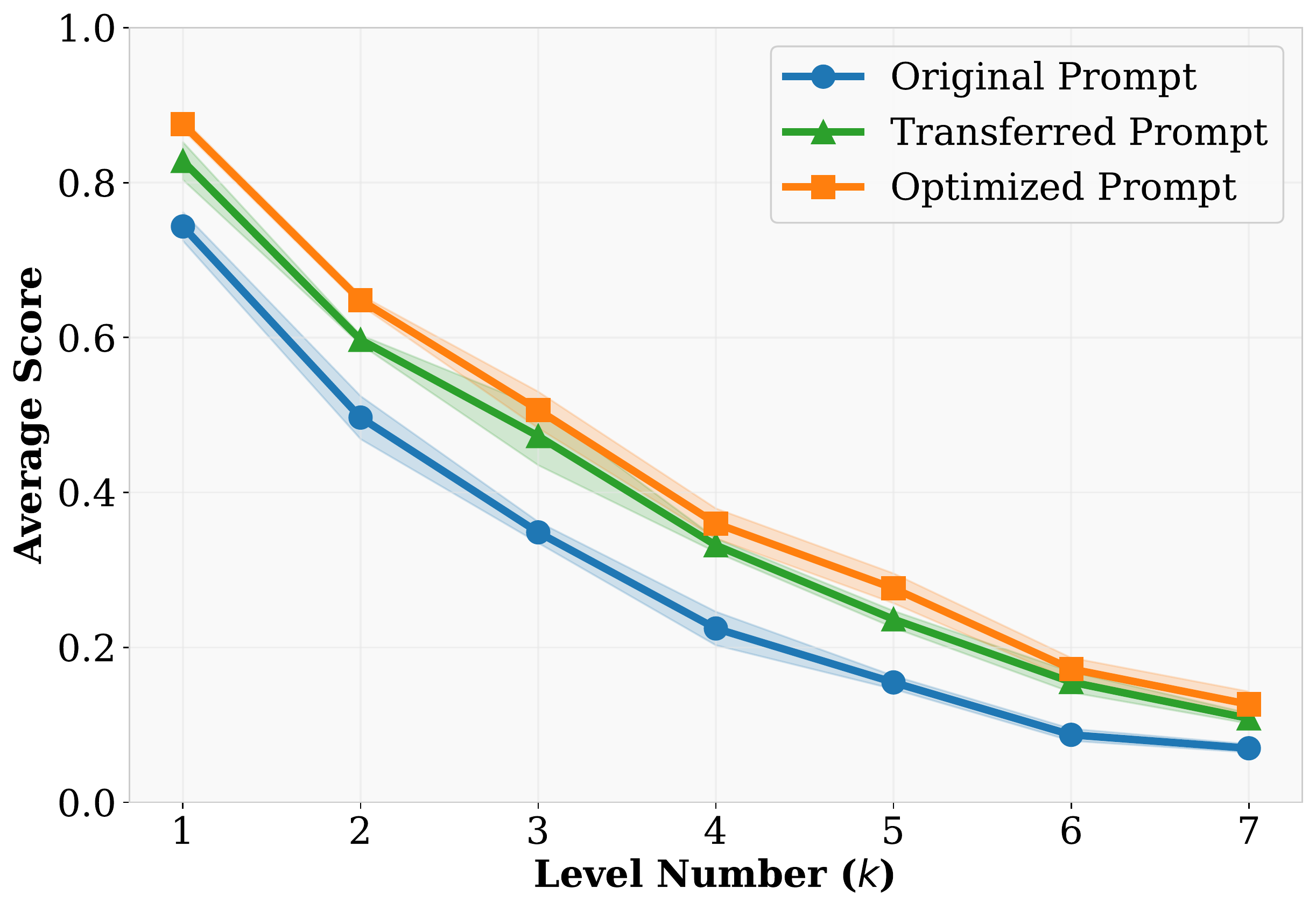}
    \caption{Average score of \sd{} using original prompts, prompts optimized for \sd{} itself, and prompts optimized for \dalle{}.
    }
    \label{fig:sd_dalle3_tr}
\end{figure}

As a representative example, Figure~\ref{fig:sd_dalle3_tr} shows the results of evaluating \sd{} when using prompts that were optimized for \dalle{}. The results compare the average scores of \sd{} using original prompts, prompts optimized for \sd{} itself, and prompts optimized for \dalle{}. We observe that \sd{} performs significantly better with \dalle{}-optimized prompts compared to original prompts, and its performance closely approaches that achieved with prompts optimized for itself. This indicates a high degree of cross-model transferability, suggesting that the optimized prompts capture phrasing patterns that are effective among all models.

Our comprehensive transferability analysis across all model pairs (presented in Appendix~\ref{sec:transferability}) demonstrates that transferred prompts consistently achieve better performance than original prompts in most cases. These findings imply that different diffusion models may share underlying prompt preferences, making cross-model prompt reuse a viable strategy for enhancing generation quality with reduced optimization overhead.

The transferability results also have important practical implications, suggesting that practitioners can leverage a cost-effective optimization workflow: (1) optimize prompts using more accessible or computationally efficient models, and (2) apply these enhanced prompts to more powerful models for final generation. This approach significantly reduces the computational overhead and API costs associated with prompt optimization while maintaining substantial performance improvements.

\subsection{Ablation Study: Optimization Iteration Count}
\label{sec:iteration_ablation}

A key hyperparameter in our prompt optimization framework is the number of optimization iterations $T$. While more iterations intuitively seem beneficial, they also increase computational cost and potentially risk overfitting to the evaluation metric. To investigate the impact of this parameter, we conducted an ablation study on DALL·E 3 with varying numbers of iterations: $T \in \{0, 1, 2, 3, 4, 5, 10, 15\}$, where $T=0$ means we use the original prompt. For this experiment, we fix $k=4$.

\begin{figure}[ht]
    \centering
    \begin{subfigure}[b]{0.48\textwidth}
        \centering
        \includegraphics[width=\textwidth]{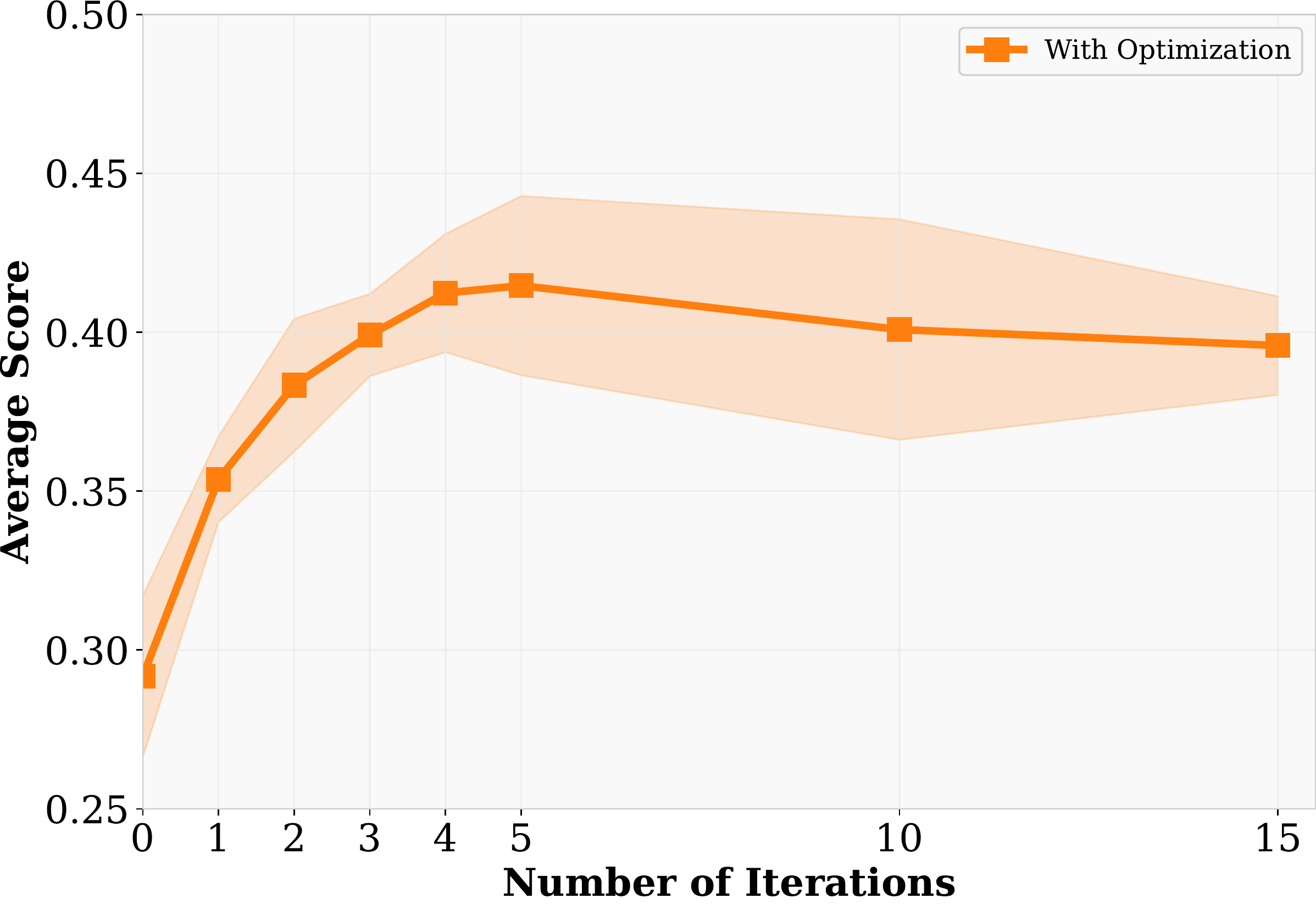}
        \caption{Average Score}
        \label{fig:iteration_ablation_avg_main}
    \end{subfigure}
    \hfill
    \begin{subfigure}[b]{0.48\textwidth}
        \centering
        \includegraphics[width=\textwidth]{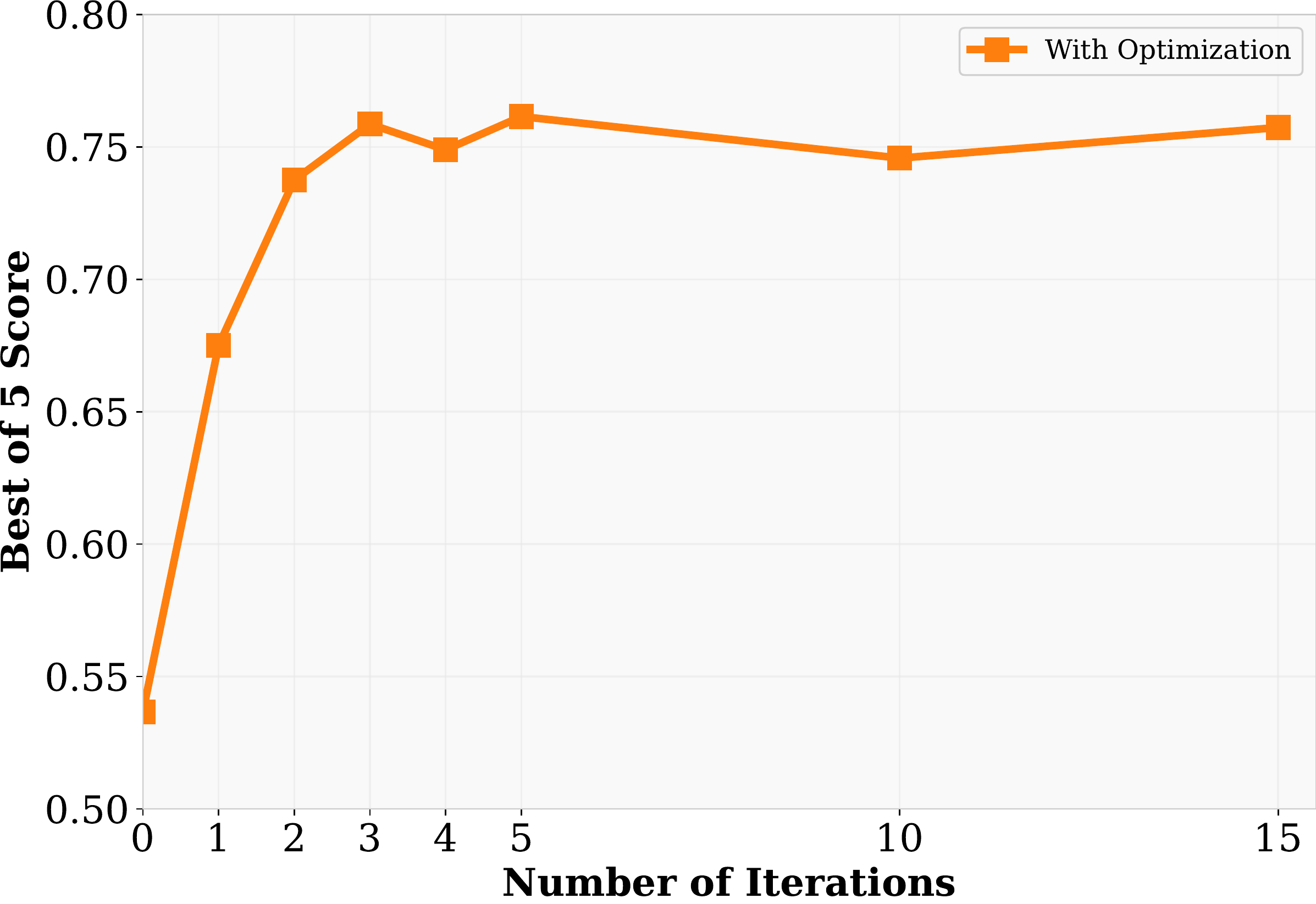}
        \caption{Best-of-5 Score}
        \label{fig:iteration_ablation_best_main}
    \end{subfigure}
    \caption{Performance comparison of different iteration numbers ($T$) in our prompt optimization framework on DALL·E 3 with fixed complexity $k=4$. The x-axis represents the number of iterations, while the y-axis shows scores.}
    \label{fig:iteration_ablation_main}
\end{figure}

From Figure~\ref{fig:iteration_ablation_main}, we observe that higher iteration numbers do not necessarily yield better prompts. This counterintuitive finding can be attributed to several factors:
\begin{enumerate}
    \item \textbf{Inherent Stochasticity:} Diffusion models like DALL·E 3 have inherent randomness in their generation process. A prompt that produces high-quality images during optimization may not consistently yield the same quality during test time, even with the same sampling parameters.
    
    \item \textbf{Overfitting to Specific Instances:} Later iterations may overfit to the specific random seed or generation parameters used during optimization, reducing generalizability to new generation instances.
\end{enumerate}

Based on these results, we find that $T=5$ represents an optimal choice that balances performance gains with computational efficiency for most applications. This finding is particularly valuable for deployment scenarios where optimization time is a concern.
\subsection{Alternative VLM Judge to Mitigate Preference Leakage}
\label{sec:preference_leakage}

A potential concern with our optimization framework is preference leakage -- the possibility that using the same vision-language model (GPT-4o) for both prompt optimization and final evaluation could introduce bias. Specifically, the optimization process might favor prompts that align with GPT-4o's particular evaluation preferences rather than genuinely improving visual generation quality. To address this concern and validate that our observed improvements reflect true visual synthesis enhancement, we conduct an additional experiment using InternVL3-8B \cite{zhu2025internvl3} as the optimization judge while maintaining GPT-4o as the final evaluator.

We repeat our optimization process for all three models (DALL·E 3, Stable Diffusion 3.5, and Playground v2.5) at complexity level $k=4$ using InternVL3-8B to provide scores and feedback during the iterative refinement process. All other experimental conditions remain identical to our main setup. The optimized prompts are then evaluated using GPT-4o to maintain consistency with our evaluation protocol.

\begin{figure}[ht]
    \centering
    \includegraphics[width=\linewidth]{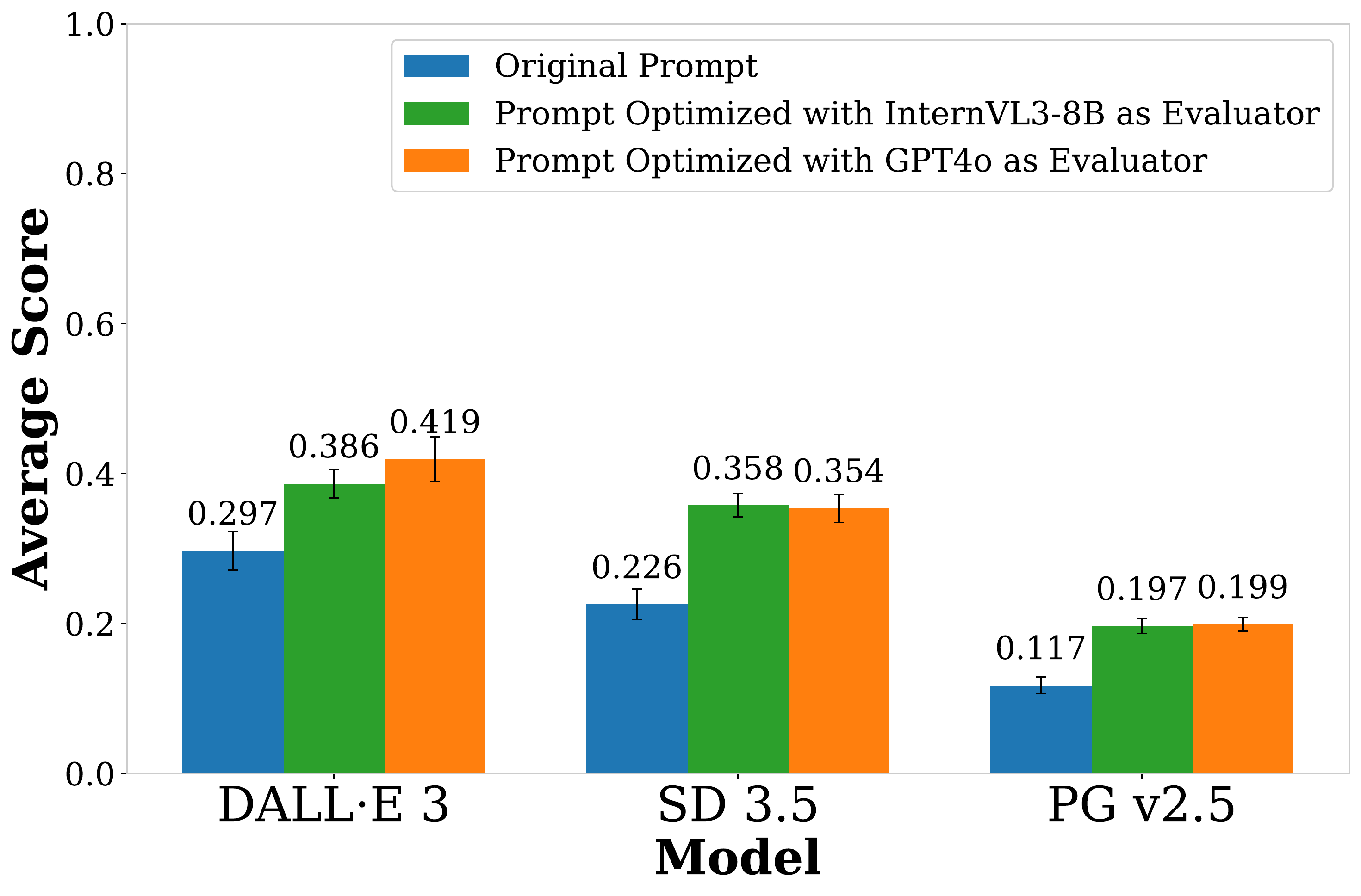}
    \caption{Performance comparison between prompts optimized using GPT-4o versus InternVL3-8B as the optimization judge, evaluated on GPT-4o. Results show substantial improvements persist even when using a different VLM for optimization.}
    \label{fig:internvl_comparison}
\end{figure}

Figure \ref{fig:internvl_comparison} presents the results of this experiment. The key finding is that prompts optimized using InternVL3-8B as the judge still achieve substantial performance improvements, which are nearly identical to prompts optimized by GPT-4o when evaluated by GPT-4o. Specifically, across all three models, InternVL3-8B-optimized prompts achieve performance levels that closely match or exceed GPT-4o-optimized prompts, with DALL·E 3 showing 0.386 vs 0.419, Stable Diffusion 3.5 showing 0.358 vs 0.354, and Playground v2.5 showing 0.197 vs 0.199.

These results demonstrate that the performance improvements from prompt optimization are not merely artifacts of evaluator bias or preference leakage. The fact that prompts optimized with a different VLM (InternVL3-8B) still yield substantial improvements when evaluated by GPT-4o provides strong evidence that our framework genuinely enhances visual synthesis capabilities rather than simply exploiting evaluation-specific preferences. The slight performance differences between GPT-4o-optimized and InternVL3-8B-optimized prompts are expected, as different VLMs may have varying strengths in identifying specific visual concepts. However, the remarkably similar performance levels validate that our optimization framework reveals genuine model capabilities that are underestimated by rigid prompt formulations, regardless of the specific evaluator used during optimization.

\section{Discussion}

Our comprehensive evaluation through ConceptMix++ reveals several important insights about the current state of text-to-image generation and evaluation methodologies.

\subsection{Implications for Model Evaluation}

The substantial performance gains achieved through prompt optimization across all tested models highlight a fundamental issue with current benchmarking practices. The fact that optimized prompts can improve performance by up to 20\% suggests that existing benchmarks may be systematically underestimating model capabilities. This finding has significant implications for how we understand progress in the field -- models that appear inferior under rigid benchmarking may actually possess comparable or superior visual synthesis abilities when evaluated under optimal conditions.

The category-wise analysis further emphasizes this point. Our results show that spatial relationships, shapes, and textures are particularly sensitive to prompt formulation, with these categories showing the largest improvements after optimization. This suggests that previous comparative studies may have drawn inaccurate conclusions about model strengths and weaknesses in these specific domains. For instance, a model that appears to struggle with spatial reasoning under standard prompts might actually excel in this area when provided with appropriately optimized instructions.

\subsection{Cross-Model Insights and Transferability}

The remarkable transferability of optimized prompts across different model architectures reveals shared underlying preferences for effective prompt phrasing. This finding suggests that despite architectural differences, modern diffusion models may have converged on similar representations of visual concepts and language understanding. The transferability also has practical implications, enabling cost-effective optimization workflows where prompts can be optimized on more accessible models and then applied to more powerful or expensive alternatives.

However, the asymmetric nature of transfer effectiveness shown in Appendix \ref{sec:transferability} provides insights into model-specific capabilities. The observation that DALL·E 3-optimized prompts transfer less effectively to other models suggests that this model may have developed more specialized prompt understanding mechanisms, potentially due to its architectural choices or capability difference.

\subsection{Persistent Limitations and Future Directions}

While prompt optimization significantly improves performance across most categories, certain limitations persist. The marginal improvements in the number category across all models confirm that counting remains a fundamental challenge that cannot be easily addressed through better prompting alone. This suggests that architectural innovations specifically targeting numerical reasoning may be necessary.

The diminishing returns at higher complexity levels ($k=6$, $k=7$) reveal another important limitation. Even with optimal prompting, models struggle with highly complex compositional scenes, indicating that current architectures may have inherent capacity constraints that limit their ability to handle multiple simultaneous requirements.

\section{Limitations}

While ConceptMix++ provides valuable insights into text-to-image model capabilities, several limitations should be acknowledged:

\subsection{Evaluation Methodology Constraints}

Our framework relies on GPT-4o as both the optimizer and evaluator, which introduces potential biases. Although our transferability experiments suggest that improvements are genuine rather than artifacts of evaluator preferences, the use of a single VLM for evaluation may not capture all aspects of visual quality that human evaluators would consider. Future work could incorporate multiple evaluation models or human assessment to provide more comprehensive validation.

\subsection{Computational Overhead}

The prompt optimization process requires additional computational resources compared to standard benchmarking. While our budget-constrained experiments demonstrate that optimization can be beneficial even with limited resources, the iterative nature of our approach does increase evaluation time and cost. This may limit the practical applicability of our framework for large-scale benchmarking studies.

\subsection{Prompt Distribution Bias}

Our optimization process may push prompts toward distributions that differ significantly from typical user inputs. While this reveals model capabilities, it may not reflect real-world usage patterns. The optimized prompts might become overly technical or artificial, potentially limiting their relevance for practical applications.

\subsection{Limited Model Coverage}

Our experiments focus on three state-of-the-art diffusion models. While these represent different architectural approaches, the generalizability of our findings to other model families (e.g., autoregressive models, GANs) remains unclear. Additionally, our analysis is limited to the ConceptMix benchmark, and the effectiveness of our approach on other evaluation frameworks requires further investigation.

\subsection{Category Definition Constraints}

The eight visual concept categories defined in ConceptMix, while comprehensive, may not capture all relevant aspects of visual generation. Our category-wise analysis is inherently limited by these predefined categories, and important capabilities or limitations might exist outside this framework.

\section{Conclusion}
In this work, we introduce \method{}, a novel framework for fair benchmarking of text-to-image diffusion models through iterative prompt optimization. Our experiments show that \method{} consistently enhances model performance, revealing their true compositional generation capabilities. We further analyze the impact of our framework across different visual concept categories, identifying which aspects benefit most from prompt refinement. Additionally, we demonstrate that optimized prompts exhibit strong cross-model transferability, suggesting shared prompt preferences among models. These findings indicate that fixed-prompt benchmarks substantially underestimate the capabilities of text-to-image models and highlight the critical role of prompt formatting in fully realizing their generative potential.

\section{Acknowledgements}
We would like to thank Microsoft for an Accelerating Foundation Models Research grant that provided the OpenAI credits enabling this work.
\newpage
{
    \small
    \bibliographystyle{ieeenat_fullname}
    \bibliography{output}
}
\newpage
\appendix
\clearpage
\setcounter{page}{1}
\maketitlesupplementary
\appendix
\input{sec/T2I_details}
\section{Extended Model Performance Analysis}
\label{sec:model_charts}

\begin{figure*}[htbp]
    \centering
    \begin{subfigure}[b]{0.49\textwidth}
        \centering
        \includegraphics[width=\textwidth]{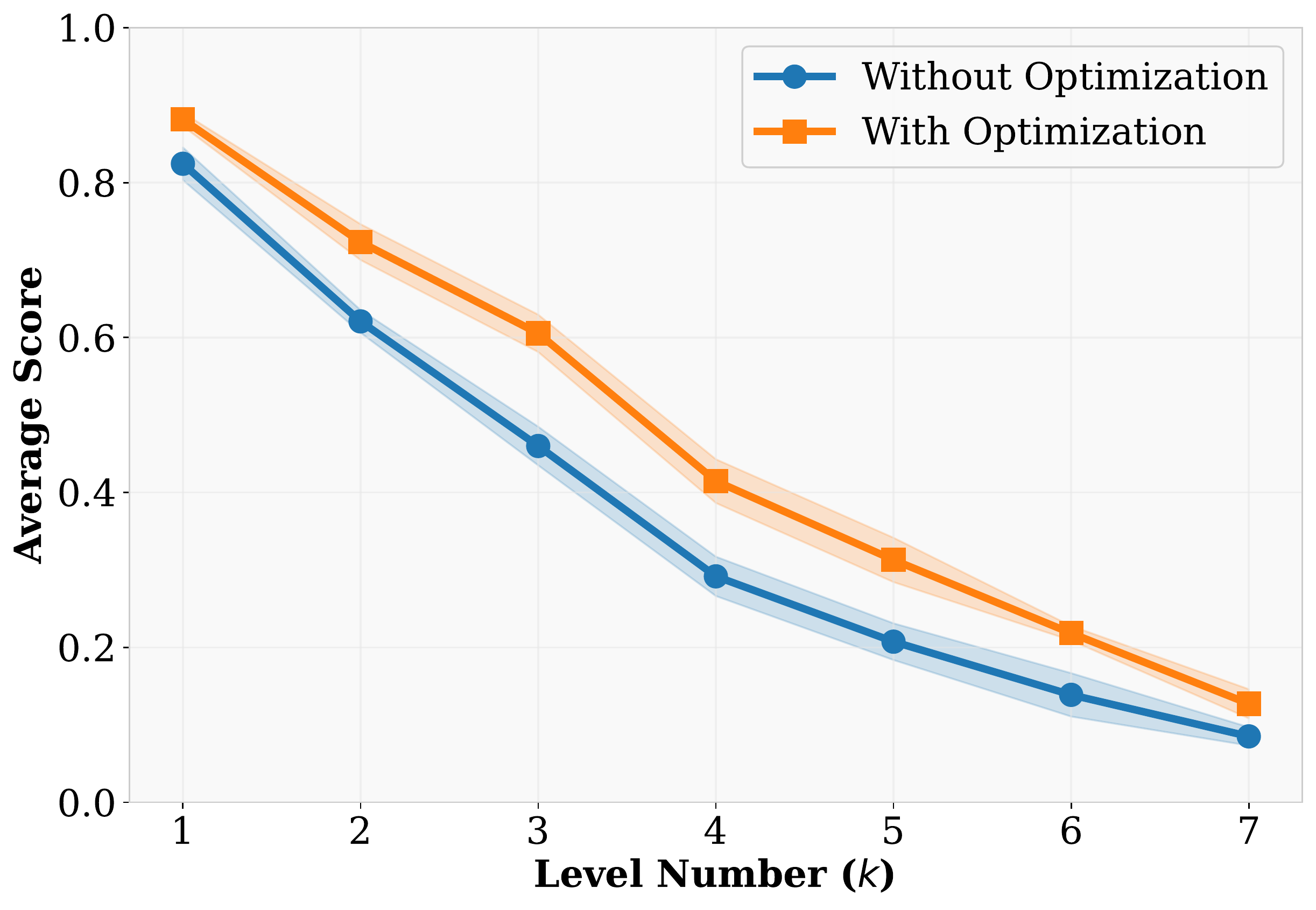}
        \caption{DALL·E 3: Average Score}
        \label{fig:dalle3_avg}
    \end{subfigure}
    \hfill
    \begin{subfigure}[b]{0.49\textwidth}
        \centering
        \includegraphics[width=\textwidth]{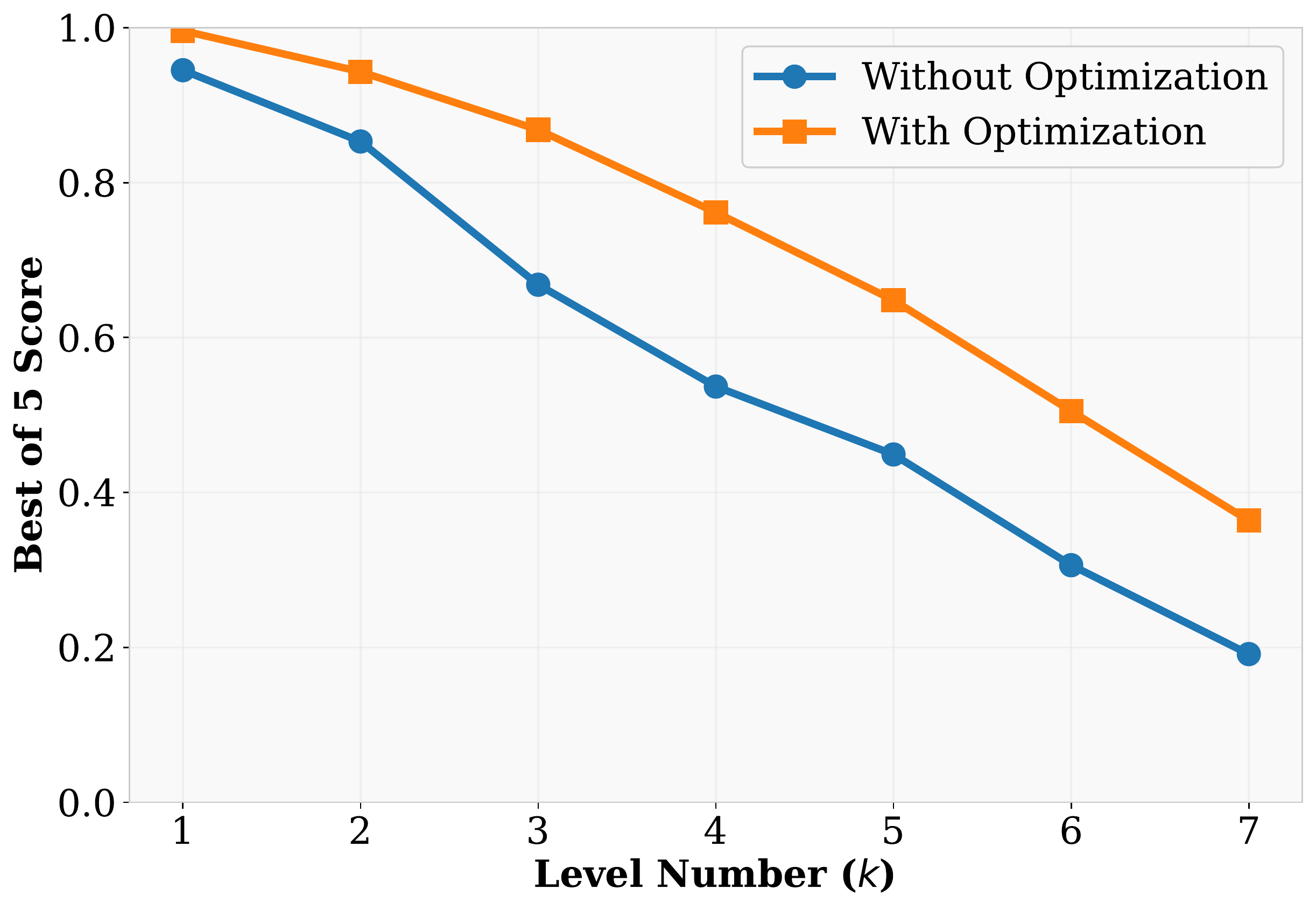}
        \caption{DALL·E 3: Best-of-5 Score}
        \label{fig:dalle3_best}
    \end{subfigure}
    
    \vspace{0.5cm}
    
    \begin{subfigure}[b]{0.49\textwidth}
        \centering
        \includegraphics[width=\textwidth]{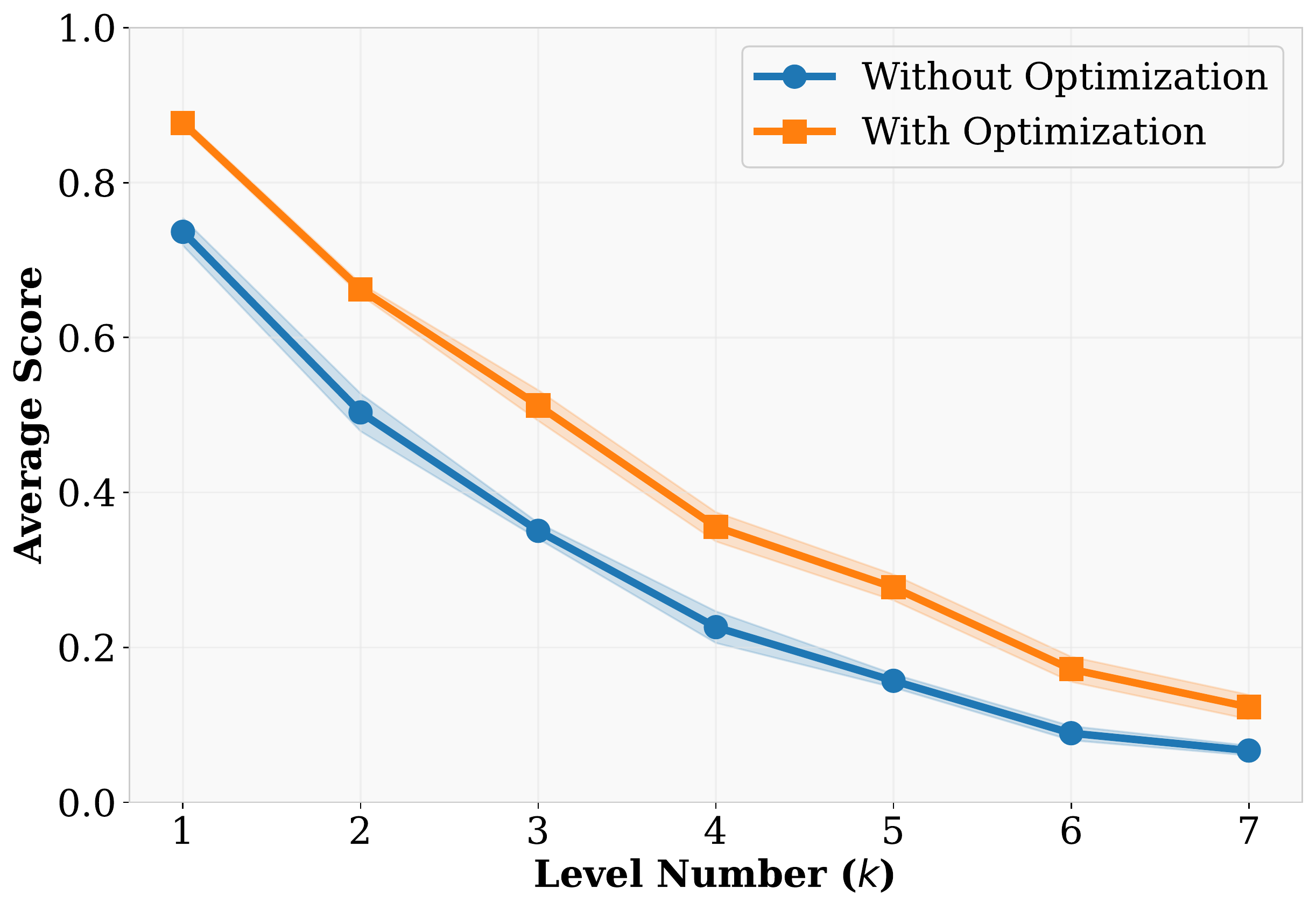}
        \caption{SD 3.5: Average Score}
        \label{fig:sd35_avg}
    \end{subfigure}
    \hfill
    \begin{subfigure}[b]{0.49\textwidth}
        \centering
        \includegraphics[width=\textwidth]{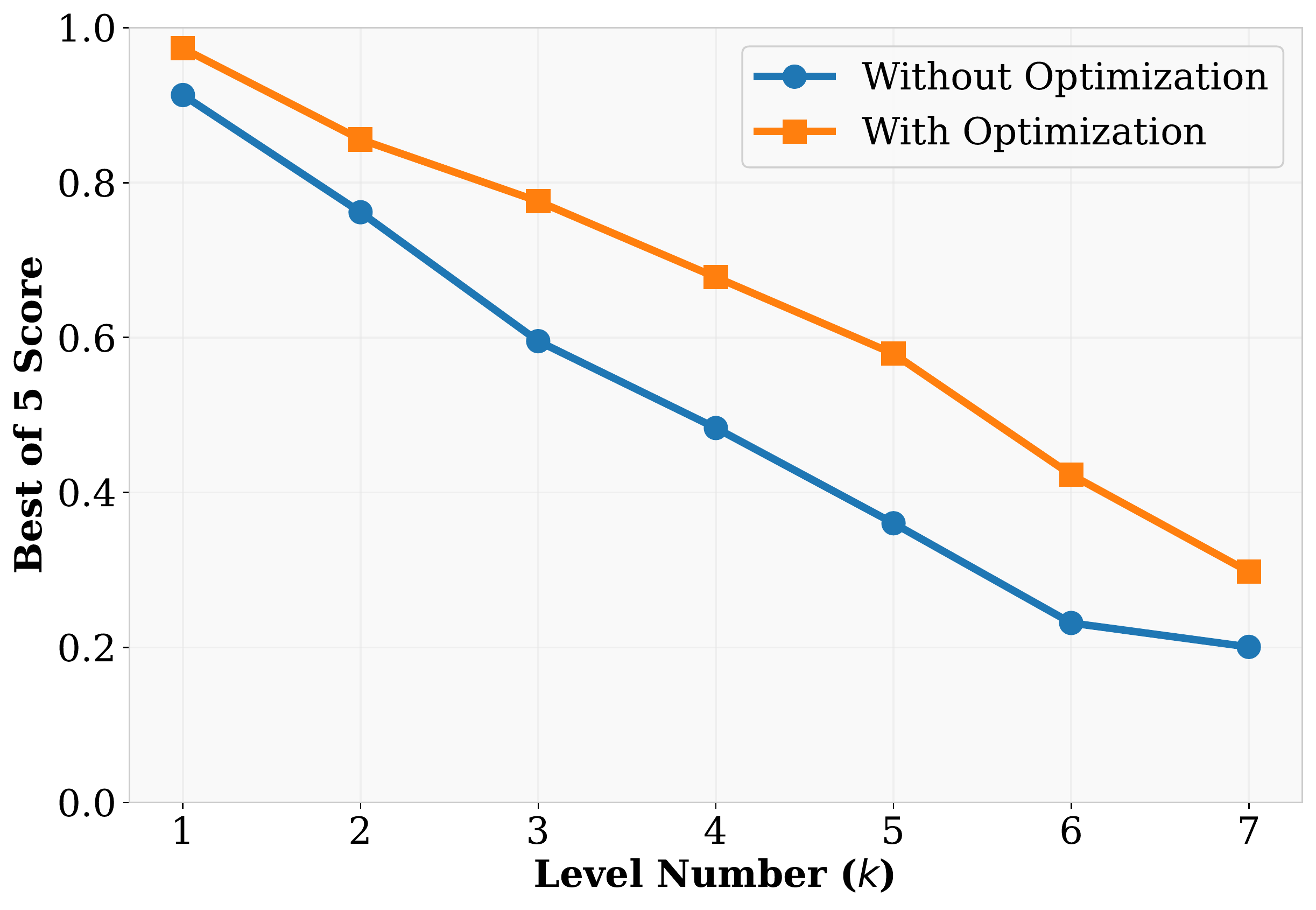}
        \caption{SD 3.5: Best-of-5 Score}
        \label{fig:sd35_best}
    \end{subfigure}
    
    \vspace{0.5cm}
    
    \begin{subfigure}[b]{0.49\textwidth}
        \centering
        \includegraphics[width=\textwidth]{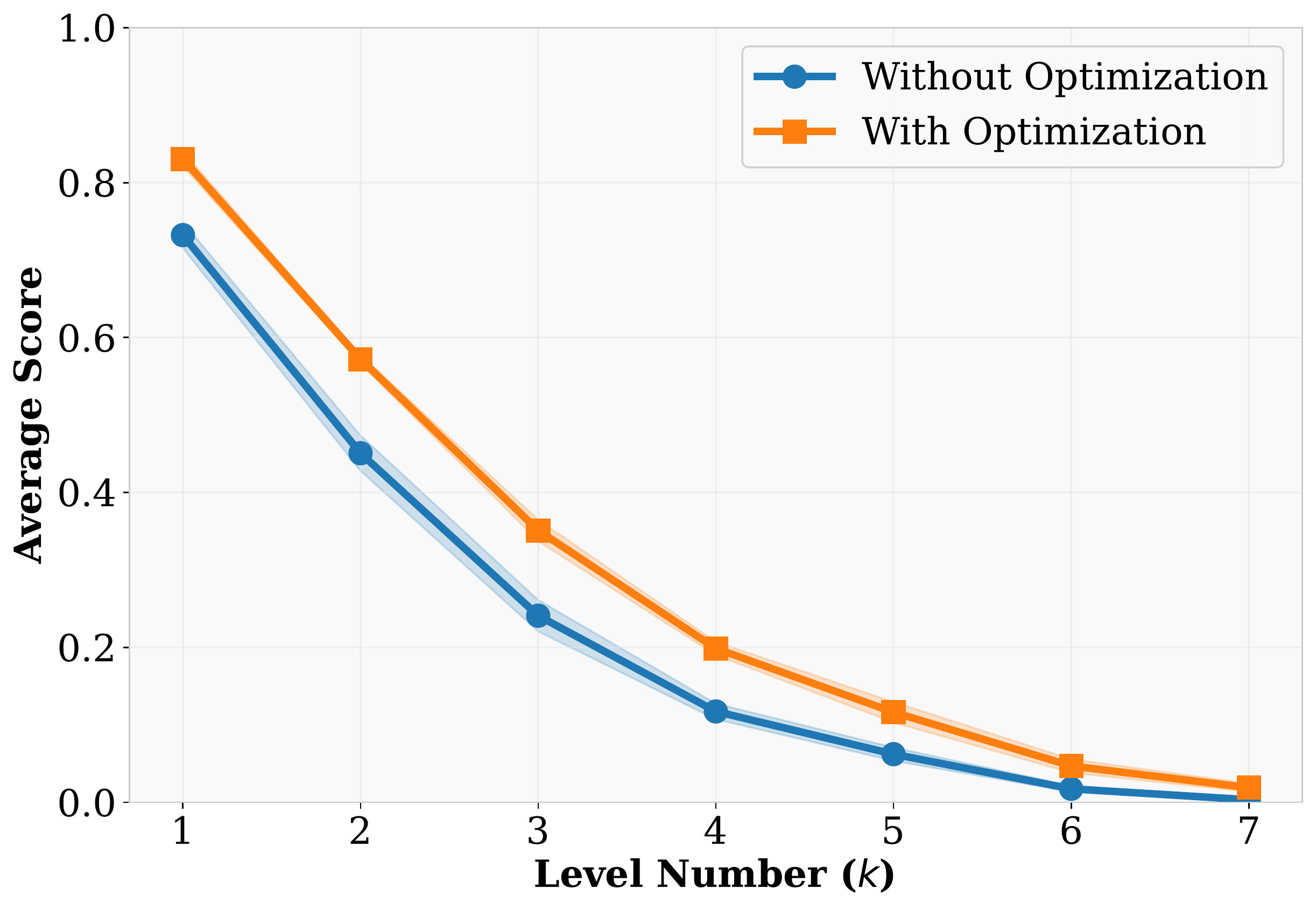}
        \caption{Playground v2.5: Average Score}
        \label{fig:playground_avg}
    \end{subfigure}
    \hfill
    \begin{subfigure}[b]{0.49\textwidth}
        \centering
        \includegraphics[width=\textwidth]{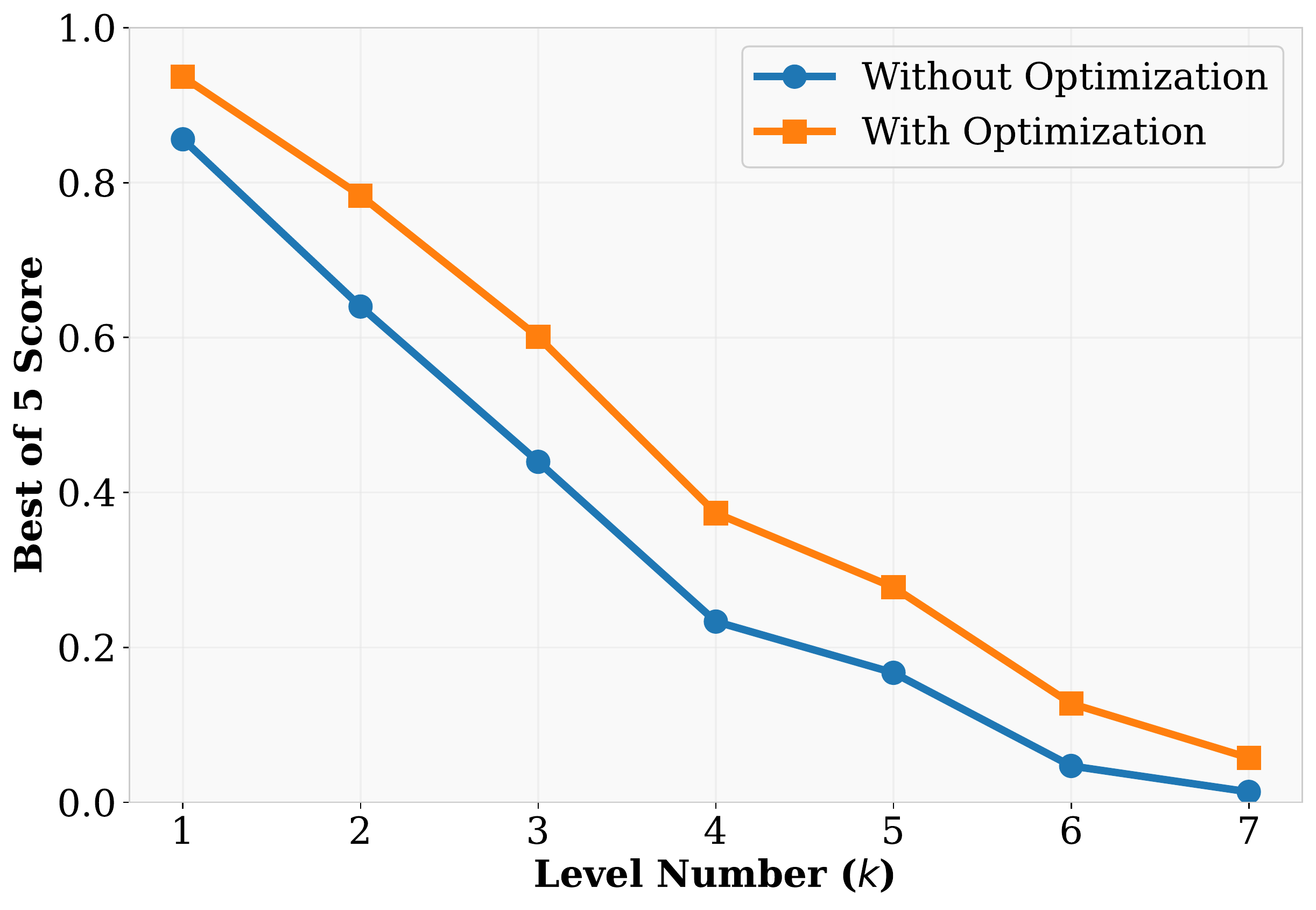}
        \caption{Playground v2.5: Best-of-5 Score}
        \label{fig:playground_best}
    \end{subfigure}
    
    \caption{Performance comparison across models and complexity levels, showing original vs. optimized prompt performance. Left column: Average Score; Right column: Best-of-5 Score. Top row: DALL·E 3; Middle row: Stable Diffusion 3.5; Bottom row: Playground v2.5.}
    \label{fig:model_comparison}
\end{figure*}

Figure~\ref{fig:model_comparison} provides comprehensive visualization for performance comparisons across all tested diffusion models, showing both \textit{mean} score and \textit{best-of-5} score metrics before and after prompt optimization across complexity levels ($k=1$ to $k=7$).

\section{Additional radar graph for Category-wise analysis}
\label{sec:category_analysis}

\begin{figure*}[htbp]
    \centering
     \begin{subfigure}[b]{0.3\textwidth}
        \centering
        \includegraphics[width=\textwidth]{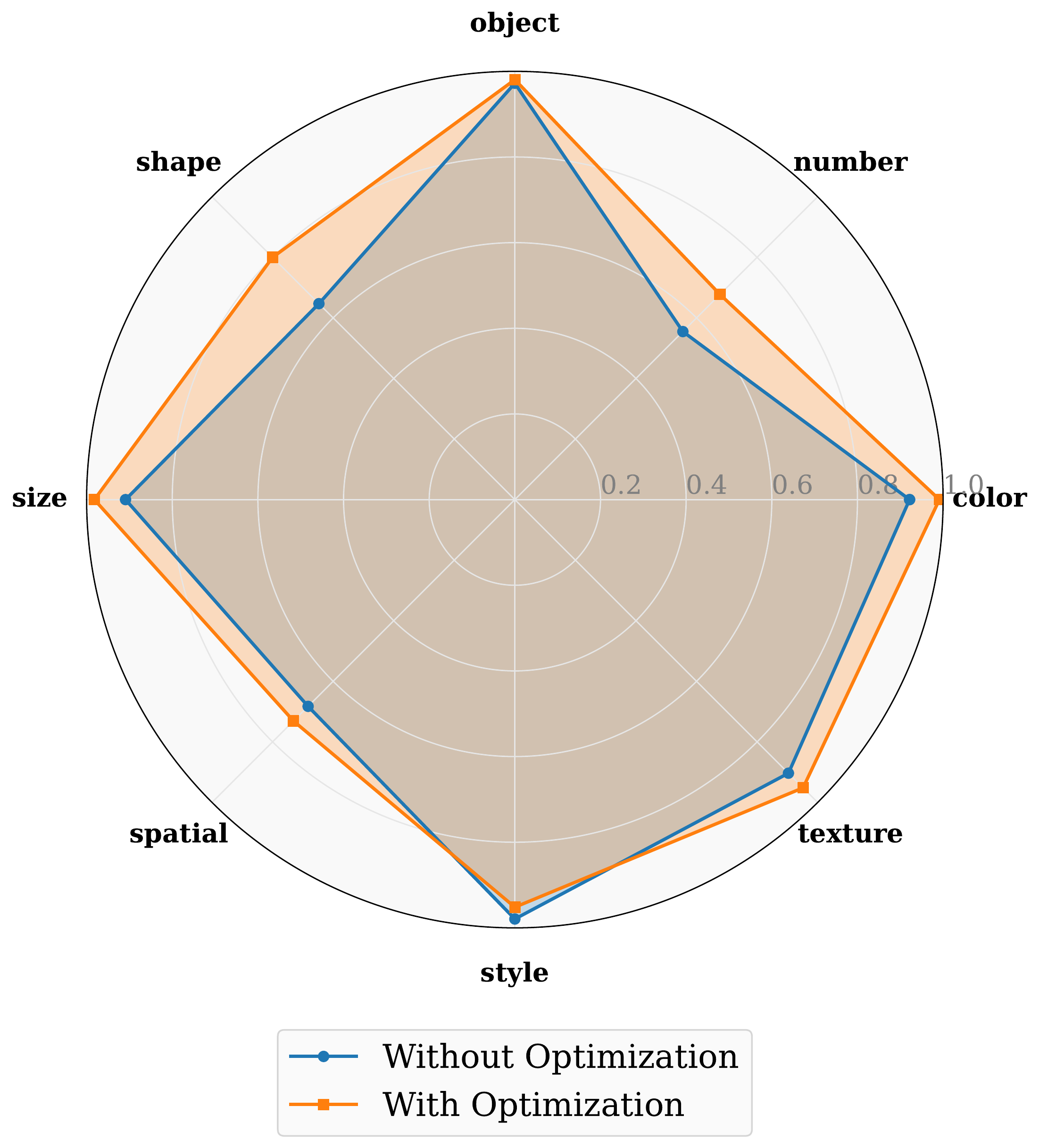}
        \caption{DALL·E 3}
        \label{fig:dalle3_radar_k1}
    \end{subfigure}
    \hfill
     \begin{subfigure}[b]{0.3\textwidth}
        \centering
        \includegraphics[width=\textwidth]
        {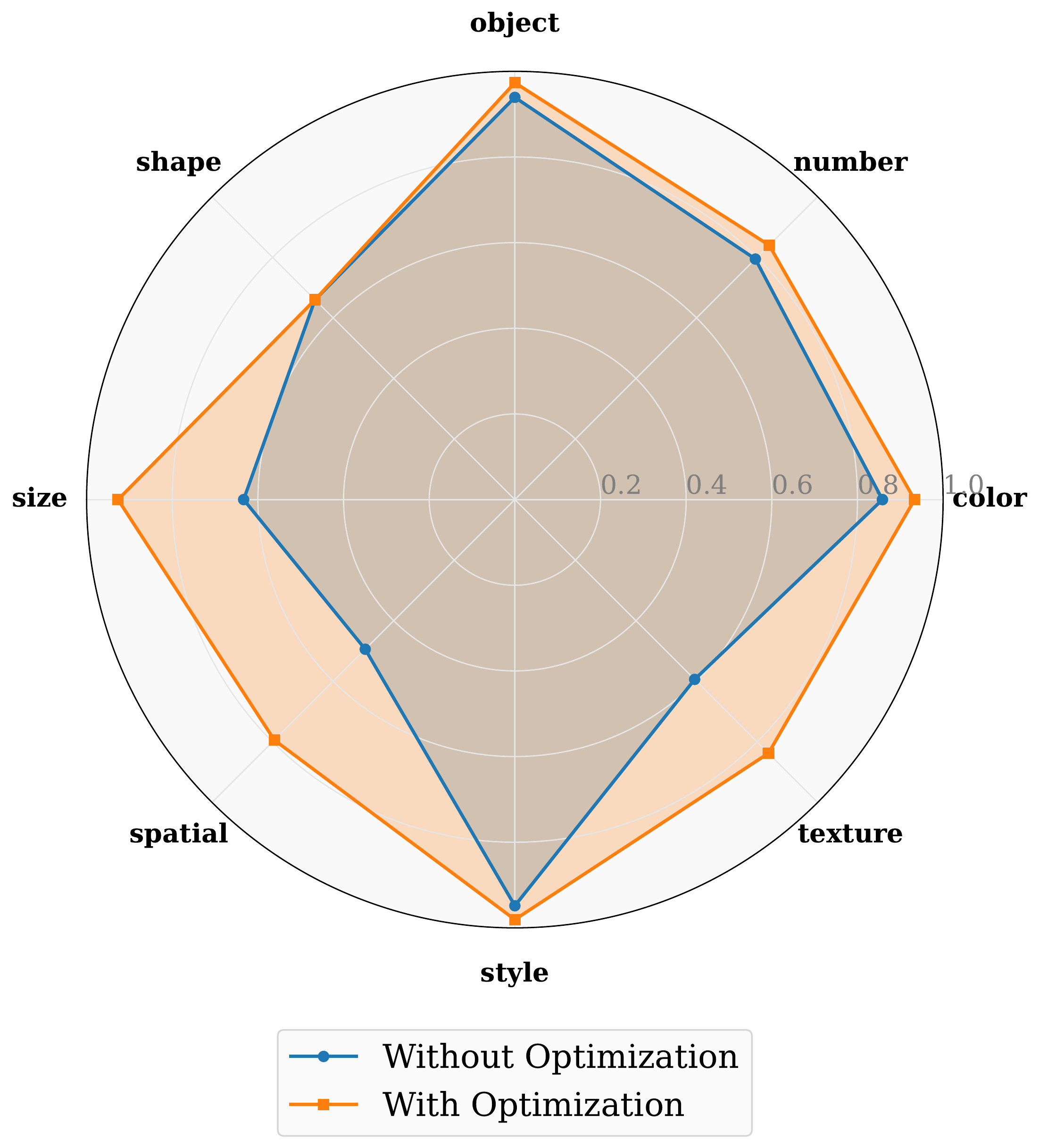}
        \caption{Stable Diffusion 3.5}
        \label{fig:sd35_radar_k1}
    \end{subfigure}
    \hfill
    \begin{subfigure}[b]{0.3\textwidth}
        \centering
        \includegraphics[width=\textwidth]
        {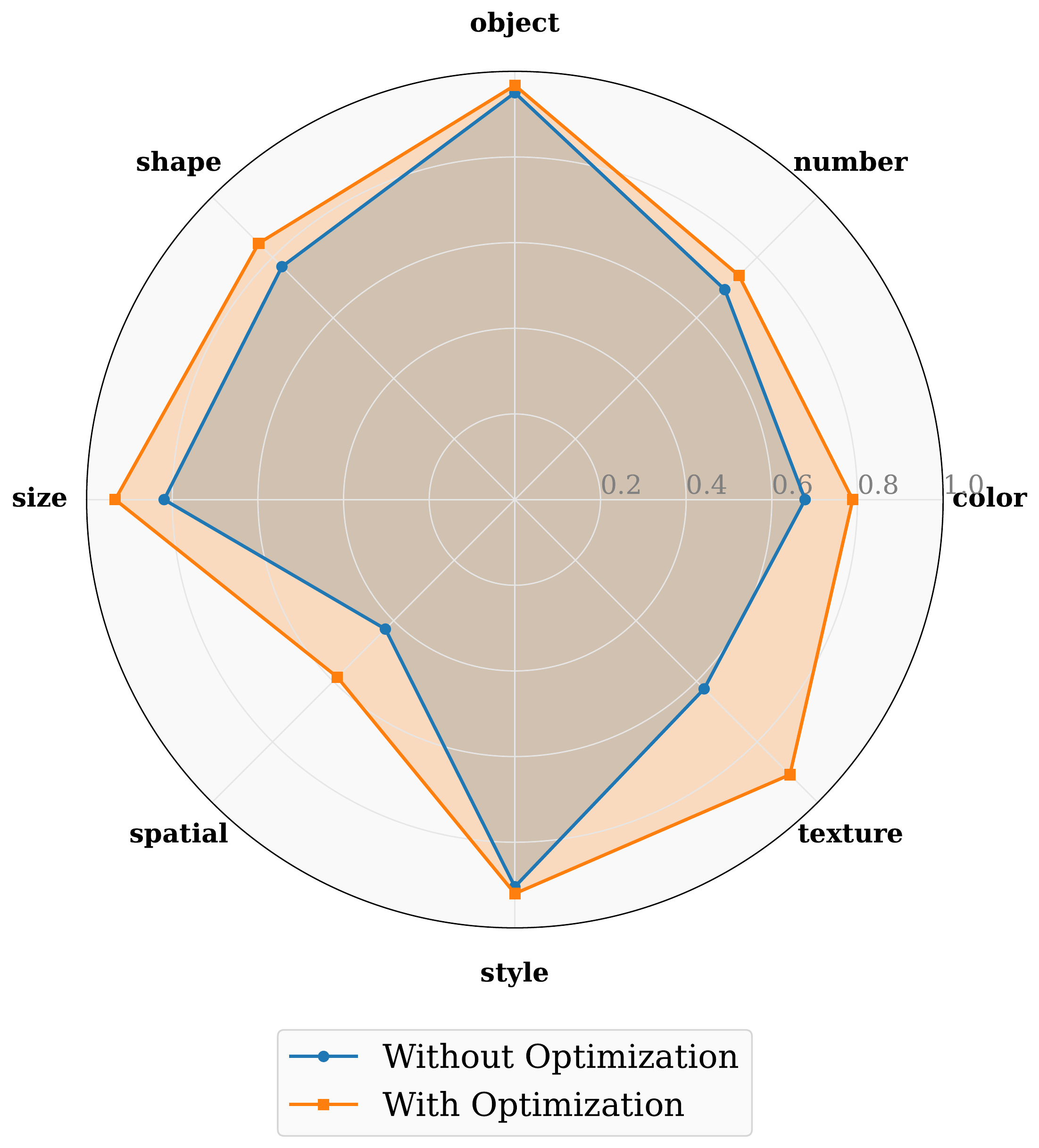}
        \caption{Playground v2.5}
        \label{fig:playground_radar_k1}
    \end{subfigure}
    \caption{Radar graphs showing category improvements for complexity level $k=1$ across all three models.}
    \label{fig:radar_k1}
\end{figure*}

\begin{figure*}[htbp]
    \centering
    \begin{subfigure}[b]{0.3\textwidth}
        \centering
        \includegraphics[width=\textwidth]{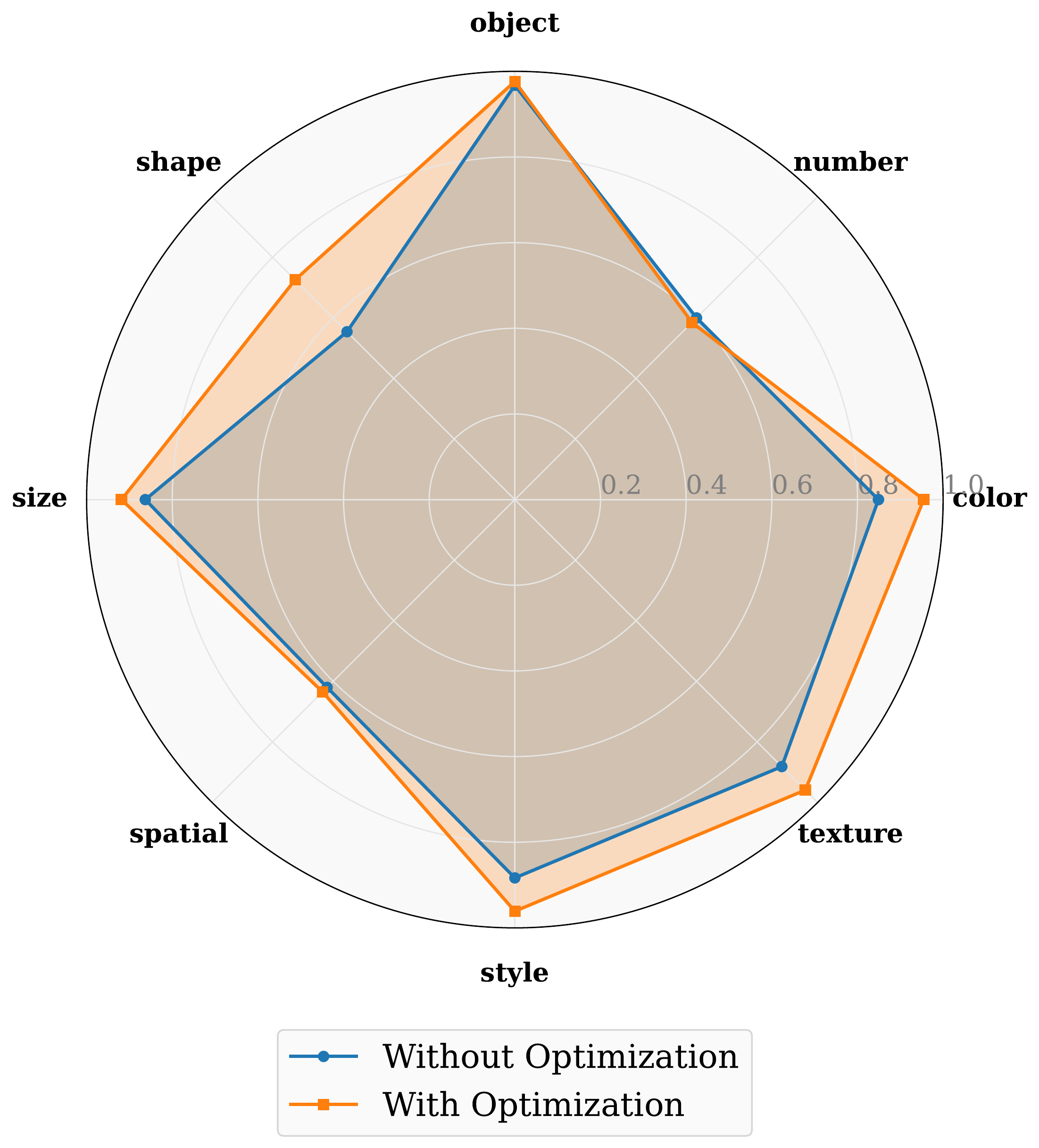}
        \caption{DALL·E 3}
        \label{fig:dalle3_radar_k2}
    \end{subfigure}
    \hfill
     \begin{subfigure}[b]{0.3\textwidth}
        \centering
        \includegraphics[width=\textwidth]{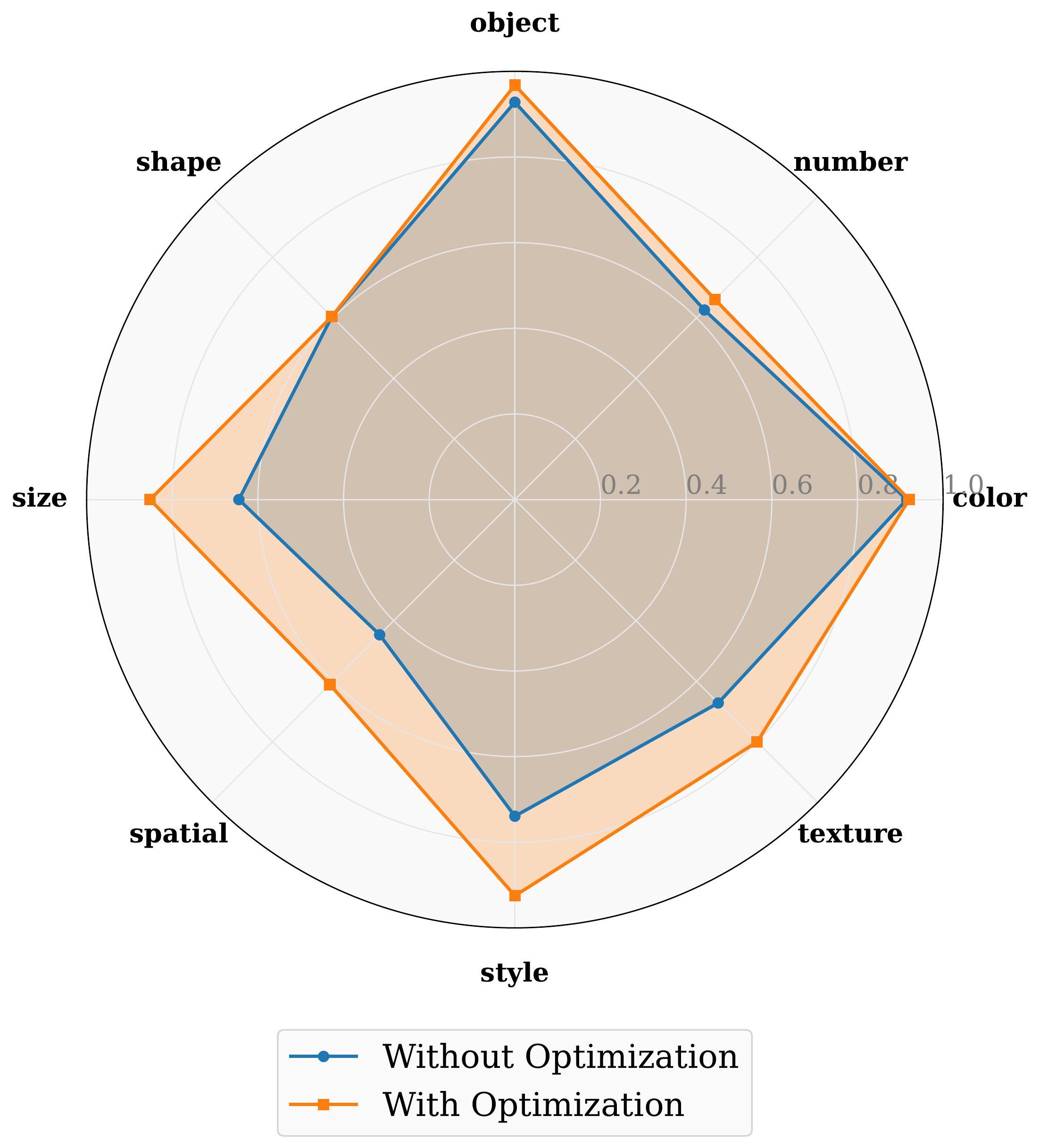}
        \caption{Stable Diffusion 3.5}
        \label{fig:sd35_radar_k2}
    \end{subfigure}
    \hfill
    \begin{subfigure}[b]{0.3\textwidth}
        \centering
        \includegraphics[width=\textwidth]{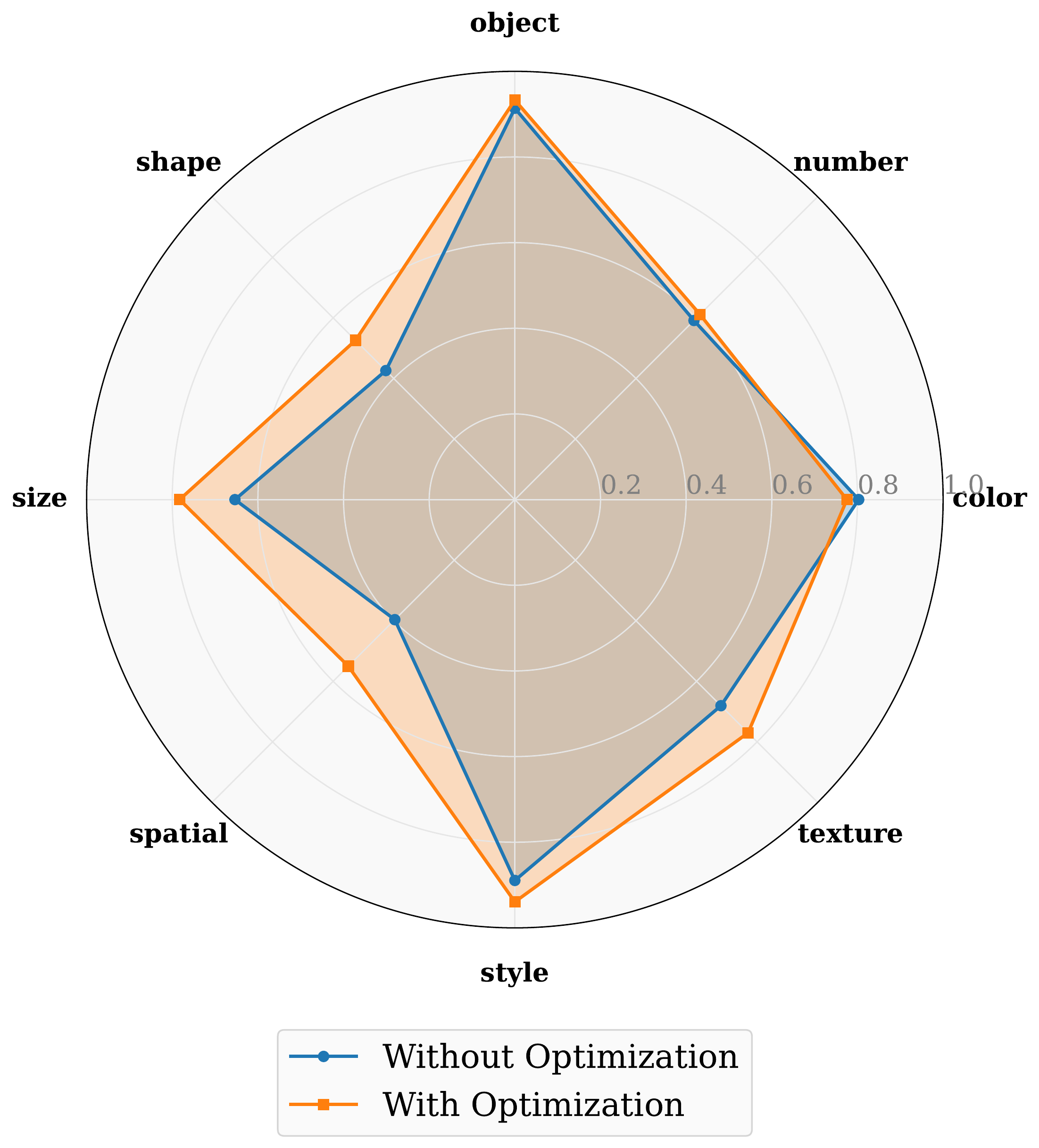}
        \caption{Playground v2.5}
        \label{fig:playground_radar_k2}
    \end{subfigure}
    \caption{Radar graphs showing category improvements for complexity level $k=2$ across all three models.}
    \label{fig:radar_k2}
\end{figure*}

\begin{figure*}[htbp]
    \centering
    \begin{subfigure}[b]{0.3\textwidth}
        \centering
        \includegraphics[width=\textwidth]{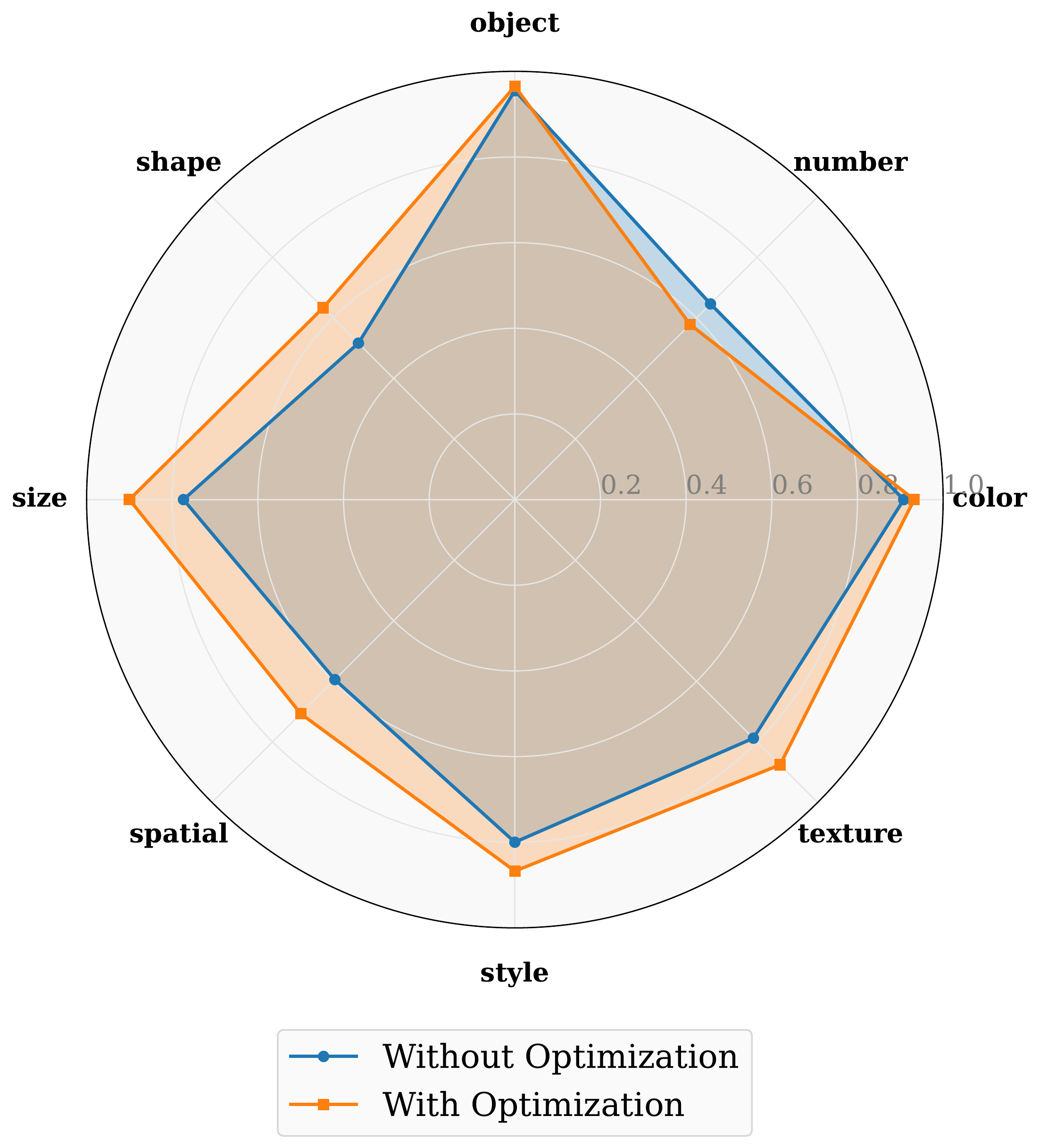}
        \caption{DALL·E 3}
        \label{fig:dalle3_radar_k3}
    \end{subfigure}
    \hfill
     \begin{subfigure}[b]{0.3\textwidth}
        \centering
        \includegraphics[width=\textwidth]{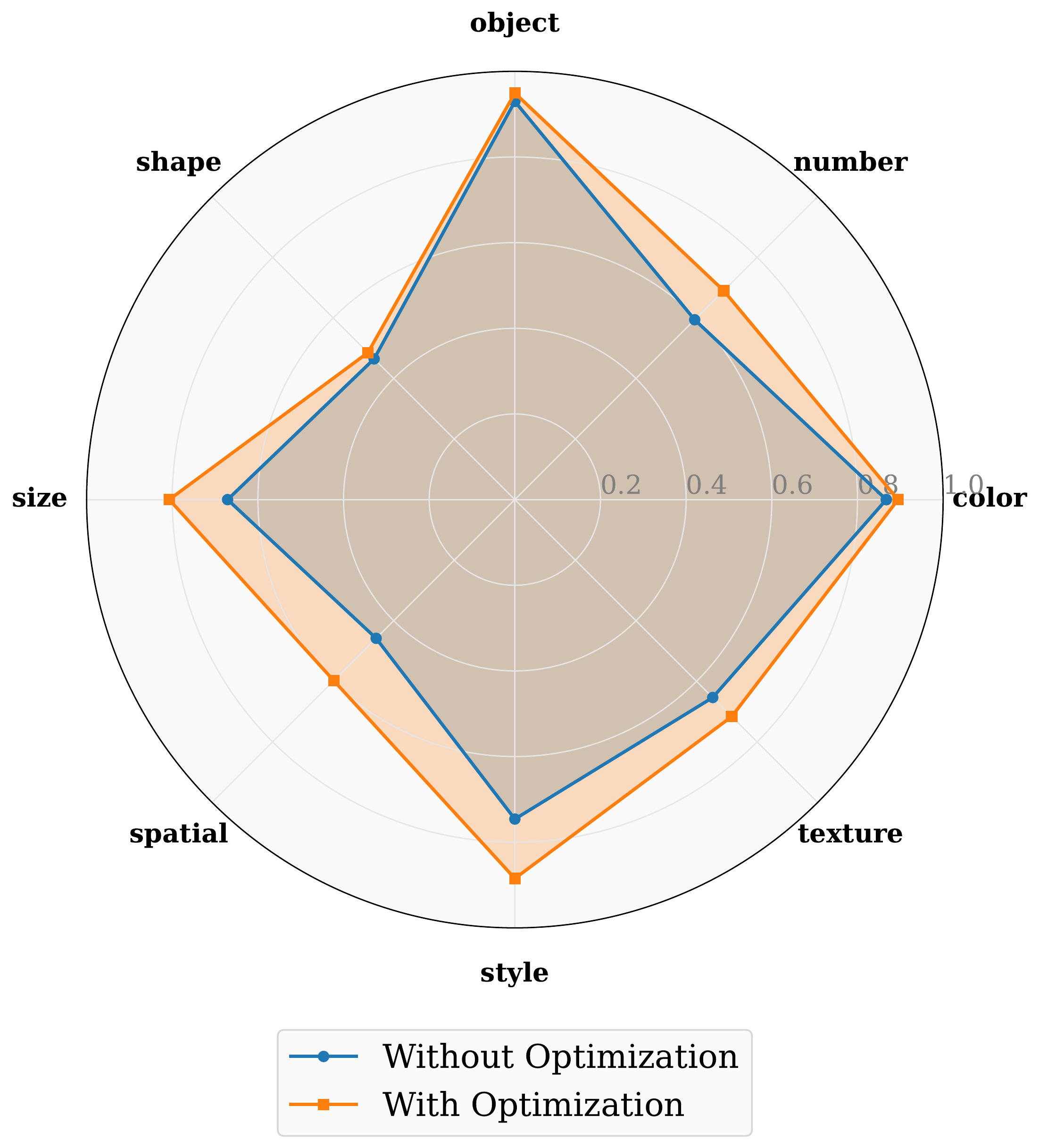}
        \caption{Stable Diffusion 3.5}
        \label{fig:sd35_radar_k3}
    \end{subfigure}
    \hfill
    \begin{subfigure}[b]{0.3\textwidth}
        \centering
        \includegraphics[width=\textwidth]{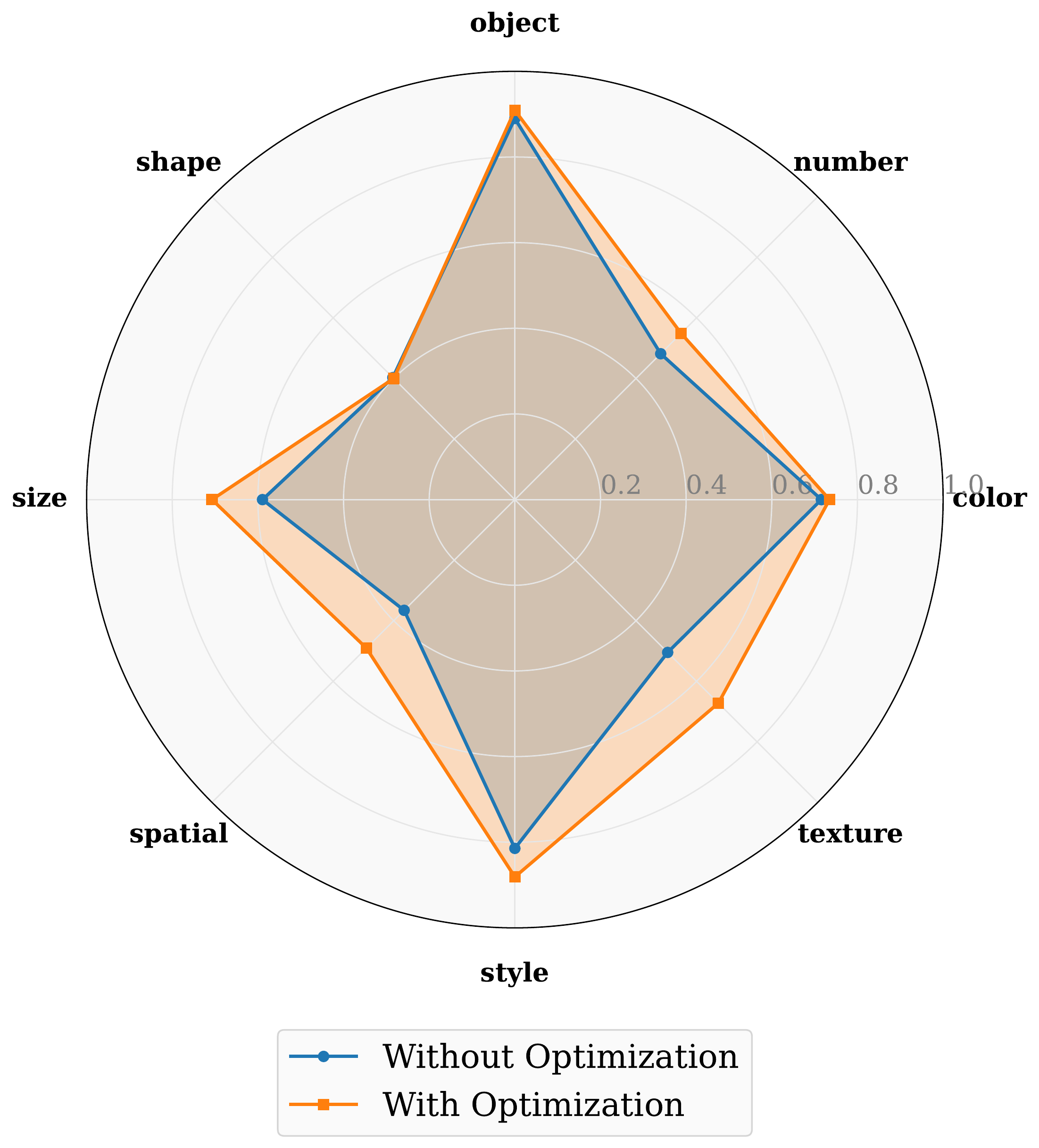}
        \caption{Playground v2.5}
        \label{fig:playground_radar_k3}
    \end{subfigure}
    \caption{Radar graphs showing category improvements for complexity level $k=3$ across all three models.}
    \label{fig:radar_k3}
\end{figure*}

\begin{figure*}[htbp]
    \centering
    \begin{subfigure}[b]{0.3\textwidth}
        \centering
        \includegraphics[width=\textwidth]{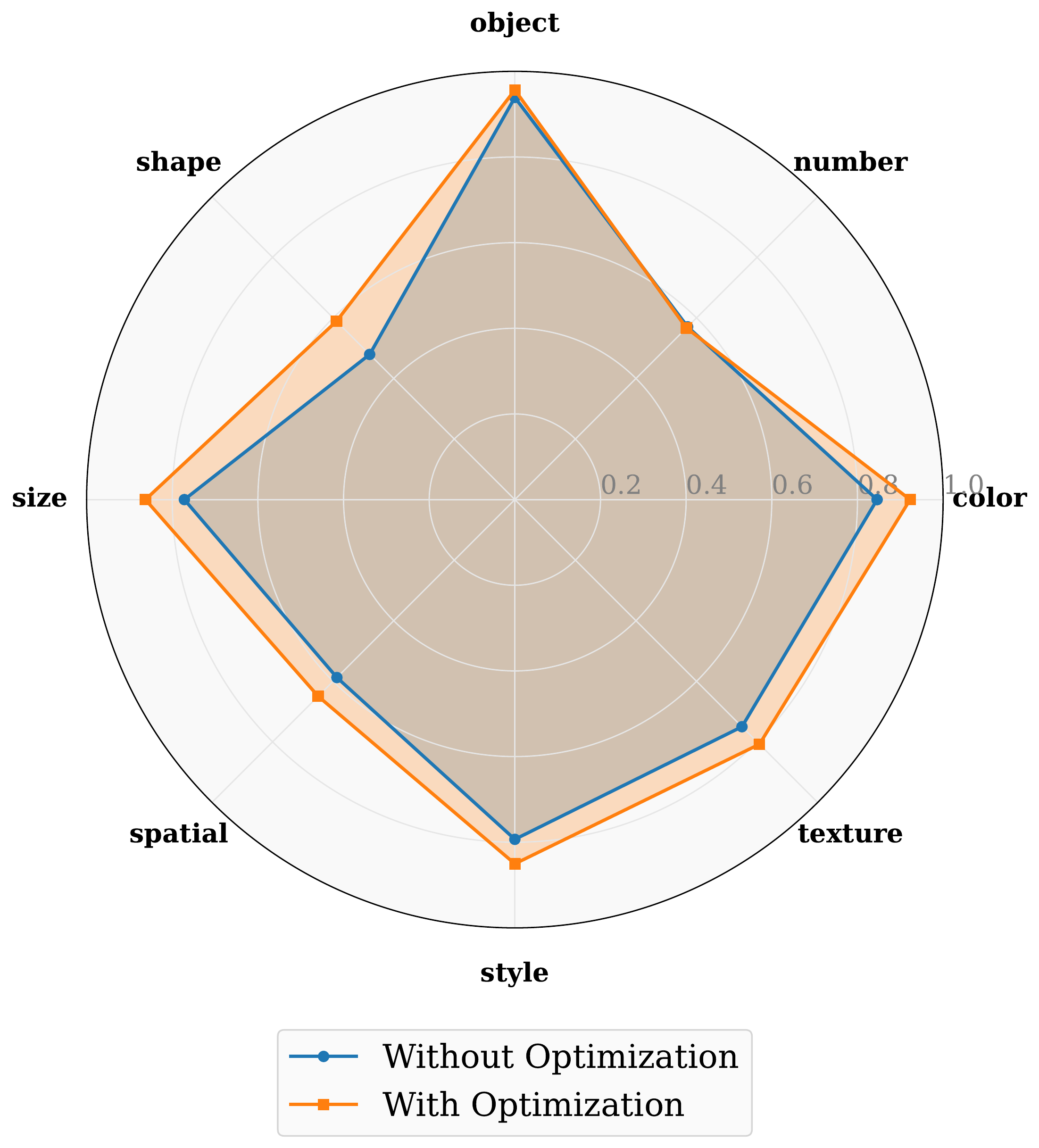}
        \caption{DALL·E 3}
        \label{fig:dalle3_radar_k4}
    \end{subfigure}
    \hfill
     \begin{subfigure}[b]{0.3\textwidth}
        \centering
        \includegraphics[width=\textwidth]{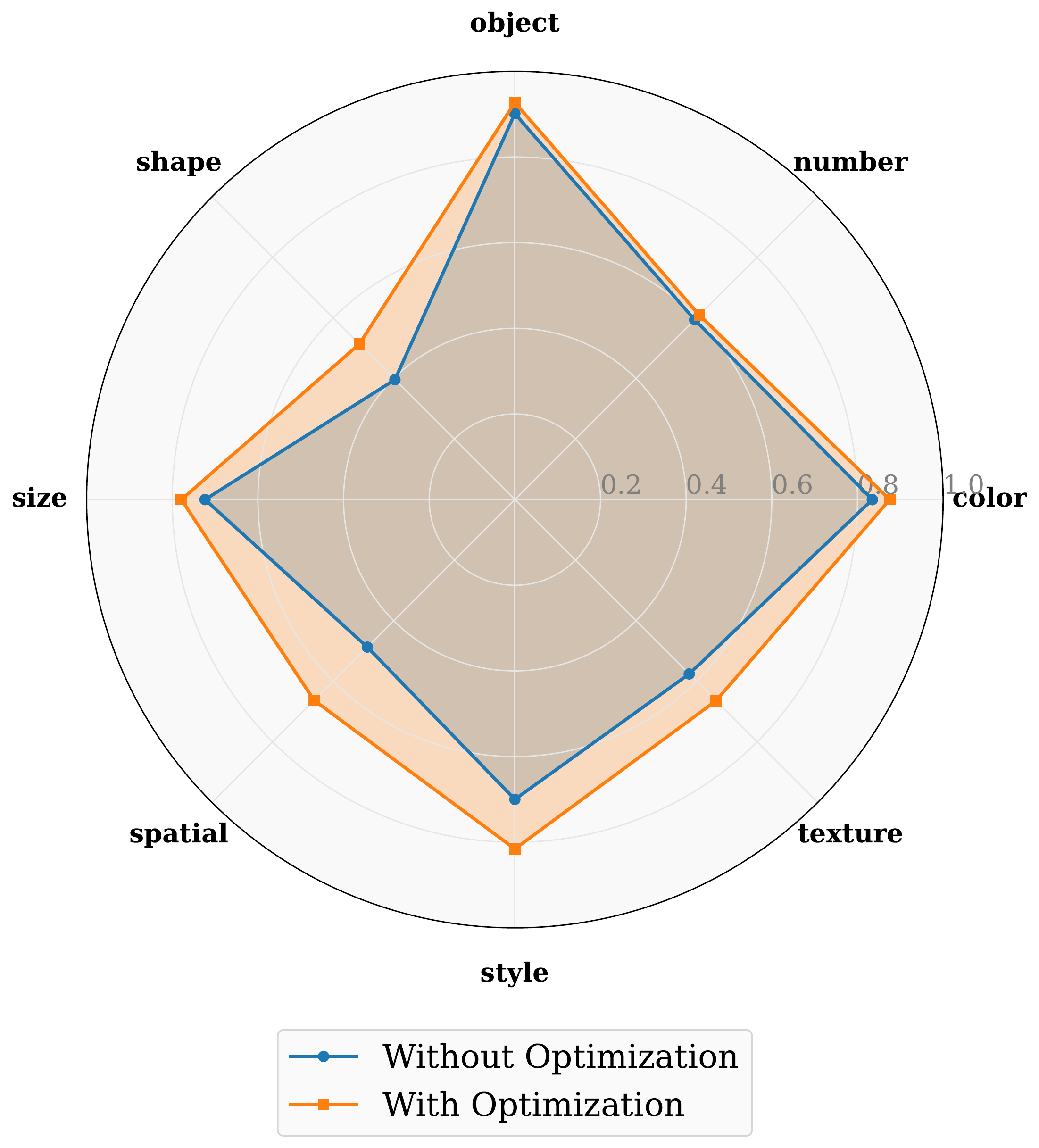}
        \caption{Stable Diffusion 3.5}
        \label{fig:sd35_radar_k4}
    \end{subfigure}
    \hfill
    \begin{subfigure}[b]{0.3\textwidth}
        \centering
        \includegraphics[width=\textwidth]{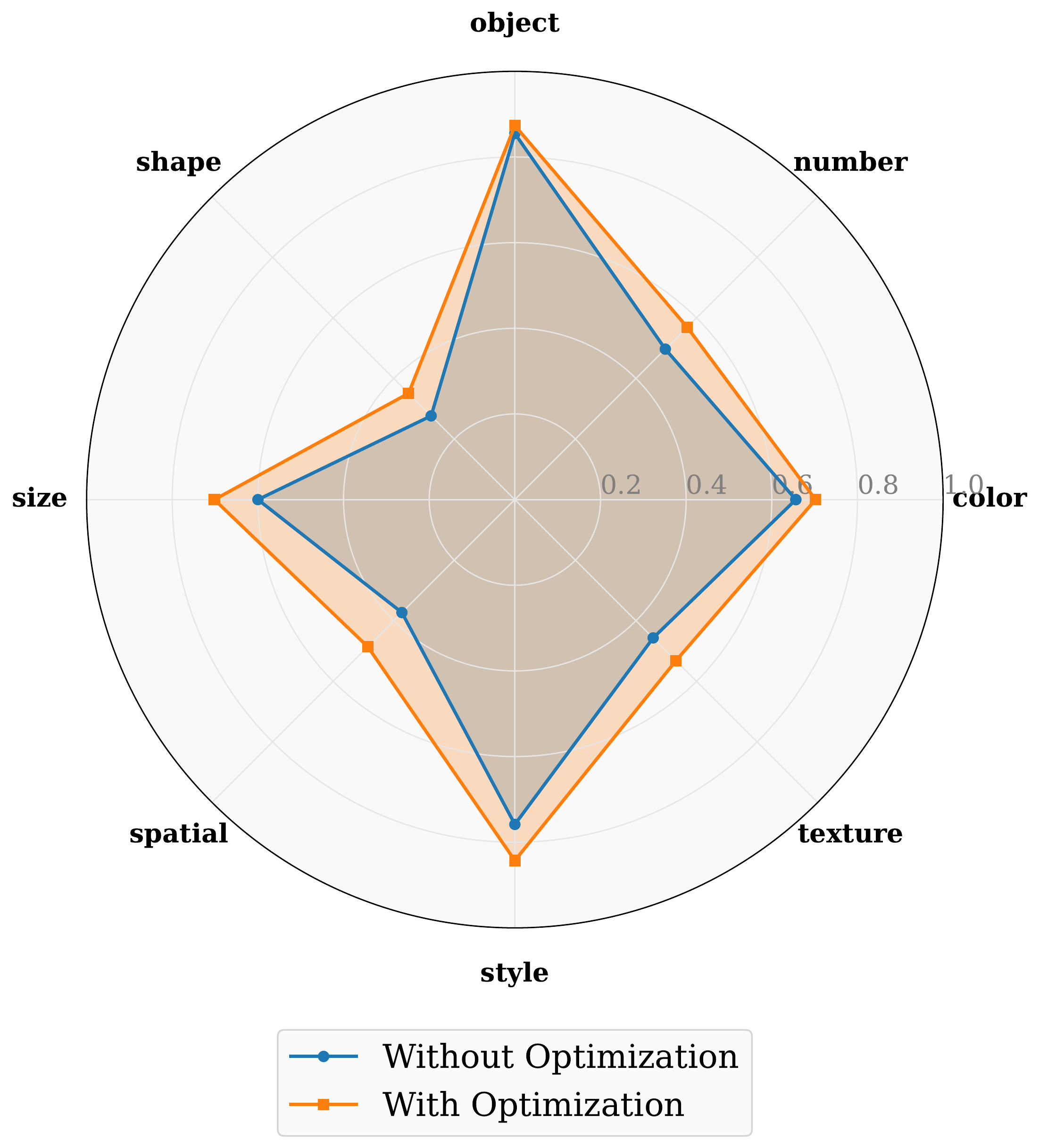}
        \caption{Playground v2.5}
        \label{fig:playground_radar_k4}
    \end{subfigure}
    \caption{Radar graphs showing category improvements for complexity level $k=4$ across all three models.}
    \label{fig:radar_k4}
\end{figure*}

\begin{figure*}[htbp]
    \centering
    \begin{subfigure}[b]{0.3\textwidth}
        \centering
        \includegraphics[width=\textwidth]{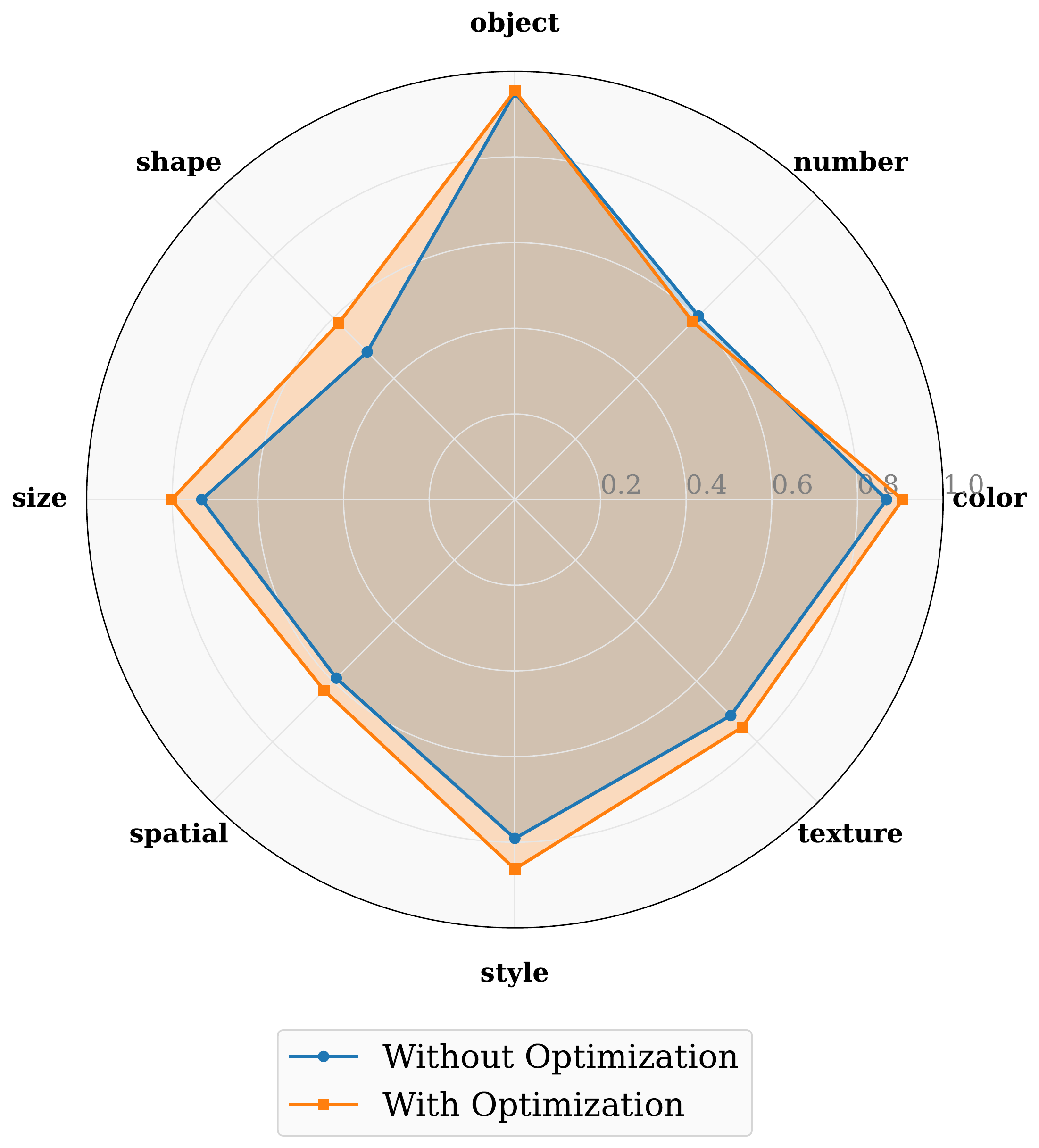}
        \caption{DALL·E 3}
        \label{fig:dalle3_radar_k5}
    \end{subfigure}
    \hfill
     \begin{subfigure}[b]{0.3\textwidth}
        \centering
        \includegraphics[width=\textwidth]{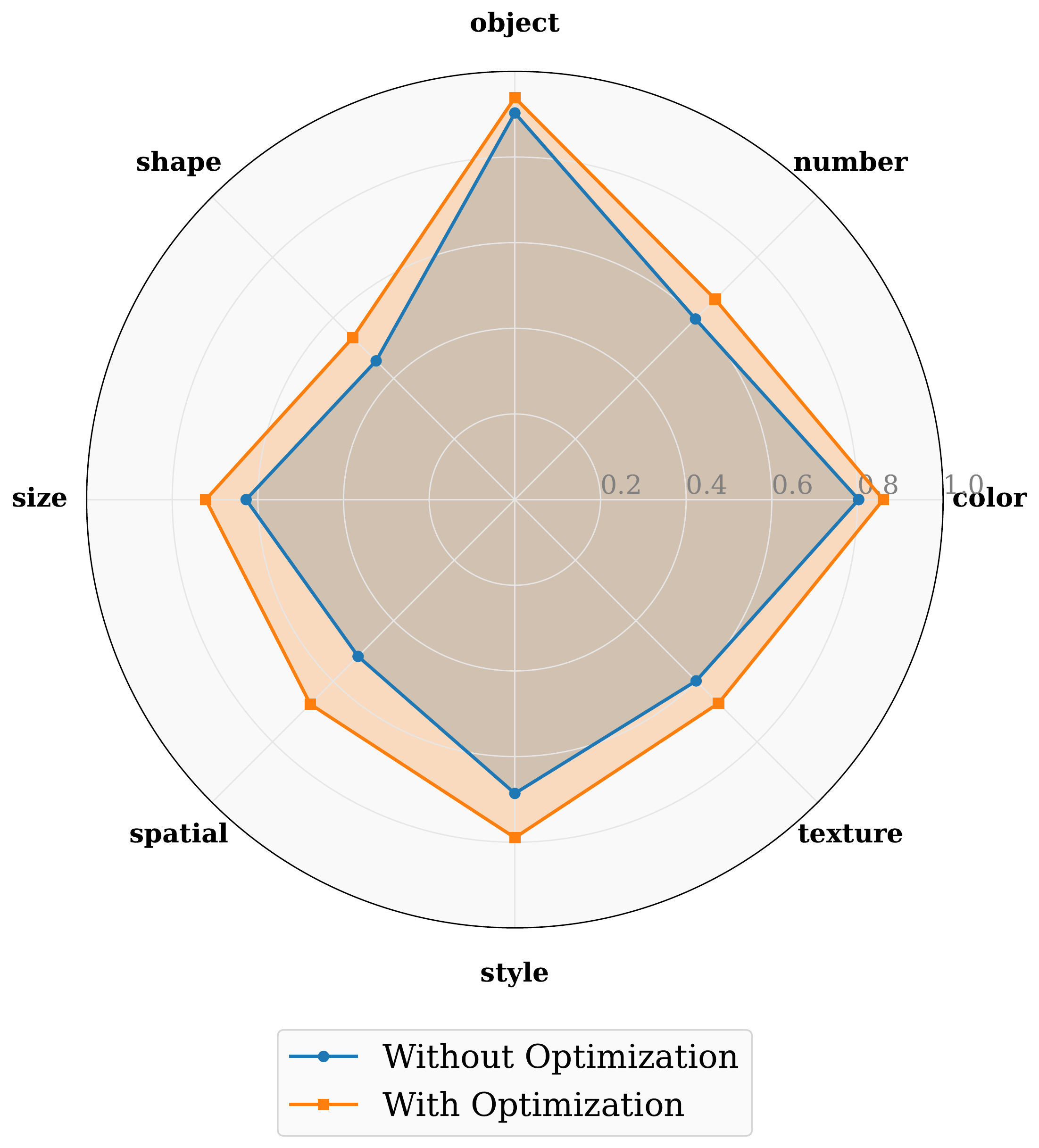}
        \caption{Stable Diffusion 3.5}
        \label{fig:sd35_radar_k5}
    \end{subfigure}
    \hfill
    \begin{subfigure}[b]{0.3\textwidth}
        \centering
        \includegraphics[width=\textwidth]{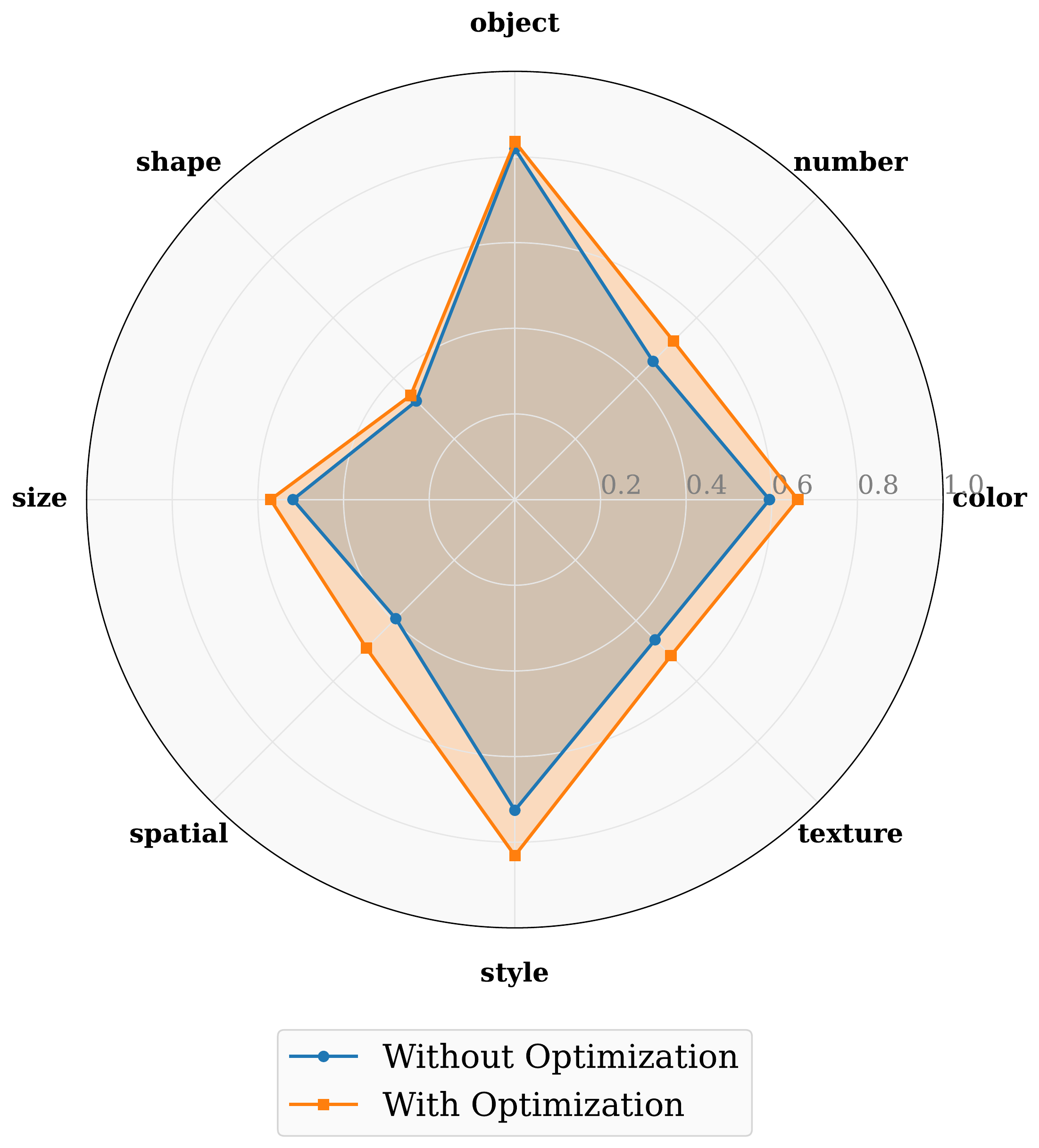}
        \caption{Playground v2.5}
        \label{fig:playground_radar_k5}
    \end{subfigure}
    \caption{Radar graphs showing category improvements for complexity level $k=5$ across all three models.}
    \label{fig:radar_k5}
\end{figure*}

\begin{figure*}[htbp]
    \centering
    \begin{subfigure}[b]{0.3\textwidth}
        \centering
        \includegraphics[width=\textwidth]{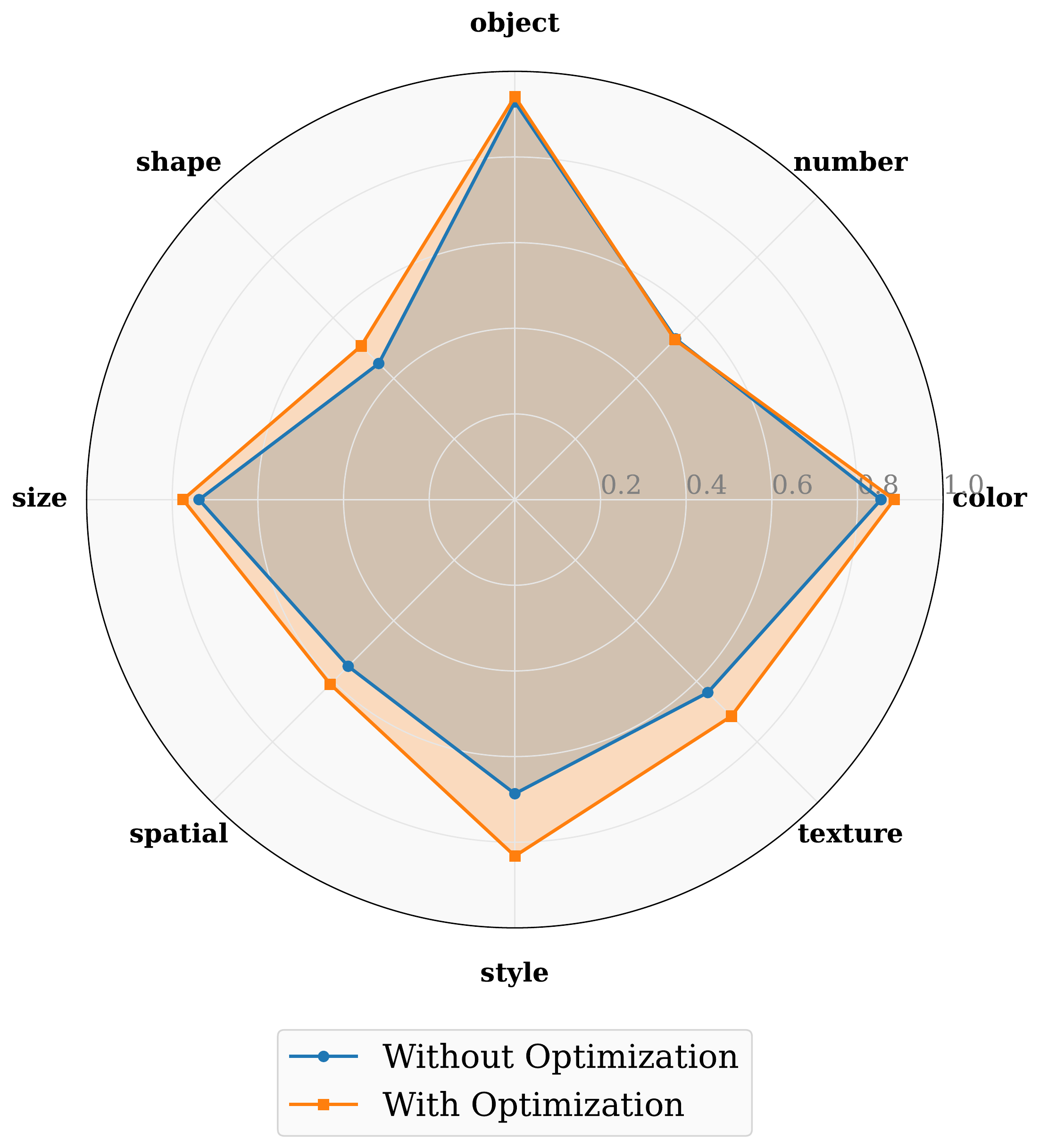}
        \caption{DALL·E 3}
        \label{fig:dalle3_radar_k6}
    \end{subfigure}
    \hfill
     \begin{subfigure}[b]{0.3\textwidth}
        \centering
        \includegraphics[width=\textwidth]{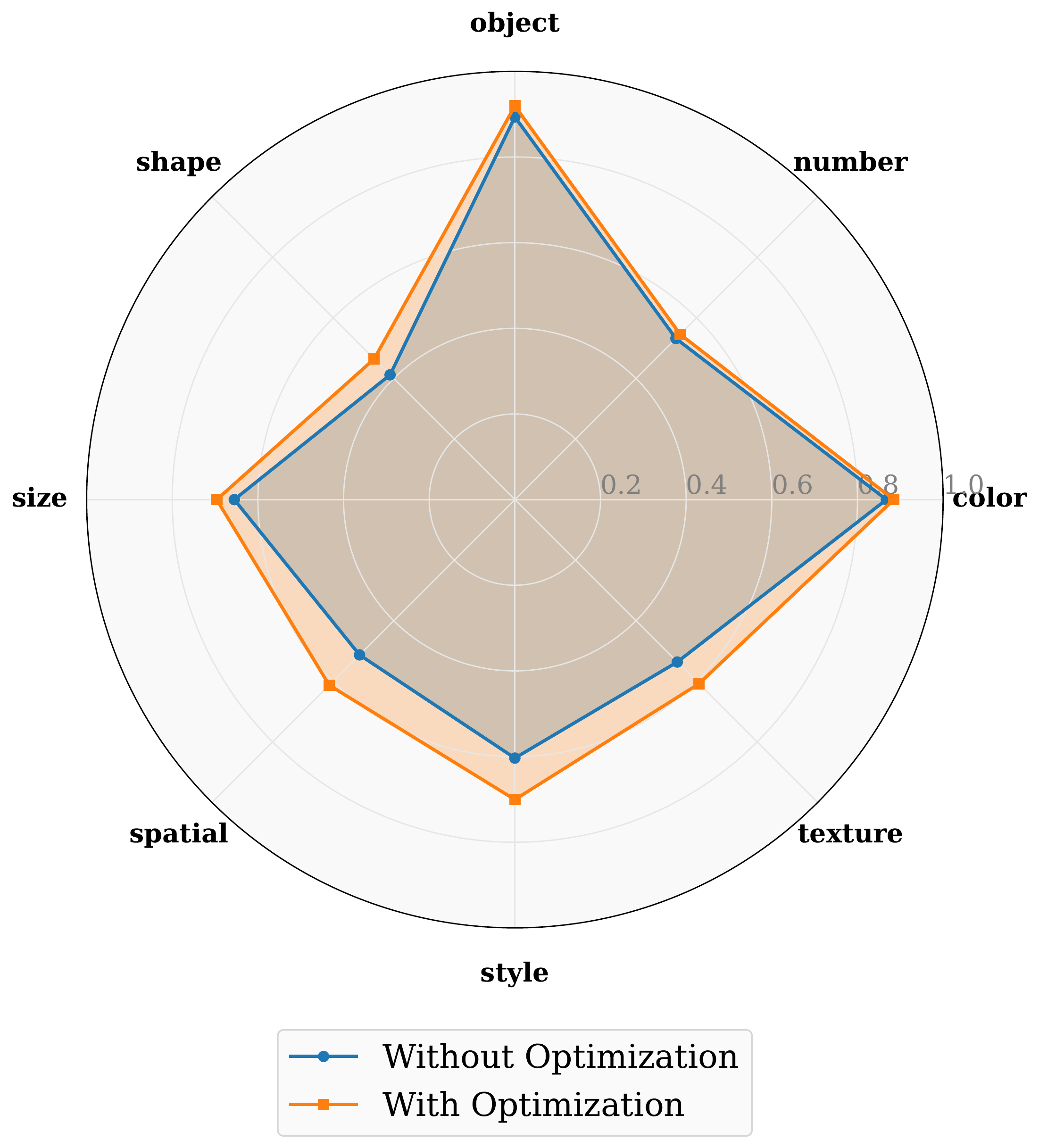}
        \caption{Stable Diffusion 3.5}
        \label{fig:sd35_radar_k6}
    \end{subfigure}
    \hfill
    \begin{subfigure}[b]{0.3\textwidth}
        \centering
        \includegraphics[width=\textwidth]{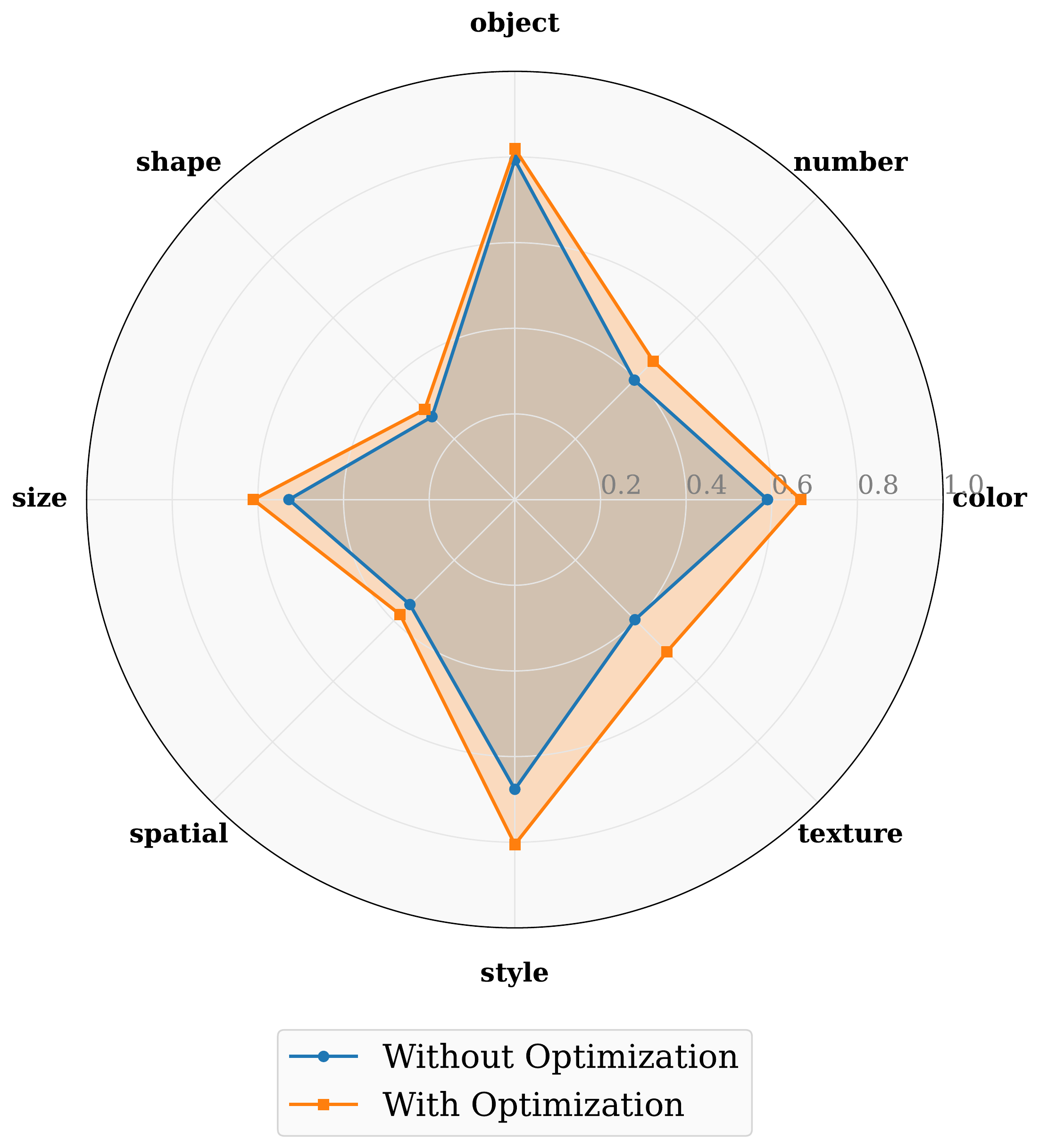}
        \caption{Playground v2.5}
        \label{fig:playground_radar_k6}
    \end{subfigure}
    \caption{Radar graphs showing category improvements for complexity level $k=6$ across all three models.}
    \label{fig:radar_k6}
\end{figure*}

\begin{figure*}[htbp]
    \centering
    \begin{subfigure}[b]{0.3\textwidth}
        \centering
        \includegraphics[width=\textwidth]{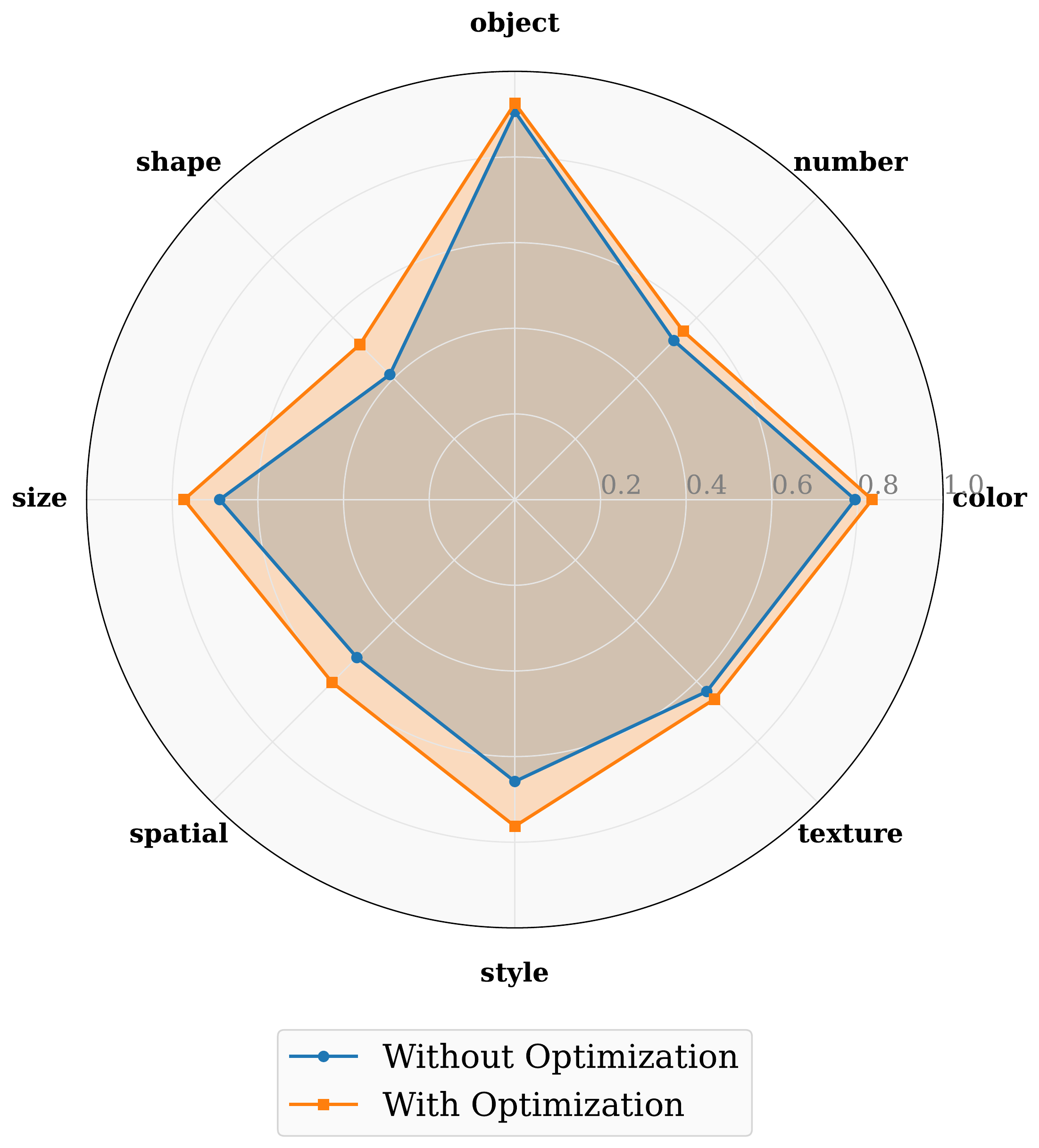}
        \caption{DALL·E 3}
        \label{fig:dalle3_radar_k7}
    \end{subfigure}
    \hfill
     \begin{subfigure}[b]{0.3\textwidth}
        \centering
        \includegraphics[width=\textwidth]{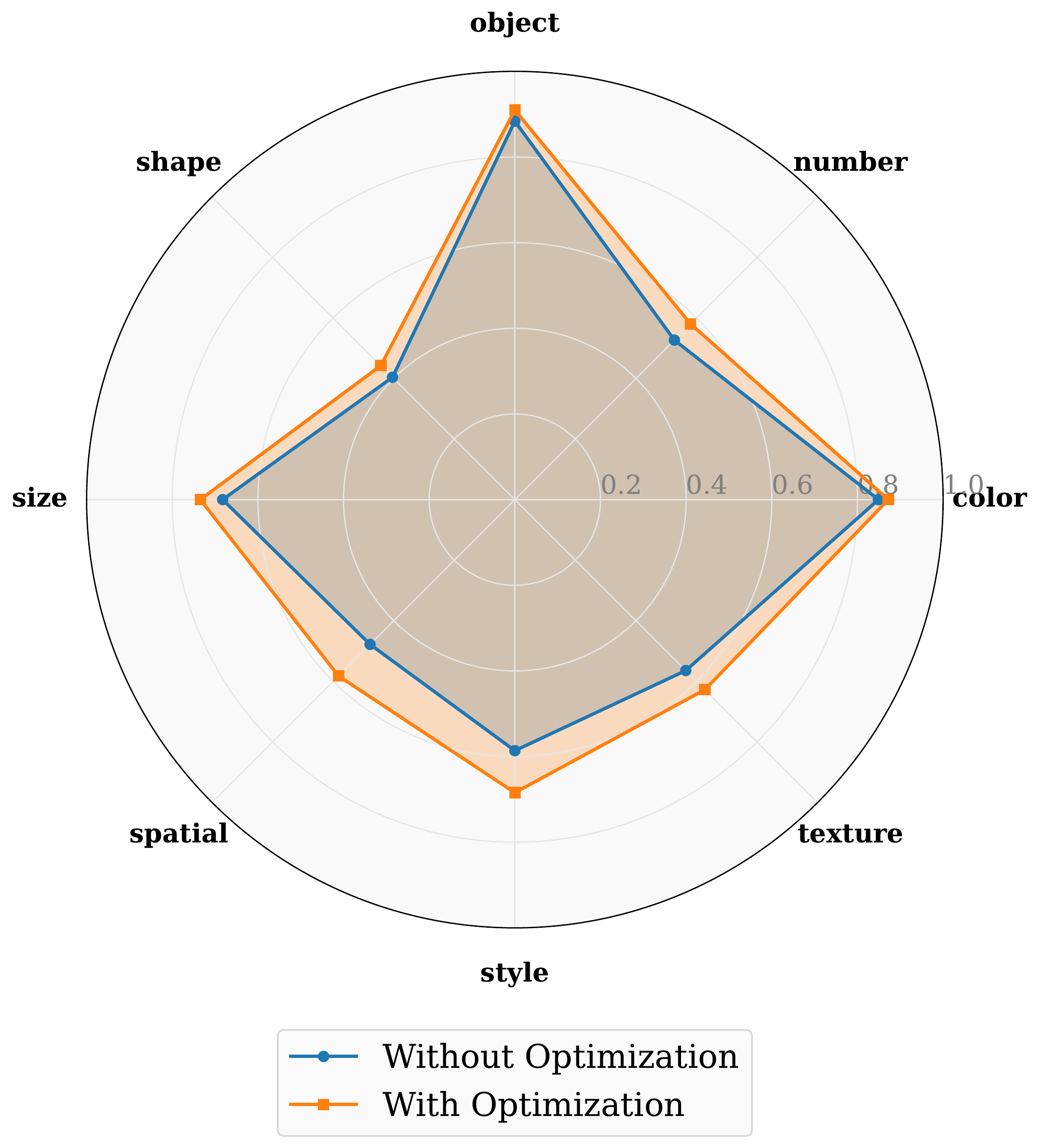}
        \caption{Stable Diffusion 3.5}
        \label{fig:sd35_radar_k7}
    \end{subfigure}
    \hfill
    \begin{subfigure}[b]{0.3\textwidth}
        \centering
        \includegraphics[width=\textwidth]{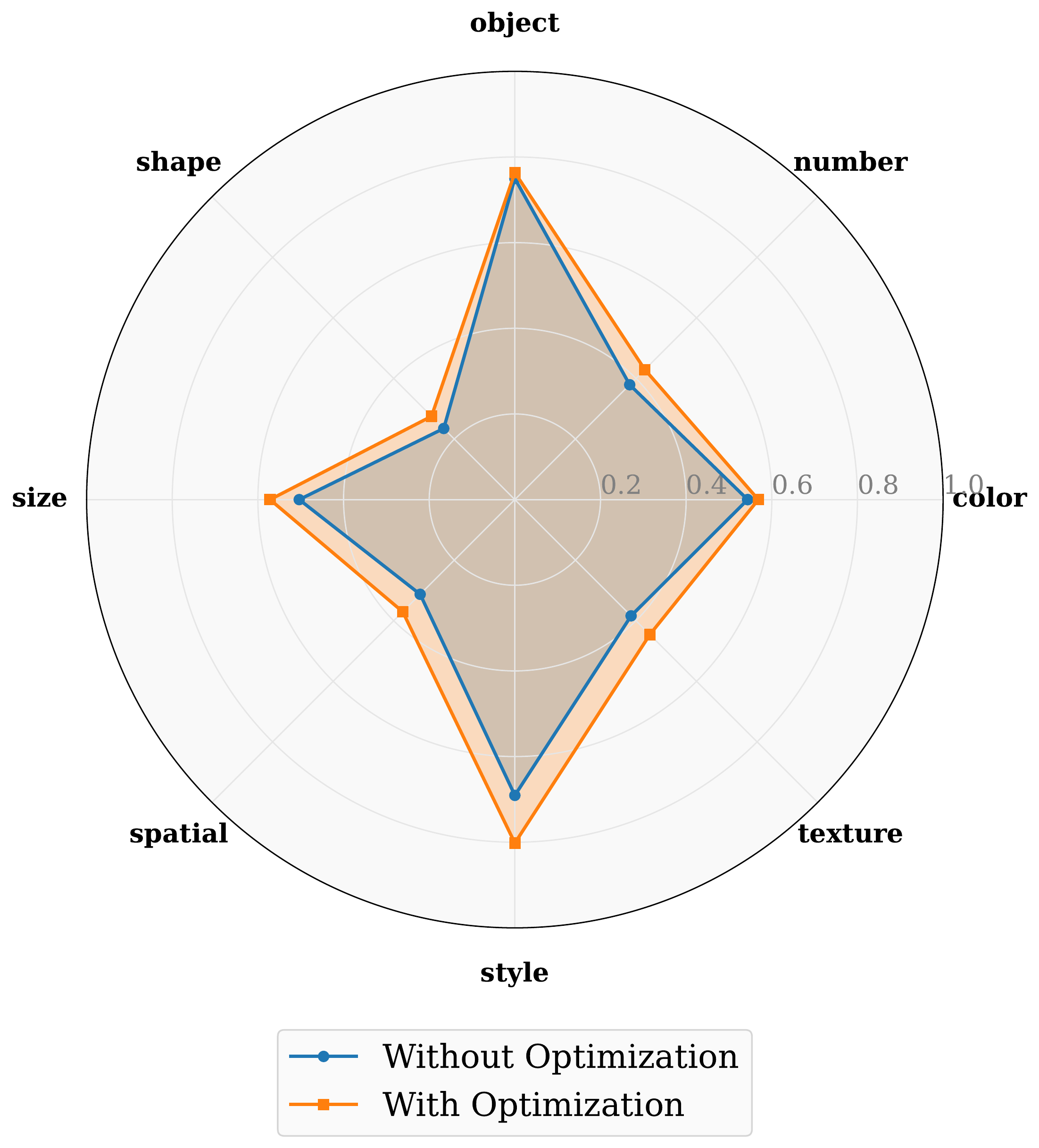}
        \caption{Playground v2.5}
        \label{fig:playground_radar_k7}
    \end{subfigure}
    \caption{Radar graphs showing category improvements for complexity level $k=7$ across all three models.}
    \label{fig:radar_k7}
\end{figure*}

Figures~\ref{fig:radar_k1} through \ref{fig:radar_k7} present detailed radar graphs comparing the three models at each complexity level, supplementing the category-wise heatmaps presented in the main paper.

\section{Complete Cross-model Transferability Results}
\label{sec:transferability}

\begin{figure*}[t]
    \centering
    \begin{subfigure}[b]{0.48\textwidth}
        \centering
        \includegraphics[width=\textwidth]{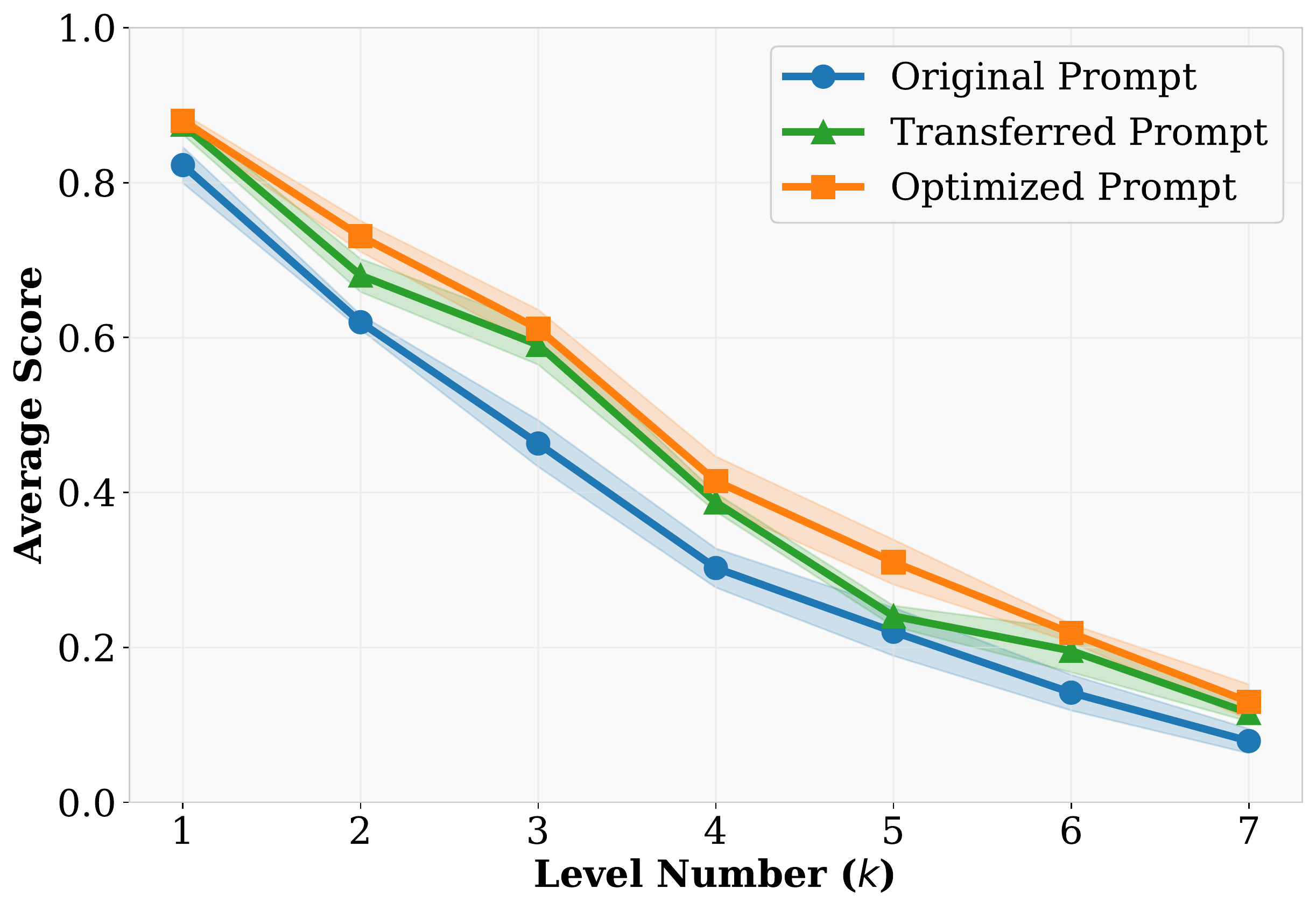}
        \caption{Average Score}
        \label{fig:dalle3_sd_avg}
    \end{subfigure}
    \hfill
    \begin{subfigure}[b]{0.48\textwidth}
        \centering
        \includegraphics[width=\textwidth]{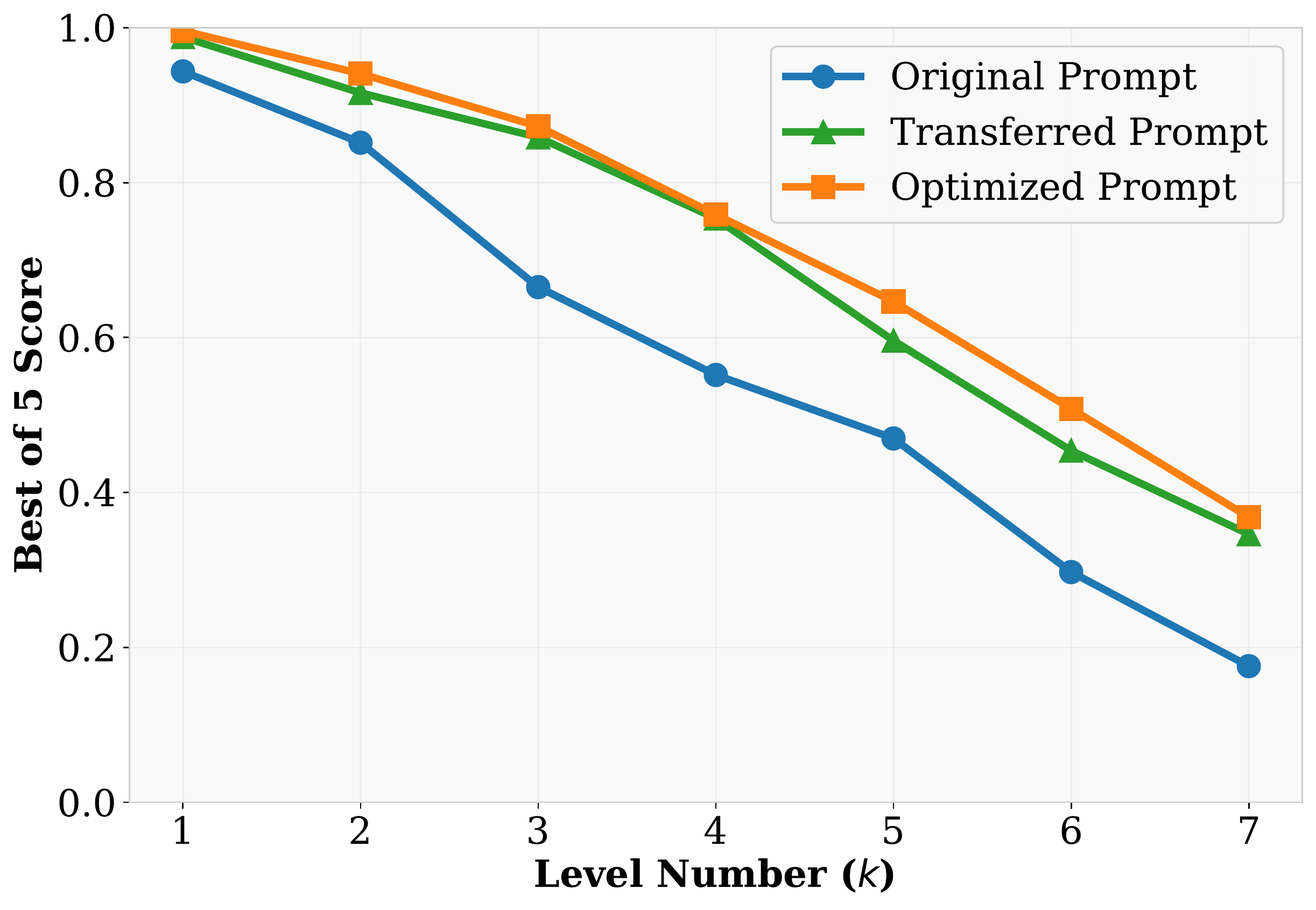}
        \caption{Best-of-5 Score}
        \label{fig:dalle3_sd_best}
    \end{subfigure}
    \caption{DALL·E 3 performance with SD 3.5 optimized prompts compared to original prompts and self-optimized prompts.}
    \label{fig:dalle3_sd}
\end{figure*}

\begin{figure*}[t]
    \centering
    \begin{subfigure}[b]{0.48\textwidth}
        \centering
        \includegraphics[width=\textwidth]{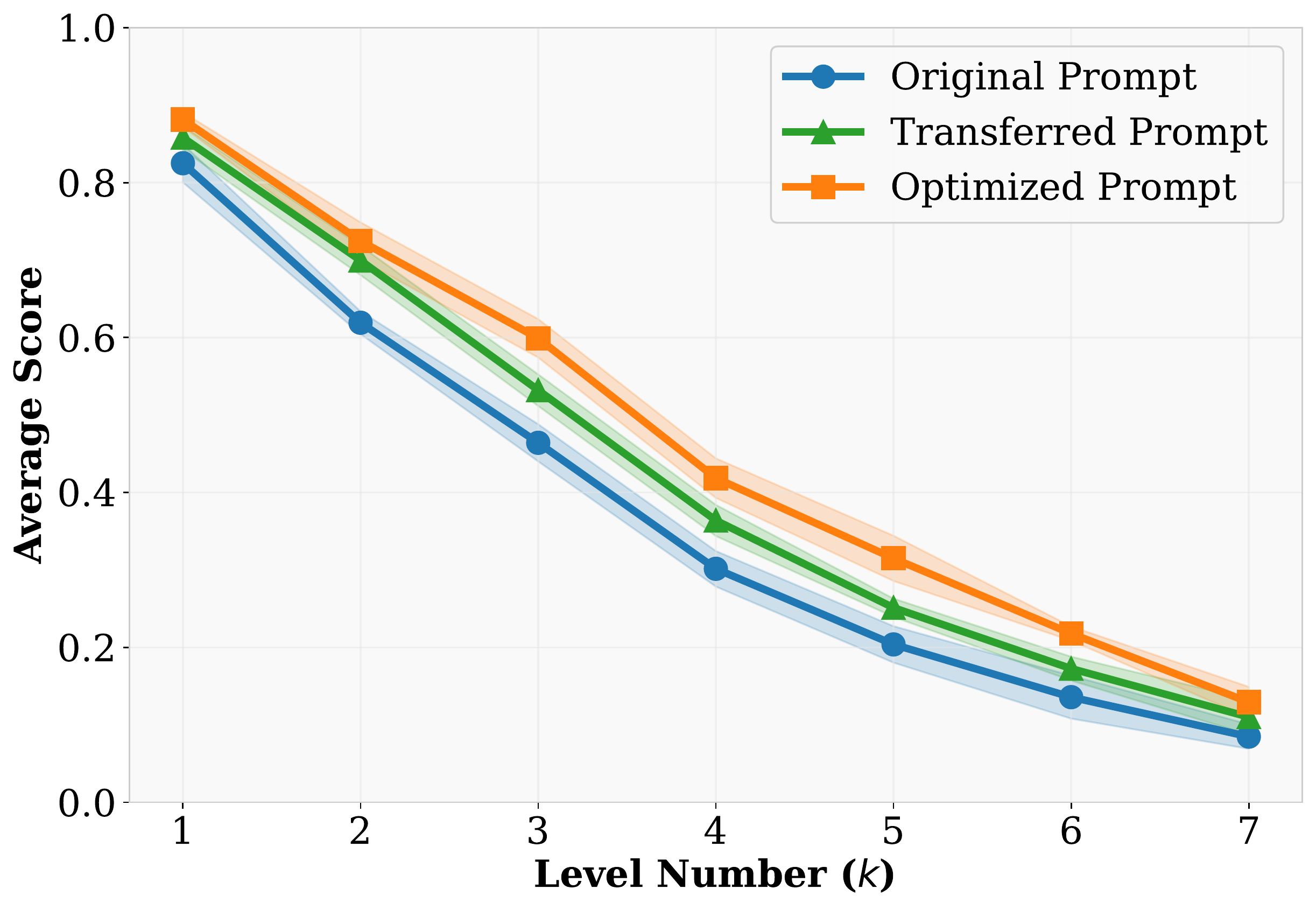}
        \caption{Average Score}
        \label{fig:dalle3_pg_avg}
    \end{subfigure}
    \hfill
    \begin{subfigure}[b]{0.48\textwidth}
        \centering
        \includegraphics[width=\textwidth]{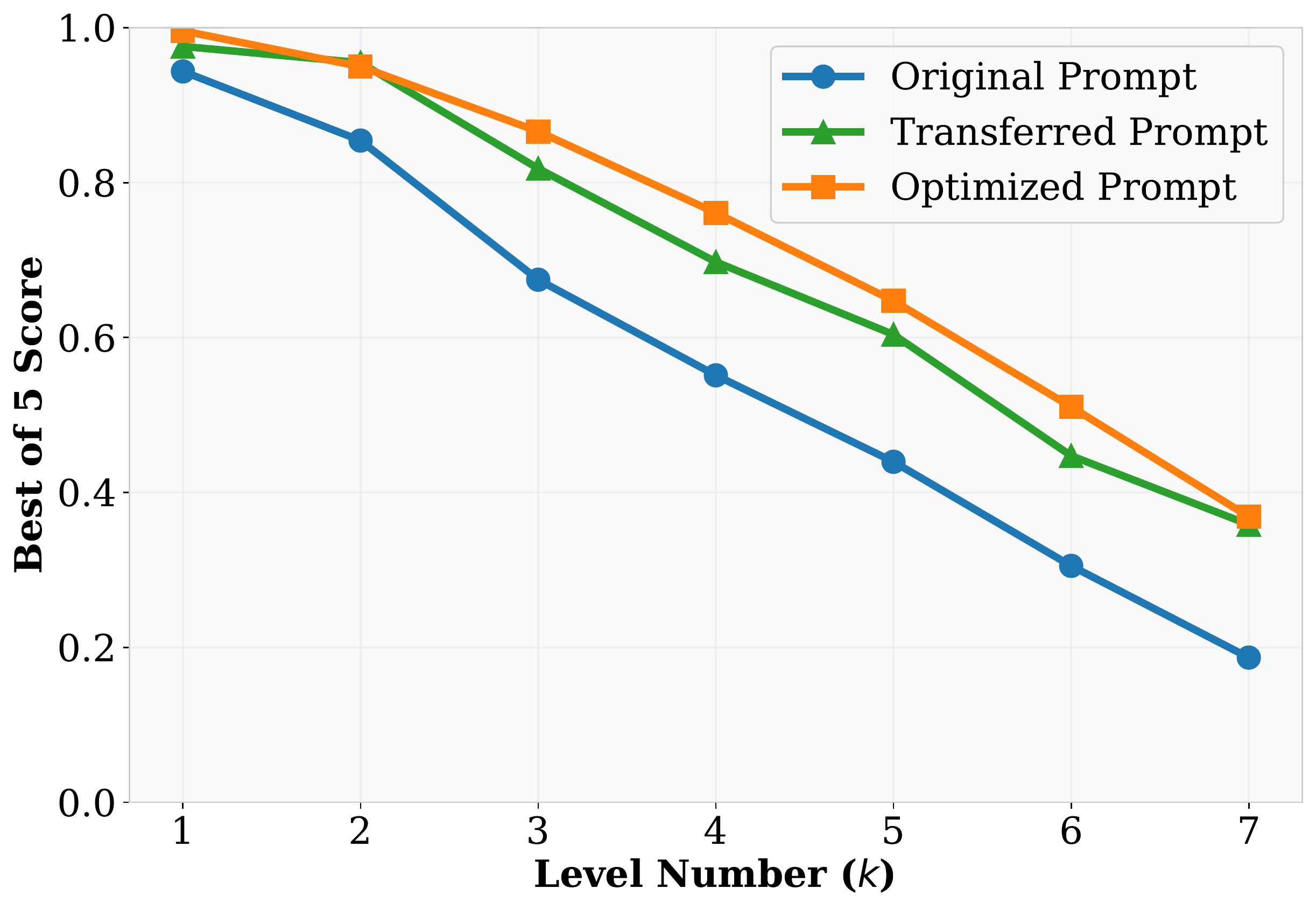}
        \caption{Best-of-5 Score}
        \label{fig:dalle3_pg_best}
    \end{subfigure}
    \caption{DALL·E 3 performance with Playground v2.5 optimized prompts compared to original prompts and self-optimized prompts.}
    \label{fig:dalle3_pg}
\end{figure*}

\begin{figure*}[t]
    \centering
    \begin{subfigure}[b]{0.48\textwidth}
        \centering
        \includegraphics[width=\textwidth]{figures/dalle3_prompt_stable-diffusion/avg_score/score_comparison.pdf}
        \caption{Average Score}
        \label{fig:sd_dalle3_avg}
    \end{subfigure}
    \hfill
    \begin{subfigure}[b]{0.48\textwidth}
        \centering
        \includegraphics[width=\textwidth]{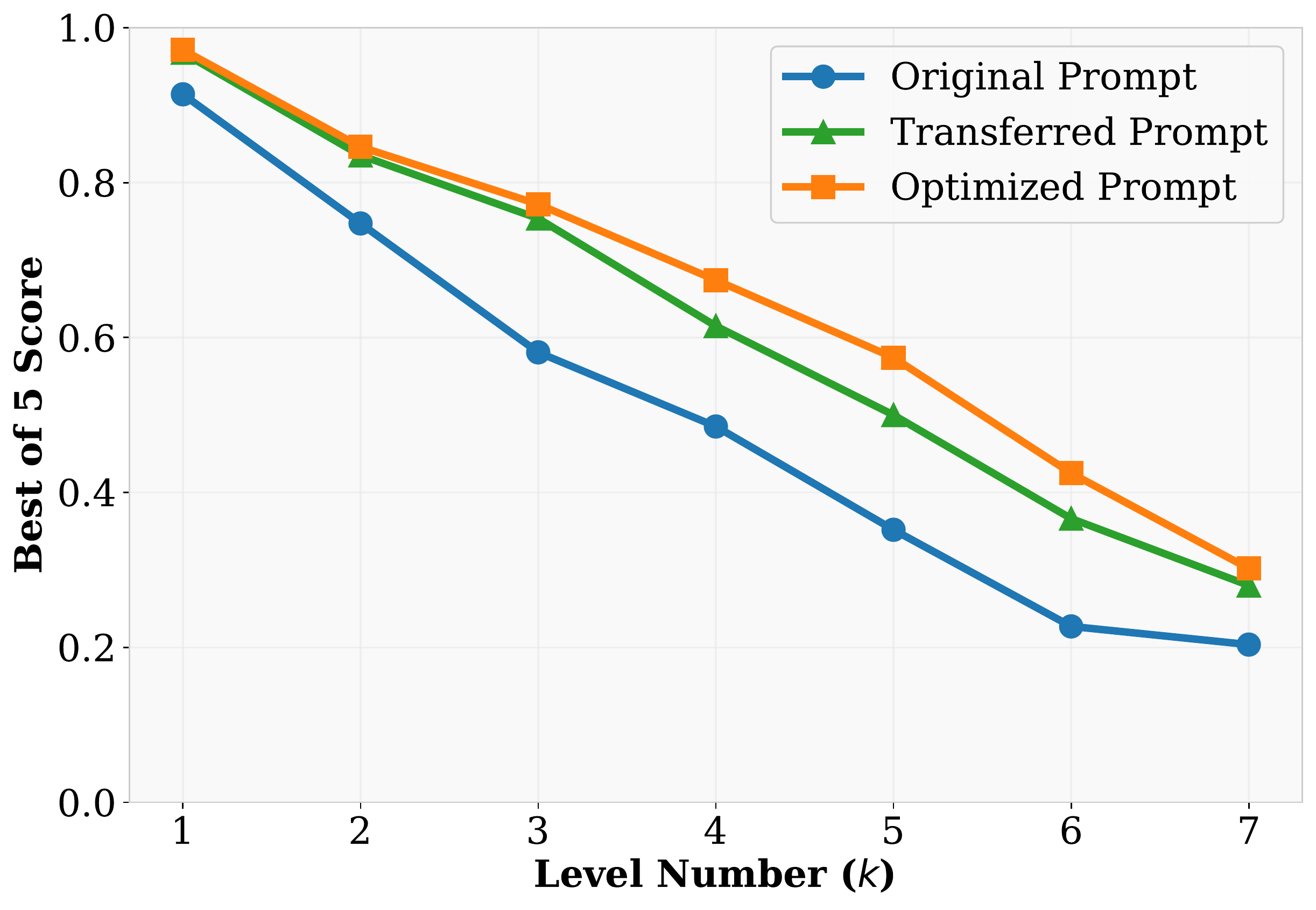}
        \caption{Best-of-5 Score}
        \label{fig:sd_dalle3_best}
    \end{subfigure}
    \caption{Stable Diffusion 3.5 performance with DALL·E 3 optimized prompts compared to original prompts and self-optimized prompts.}
    \label{fig:sd_dalle3}
\end{figure*}

\begin{figure*}[t]
    \centering
    \begin{subfigure}[b]{0.48\textwidth}
        \centering
        \includegraphics[width=\textwidth]{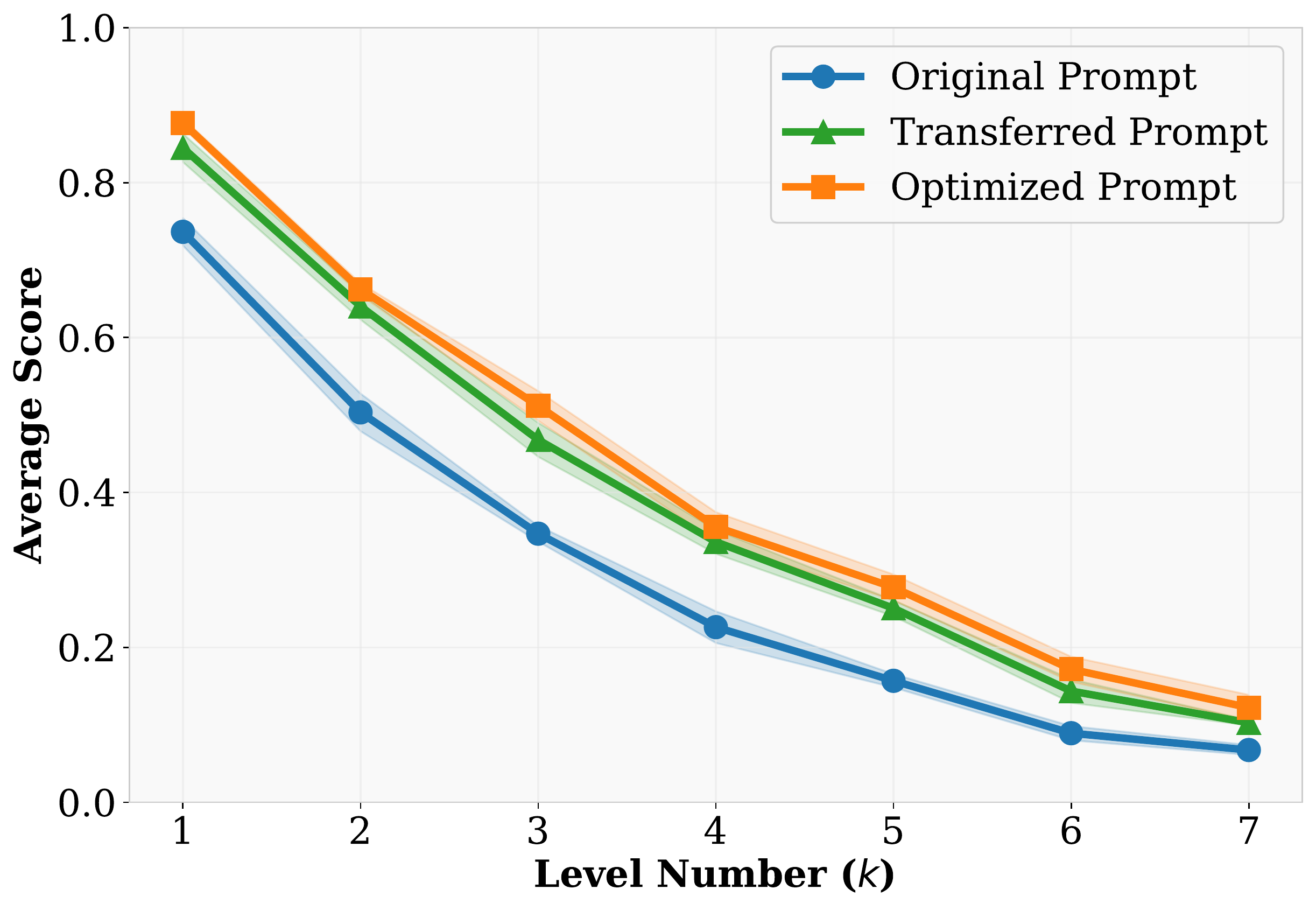}
        \caption{Average Score}
        \label{fig:sd_pg_avg}
    \end{subfigure}
    \hfill
    \begin{subfigure}[b]{0.48\textwidth}
        \centering
        \includegraphics[width=\textwidth]{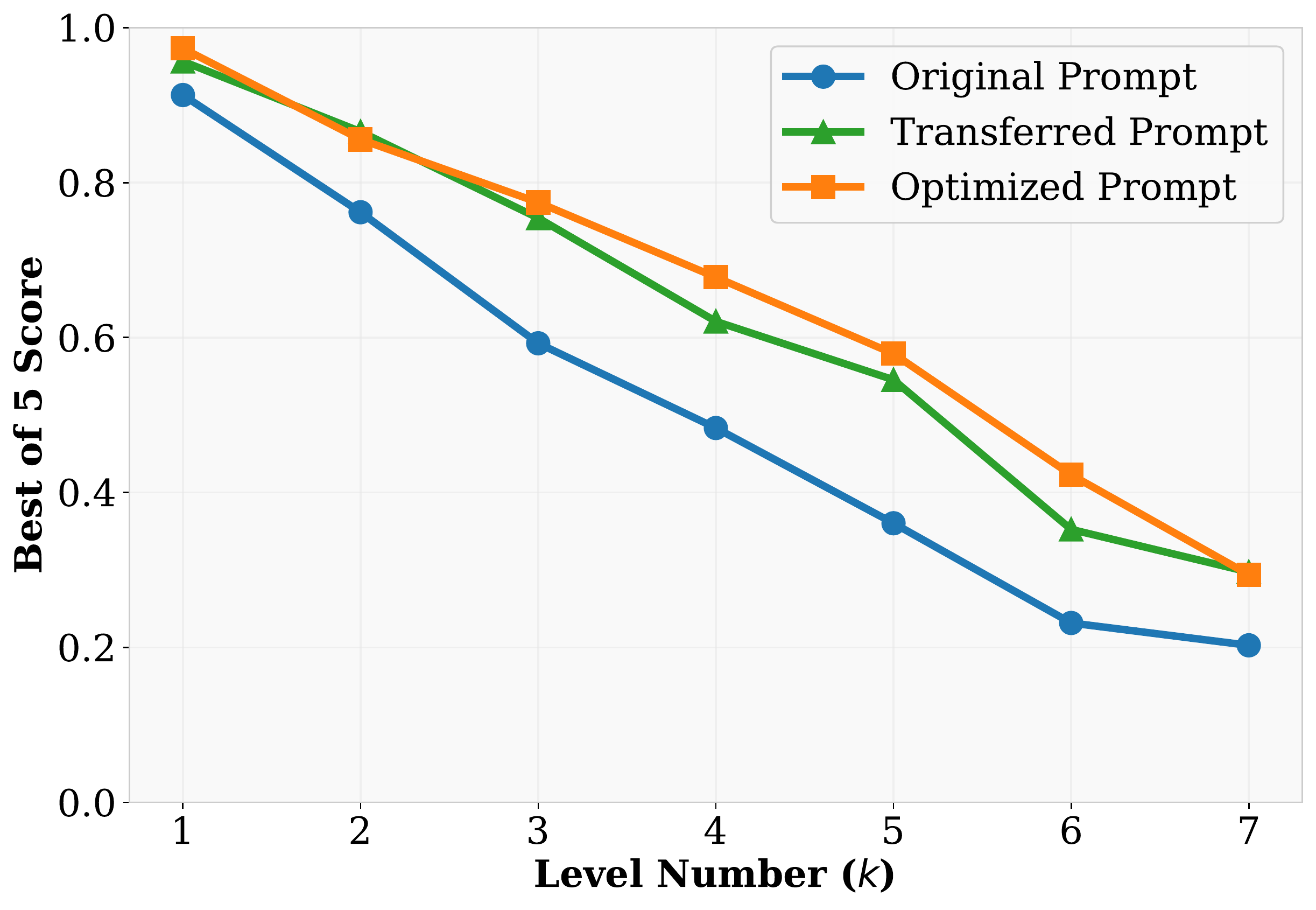}
        \caption{Best-of-5 Score}
        \label{fig:sd_pg_best}
    \end{subfigure}
    \caption{Stable Diffusion 3.5 performance with Playground v2.5 optimized prompts compared to original prompts and self-optimized prompts.}
    \label{fig:sd_pg}
\end{figure*}

\begin{figure*}[t]
    \centering
    \begin{subfigure}[b]{0.48\textwidth}
        \centering
        \includegraphics[width=\textwidth]{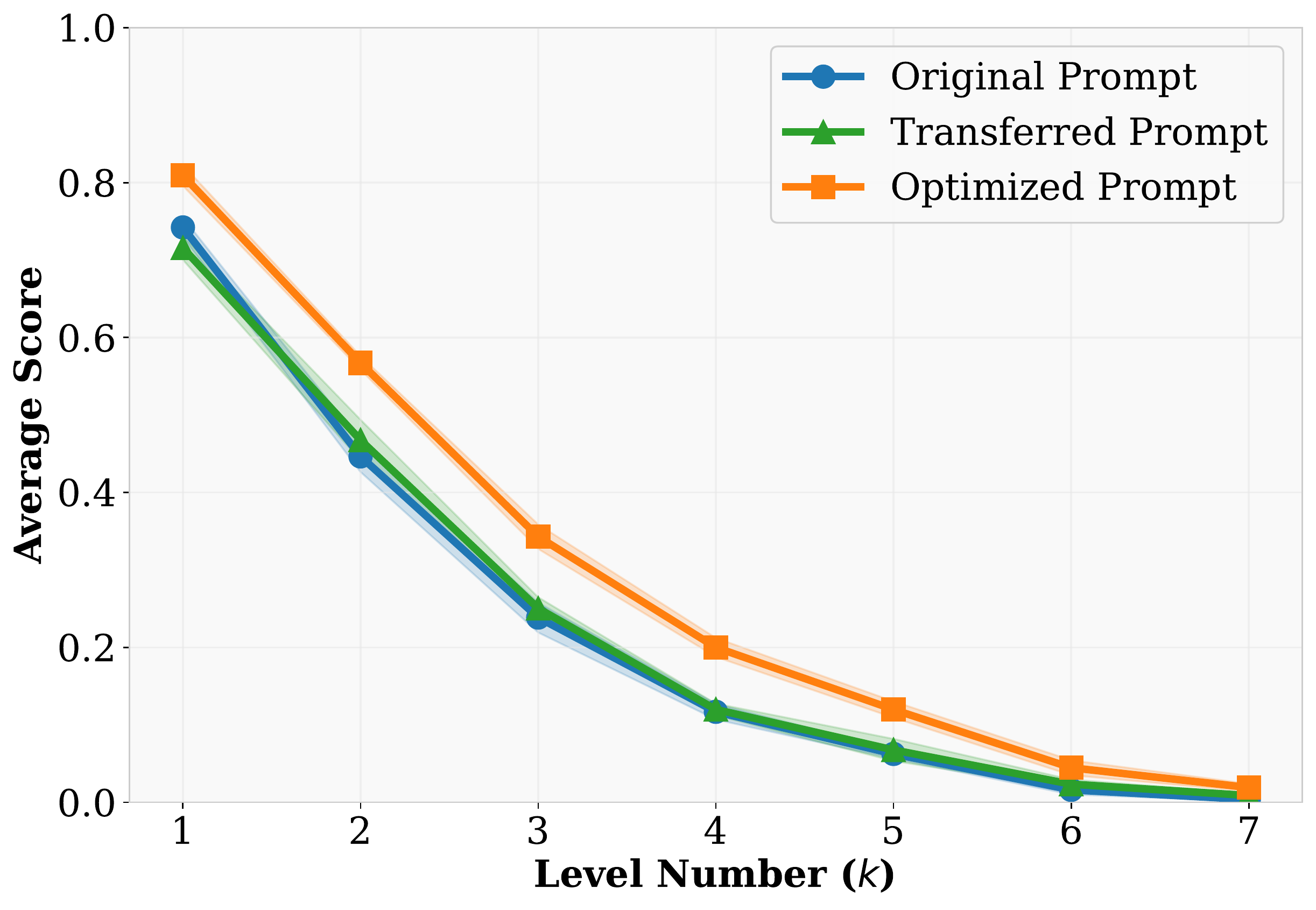}
        \caption{Average Score}
        \label{fig:pg_dalle3_avg}
    \end{subfigure}
    \hfill
    \begin{subfigure}[b]{0.48\textwidth}
        \centering
        \includegraphics[width=\textwidth]{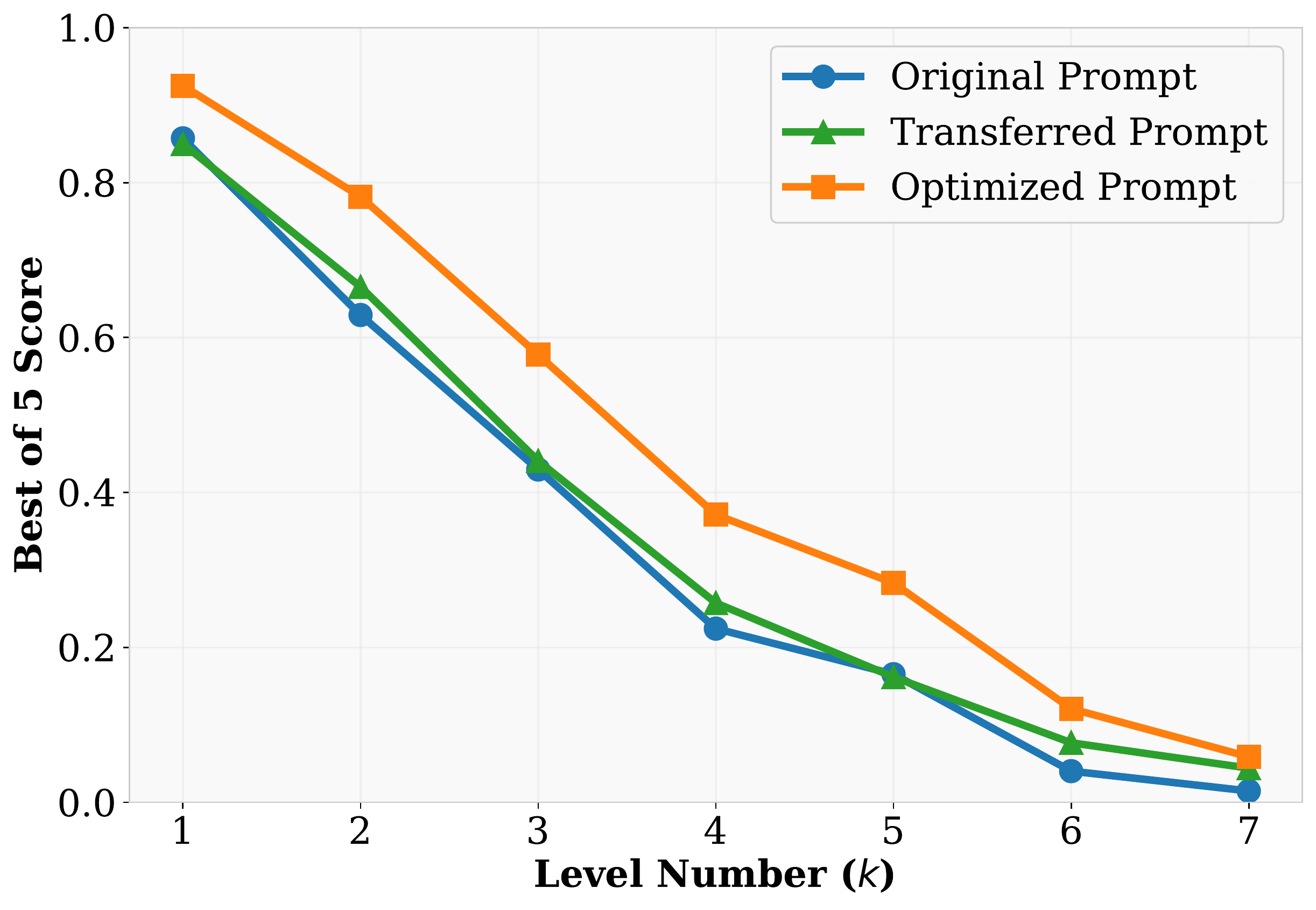}
        \caption{Best-of-5 Score}
        \label{fig:pg_dalle3_best}
    \end{subfigure}
    \caption{Playground v2.5 performance with DALL·E 3 optimized prompts compared to original prompts and self-optimized prompts.}
    \label{fig:pg_dalle3}
\end{figure*}

\begin{figure*}[t]
    \centering
    \begin{subfigure}[b]{0.48\textwidth}
        \centering
        \includegraphics[width=\textwidth]{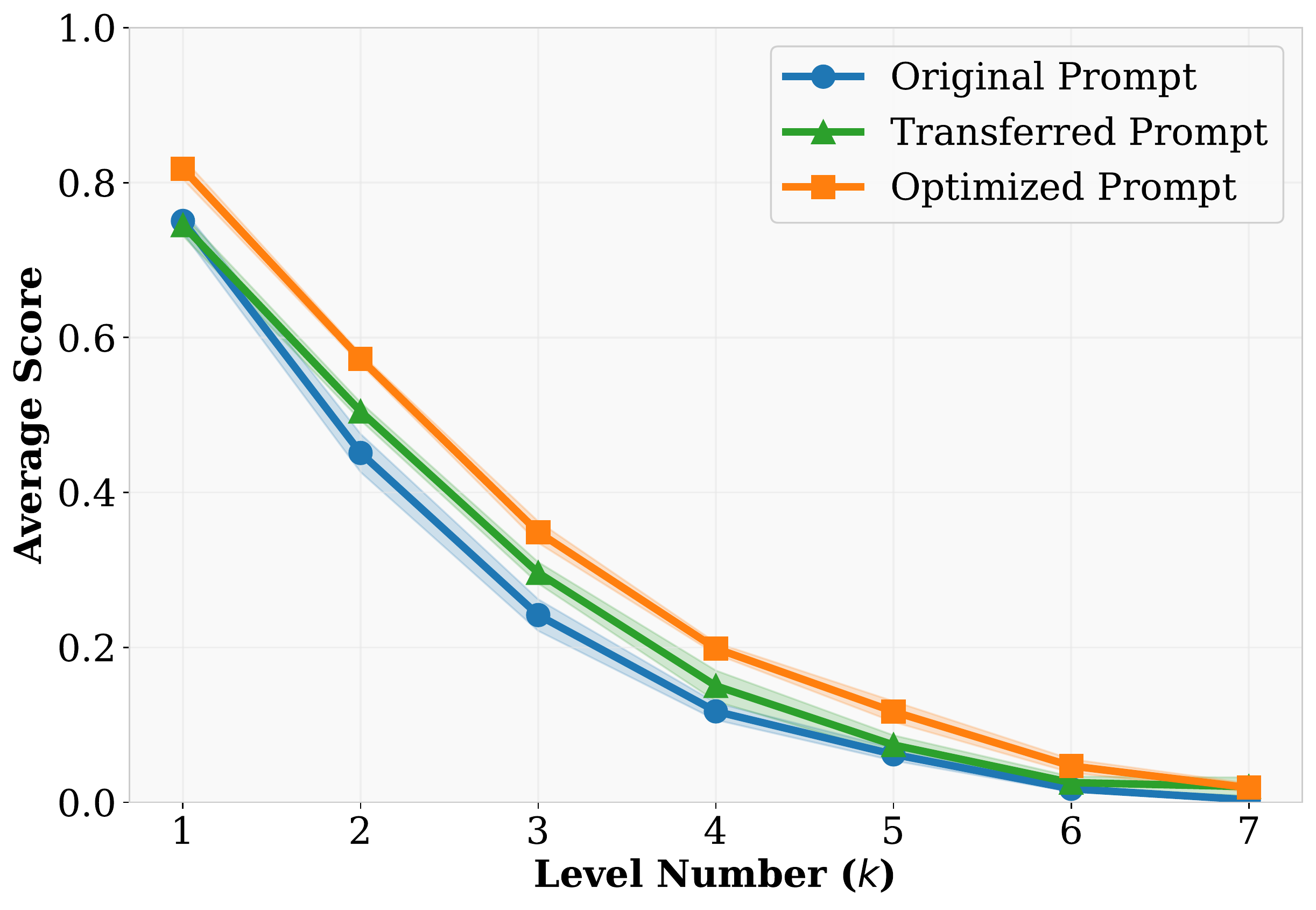}
        \caption{Average Score}
        \label{fig:pg_sd_avg}
    \end{subfigure}
    \hfill
    \begin{subfigure}[b]{0.48\textwidth}
        \centering
        \includegraphics[width=\textwidth]{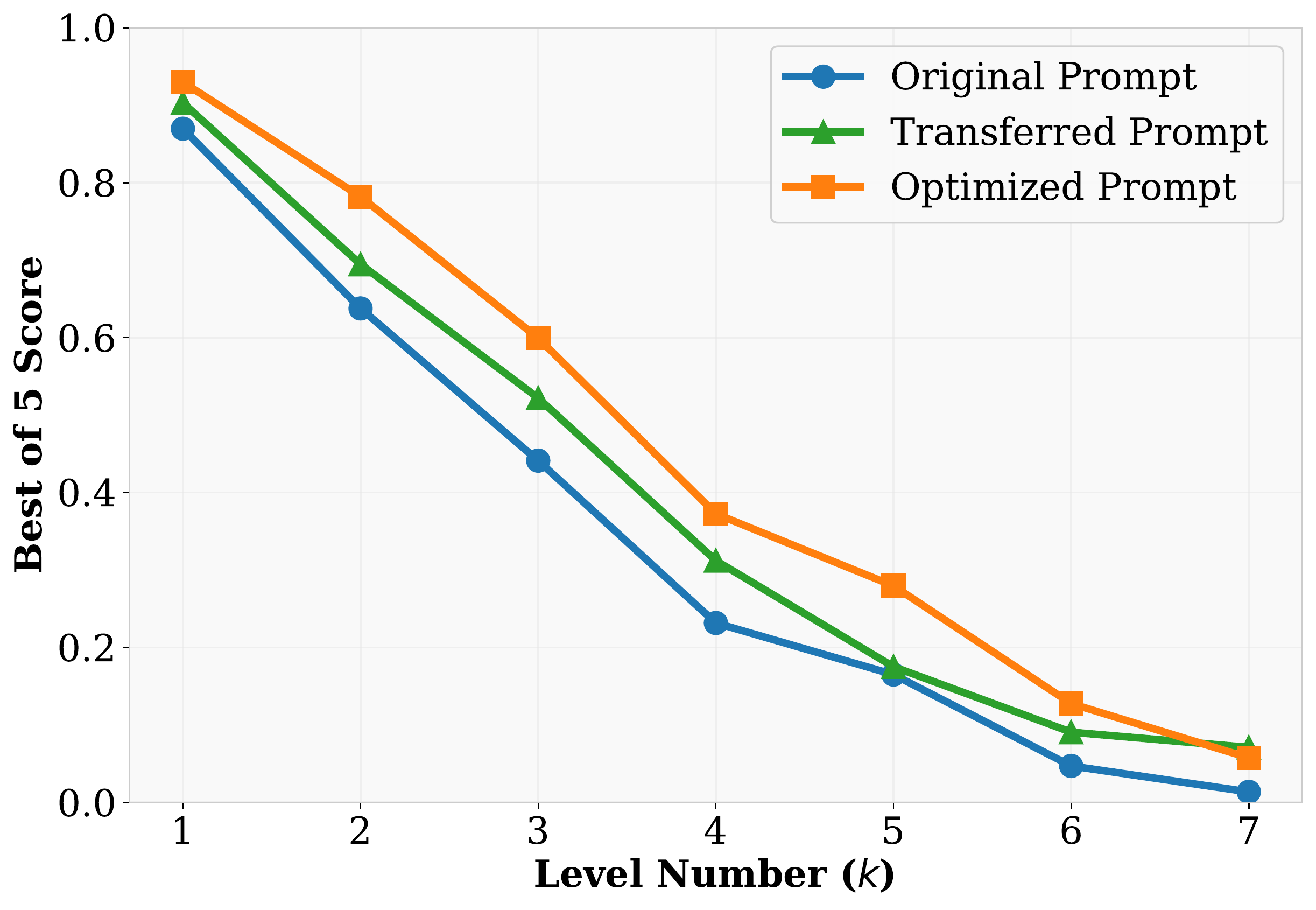}
        \caption{Best-of-5 Score}
        \label{fig:pg_sd_best}
    \end{subfigure}
    \caption{Playground v2.5 performance with Stable Diffusion 3.5 optimized prompts compared to original prompts and self-optimized prompts.}
    \label{fig:pg_sd}
\end{figure*}

This section provides comprehensive transferability results across all model pairs, extending the analysis presented in Section~\ref{sec:transferability} of the main paper. Figures~\ref{fig:dalle3_sd} through \ref{fig:pg_sd} present the complete set of transferability experiments across all model pairs, following the experimental setup described in the main paper.

The effectiveness of prompt transfer varies depending on the source and target models. Some key observations:

\begin{itemize}
    \item \textbf{SD 3.5 $\rightarrow$ DALL·E 3:} Prompts optimized for Stable Diffusion 3.5 transfer effectively to DALL·E 3, achieving approximately 85\% of the performance gain of DALL·E 3's self-optimized prompts.
    
    \item \textbf{Playground v2.5 $\rightarrow$ DALL·E 3:} Similarly, Playground-optimized prompts transfer well to DALL·E 3, suggesting these models share similar prompt understanding mechanisms.
    
    \item \textbf{Asymmetric Transfer:} Interestingly, transfer effectiveness is not always symmetric. Prompts optimized for DALL·E 3 transfer less effectively to other models, suggesting it may have developed more specialized prompt understanding capabilities.
\end{itemize}

These transfer patterns provide insights into the shared conceptual understanding across different model architectures. The transferability suggests that different diffusion models may learn similar representations of visual concepts and share underlying prompt preferences, enabling the remarkable transferability observed in our experiments.
\section{Ablation study on computational budget}
\label{sec:budget_experiment}
An important practical consideration for prompt optimization is the computational budget, particularly the number of image generations required. In our main experiments, we compared the performance of optimized prompts (which required 1 initial image + 5 optimization iterations + 5 final test generations) against baseline approaches that generated 5 images with the original prompt. This comparison, while demonstrating the effectiveness of our framework, does not account for the additional computational cost of the optimization process itself.
To address this concern, we conduct a controlled computational budget experiment where both the baseline and our method are limited to exactly 5 image generations in total. For the baseline, we maintain the same approach of generating 5 images with the original prompt and selecting the best one. For our prompt optimization approach, we modify the procedure as follows:
\begin{enumerate}
    \item We generate an initial image with the original prompt $p_0$
    \item We run 4 iterations of our optimization process, generating one image with each improved prompt $p_1, p_2, p_3, p_4$
    \item We select the best image from these 5 generations $(p_0, p_1, p_2, p_3, p_4)$ based on our evaluation metric
\end{enumerate}
This approach maintains strict budget parity between the methods, with both generating exactly 5 images. The key difference is that our approach generates images from a sequence of progressively optimized prompts, while the baseline generates multiple images from the same initial prompt.

\begin{figure*}[t]
    \centering
    \includegraphics[width=0.7\textwidth]{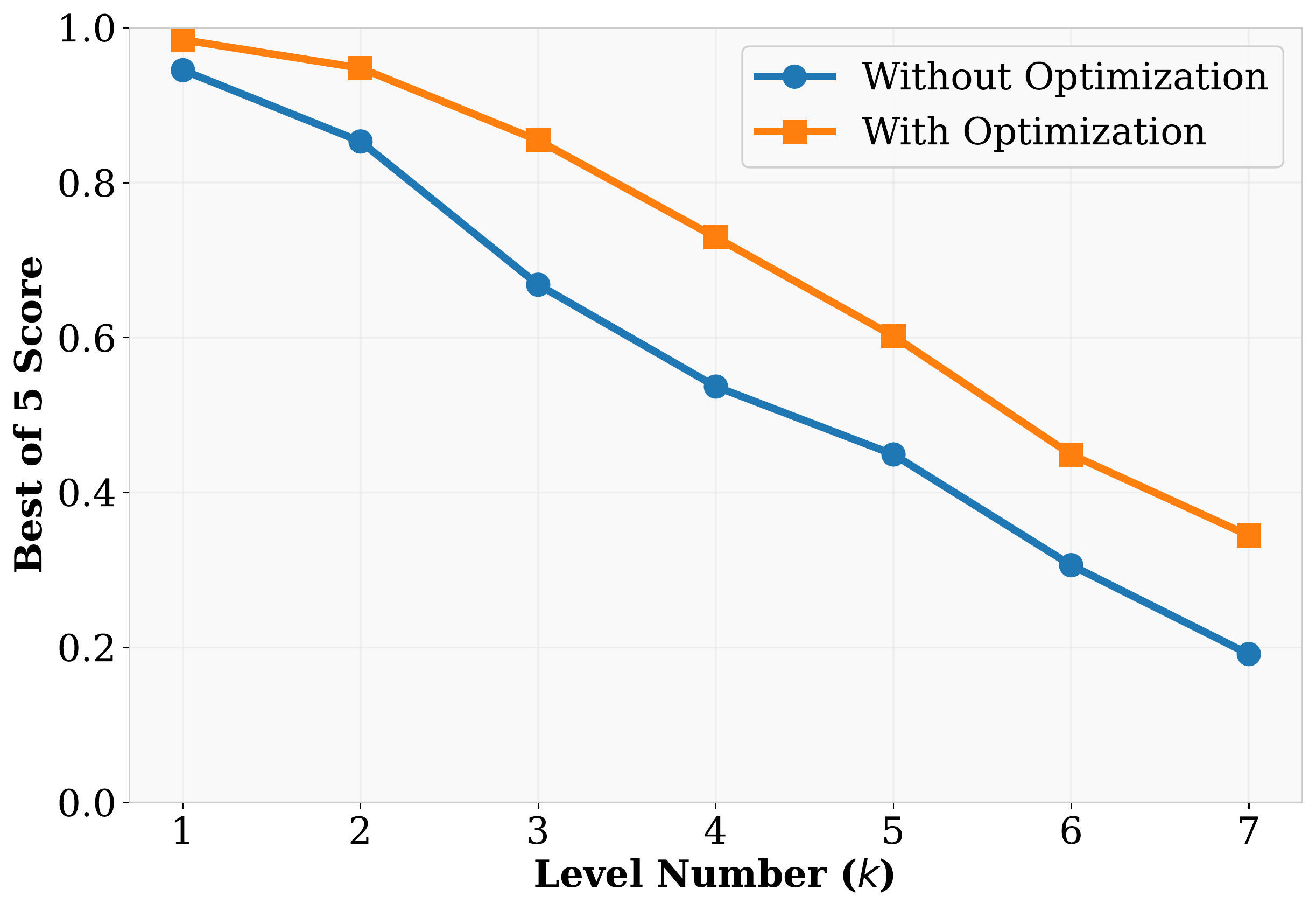}
    \caption{Comparison of DALL·E 3 performance under equal computational budget (5 images). The graph shows Best-of-5 Score for the baseline approach (5 images with the original prompt) versus our approach (5 images with progressively optimized prompts) across different complexity levels ($k=1$ to $k=7$).}
    \label{fig:budget_dalle3}
\end{figure*}

\begin{figure*}[t]
    \centering
    \includegraphics[width=0.7\textwidth]{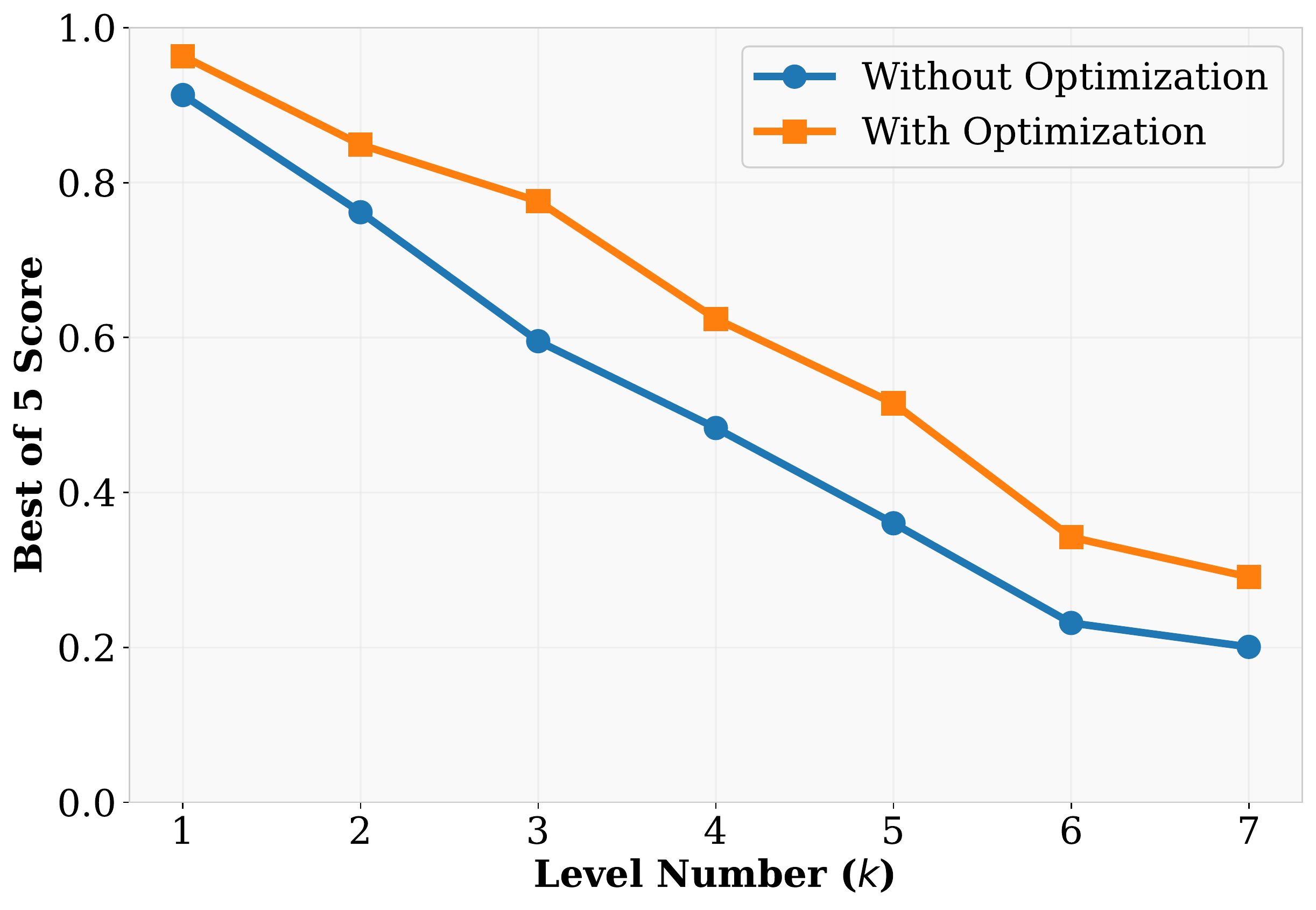}
    \caption{Comparison of Stable Diffusion 3.5 performance under equal computational budget (5 images). The graph shows Best-of-5 Score for the baseline approach (5 images with the original prompt) versus our approach (5 images with progressively optimized prompts) across different complexity levels ($k=1$ to $k=7$).}
    \label{fig:budget_sd}
\end{figure*}

\begin{figure*}[t]
    \centering
    \includegraphics[width=0.7\textwidth]{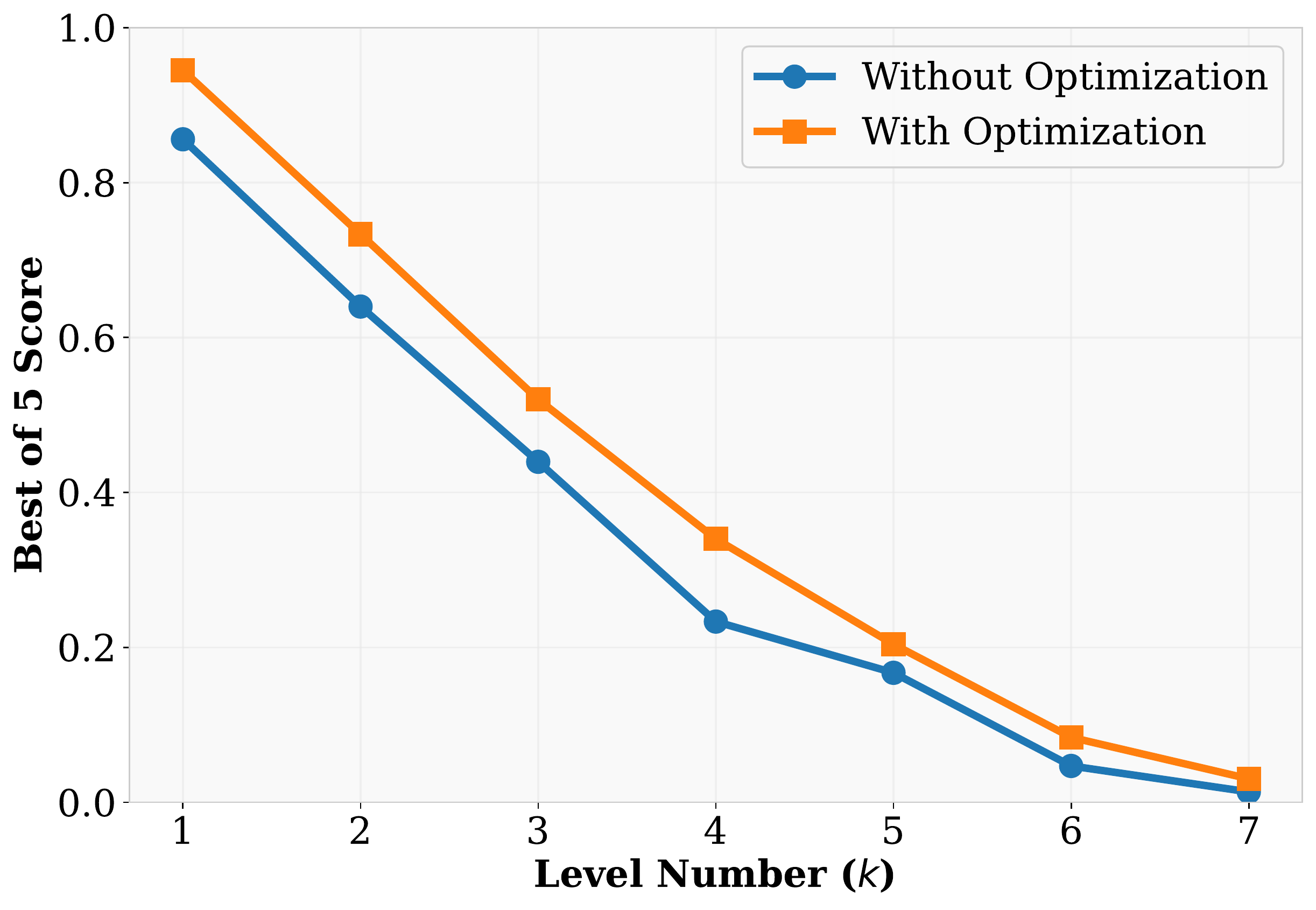}
    \caption{Comparison of Playground v2.5 performance under equal computational budget (5 images). The graph shows Best-of-5 Score for the baseline approach (5 images with the original prompt) versus our approach (5 images with progressively optimized prompts) across different complexity levels ($k=1$ to $k=7$).}
    \label{fig:budget_playground}
\end{figure*}

As shown in Figures \ref{fig:budget_dalle3}, \ref{fig:budget_sd}, and \ref{fig:budget_playground}, even with the same computational budget, our prompt optimization approach consistently outperforms the baseline across all complexity levels for all three models. This highlights another practical application of prompt optimization framework beyond benchmarking model capabilities: improving resource-constrained generation. When users have a fixed compute budget (e.g., limited to 5 DALL·E 3 API calls) and specific image criteria to meet, our framework can efficiently allocate these resources by progressively updating prompts based on previous generations, rather than repeatedly sampling from the same initial prompt.

\end{document}

%% file: preamble.tex
\usepackage{amsmath}
\usepackage{amssymb}
\usepackage{algorithm}
\usepackage{algorithmic}
\usepackage{tcolorbox}
\usepackage{tabularx}
\usepackage{balance}
\usepackage{colortbl}
\usepackage{xcolor}
\newcommand{\sd}{Stable Diffusion 3.5}
\newcommand{\dalle}{DALL·E 3}
\newcommand{\pg}{Playground v2.5}
\newcommand{\method}{ConceptMix++}

%% file: macros.tex
\newcommand{\1}{\mathbbm{1}}


%% file: Table/main_results_table.tex
\begin{table*}[t]
\caption{Performance comparison between original and optimized prompts across different complexity levels $k$. For each model, we report both Average Score and Best-of-5 Score to show the capability gap revealed through prompt optimization. Green intensity indicates the magnitude of improvement.}
\label{tab:main_results}
\centering
\resizebox{\textwidth}{!}{
\begin{tabular}{c|ccccccc}
\hline
\rowcolor{gray!20} \textbf{Model} & \textbf{$k=1$} & \textbf{$k=2$} & \textbf{$k=3$} & \textbf{$k=4$} & \textbf{$k=5$} & \textbf{$k=6$} & \textbf{$k=7$} \\
\toprule
DALL·E 3 (Avg Orig) & 0.824 ± 0.021 & 0.621 ± 0.015 & 0.460 ± 0.025 & 0.292 ± 0.025 & 0.207 ± 0.024 & 0.139 ± 0.028 & 0.085 ± 0.012 \\
DALL·E 3 (Avg Opt) & 0.882 ± 0.009 & 0.723 ± 0.023 & 0.605 ± 0.024 & 0.415 ± 0.028 & 0.313 ± 0.029 & 0.218 ± 0.009 & 0.127 ± 0.019 \\
\rowcolor{green!10} DALL·E 3 ($\Delta$ Avg) & \cellcolor{green!15}\textbf{+0.058} & \cellcolor{green!25}\textbf{+0.102} & \cellcolor{green!35}\textbf{+0.145} & \cellcolor{green!30}\textbf{+0.123} & \cellcolor{green!25}\textbf{+0.106} & \cellcolor{green!20}\textbf{+0.079} & \cellcolor{green!10}\textbf{+0.042} \\
\midrule
DALL·E 3 (Best Orig) & 0.945 & 0.853 & 0.668 & 0.537 & 0.449 & 0.306 & 0.191 \\
DALL·E 3 (Best Opt) & 0.996 & 0.943 & 0.868 & 0.761 & 0.648 & 0.505 & 0.364 \\
\rowcolor{green!15} DALL·E 3 ($\Delta$ Best) & \cellcolor{green!12}\textbf{+0.051} & \cellcolor{green!22}\textbf{+0.090} & \cellcolor{green!50}\textbf{+0.200} & \cellcolor{green!55}\textbf{+0.224} & \cellcolor{green!50}\textbf{+0.199} & \cellcolor{green!50}\textbf{+0.199} & \cellcolor{green!43}\textbf{+0.173} \\
\hline\hline
SD 3.5 (Avg Orig) & 0.736 ± 0.018 & 0.503 ± 0.024 & 0.351 ± 0.010 & 0.226 ± 0.020 & 0.157 ± 0.009 & 0.089 ± 0.009 & 0.067 ± 0.006 \\
SD 3.5 (Avg Opt) & 0.877 ± 0.006 & 0.662 ± 0.008 & 0.512 ± 0.020 & 0.356 ± 0.019 & 0.277 ± 0.017 & 0.172 ± 0.016 & 0.123 ± 0.016 \\
\rowcolor{green!10} SD 3.5 ($\Delta$ Avg) & \cellcolor{green!35}\textbf{+0.141} & \cellcolor{green!40}\textbf{+0.159} & \cellcolor{green!40}\textbf{+0.161} & \cellcolor{green!32}\textbf{+0.130} & \cellcolor{green!30}\textbf{+0.120} & \cellcolor{green!20}\textbf{+0.083} & \cellcolor{green!14}\textbf{+0.056} \\
\midrule
SD 3.5 (Best Orig) & 0.913 & 0.762 & 0.595 & 0.483 & 0.360 & 0.232 & 0.201 \\
SD 3.5 (Best Opt) & 0.973 & 0.856 & 0.776 & 0.678 & 0.579 & 0.423 & 0.298 \\
\rowcolor{green!15} SD 3.5 ($\Delta$ Best) & \cellcolor{green!15}\textbf{+0.060} & \cellcolor{green!23}\textbf{+0.094} & \cellcolor{green!45}\textbf{+0.181} & \cellcolor{green!48}\textbf{+0.195} & \cellcolor{green!54}\textbf{+0.219} & \cellcolor{green!47}\textbf{+0.191} & \cellcolor{green!24}\textbf{+0.097} \\
\hline\hline
PG v2.5 (Avg Orig) & 0.732 ± 0.017 & 0.451 ± 0.023 & 0.241 ± 0.020 & 0.117 ± 0.010 & 0.062 ± 0.009 & 0.017 ± 0.004 & 0.003 ± 0.002 \\
PG v2.5 (Avg Opt) & 0.831 ± 0.010 & 0.572 ± 0.006 & 0.351 ± 0.014 & 0.199 ± 0.009 & 0.116 ± 0.013 & 0.047 ± 0.009 & 0.019 ± 0.005 \\
\rowcolor{green!10} PG v2.5 ($\Delta$ Avg) & \cellcolor{green!25}\textbf{+0.099} & \cellcolor{green!30}\textbf{+0.121} & \cellcolor{green!27}\textbf{+0.110} & \cellcolor{green!20}\textbf{+0.082} & \cellcolor{green!13}\textbf{+0.054} & \cellcolor{green!7}\textbf{+0.030} & \cellcolor{green!4}\textbf{+0.016} \\
\midrule
PG v2.5 (Best Orig) & 0.856 & 0.640 & 0.440 & 0.233 & 0.167 & 0.047 & 0.014 \\
PG v2.5 (Best Opt) & 0.936 & 0.783 & 0.601 & 0.373 & 0.278 & 0.128 & 0.057 \\
\rowcolor{green!15} PG v2.5 ($\Delta$ Best) & \cellcolor{green!20}\textbf{+0.080} & \cellcolor{green!35}\textbf{+0.143} & \cellcolor{green!40}\textbf{+0.161} & \cellcolor{green!35}\textbf{+0.140} & \cellcolor{green!27}\textbf{+0.111} & \cellcolor{green!20}\textbf{+0.081} & \cellcolor{green!10}\textbf{+0.043} \\
\bottomrule
\end{tabular}
}
\end{table*}

%% file: sec/T2I_details.tex
\section{Additional details on prompt optimization module} \label{apx:prompt_opt}
\subsection{Algorithm}
Algorithm \ref{alg:t2i_grad} outlines the complete prompt optimization framework:

\begin{algorithm}
\caption{Text2Image Grad}
\label{alg:t2i_grad}
\begin{algorithmic}
\REQUIRE Initial prompt $p_0$, diffusion model $\mathcal{D}$, VLM $\mathcal{V}$, LLM $\mathcal{U}_{\text{LLM}}$, maximum iterations $T$
\ENSURE Optimized prompt $p^*$
\STATE Initialize history table $\mathcal{H} \leftarrow \emptyset$
\STATE Initialize best score $s_{\text{best}} \leftarrow -\infty$
\STATE Initialize best prompt $p_{\text{best}} \leftarrow p_0$
\FOR{$t = 0$ to $T$}
    \STATE Generate image $\mathcal{I}_t \leftarrow \mathcal{D}(p_t)$
    \STATE Evaluate image and generate feedback $(s_t, f_t) \leftarrow \mathcal{V}(\mathcal{I}_t)$
    \STATE Update history $\mathcal{H} \leftarrow \mathcal{H} \cup \{(p_t, s_t, f_t)\}$
    \IF{$s_t > s_{\text{best}}$}
        \STATE $s_{\text{best}} \leftarrow s_t$
        \STATE $p_{\text{best}} \leftarrow p_t$
    \ENDIF
    \IF{$t==T$}
        \STATE \textbf{break}
    \ENDIF
    \STATE Generate improved prompt $p_{t+1} \leftarrow \mathcal{U}_{\text{LLM}}(p_{\text{best}}, \mathcal{H})$
\ENDFOR
\RETURN $p_{\text{best}}$
\end{algorithmic}
\end{algorithm}

\subsection{Analogy with traditional Gradient Descent}
Traditional gradient descent optimization iteratively updates parameters using the gradient of the objective function according to the following update rule:
\begin{equation}
\theta_{t+1} = \theta_t - \eta \nabla f(\theta_t)
\end{equation}
where $\theta_t$ represents the parameters at iteration $t$, $\eta$ is the learning rate, and $\nabla f(\theta_t)$ is the gradient of the objective function $f$ with respect to $\theta_t$.

In the context of text-to-image generation, our objective is to maximize the score function $\mathcal{V}(\mathcal{D}(p))$, where $p$ is the prompt, $\mathcal{D}$ is the diffusion model, and $\mathcal{V}$ is the VLM evaluation. However, we cannot directly compute the gradient $\nabla_p \mathcal{V}(\mathcal{D}(p))$ for several reasons:

\begin{itemize}
    \item The prompt $p$ exists in a discrete, non-Euclidean space rather than a continuous parameter space.
    \item The diffusion process $\mathcal{D}$ involves complex, non-differentiable stochastic sampling procedures.
    \item The VLM evaluation $\mathcal{V}$ is similarly complex and not directly differentiable with respect to its inputs.
\end{itemize}

To address these challenges, our prompt optimization framework draws inspiration from numerical optimization methods to approximate the gradient and update direction. The key insight is that we can view our optimization process through the lens of zeroth-order optimization methods, particularly finite-difference approximations of gradients.

Consider how traditional finite-difference methods approximate gradients:
\begin{equation}
\nabla f(\theta) \approx \frac{f(\theta + \delta) - f(\theta)}{\delta}
\end{equation}

In our context, we generalize this approach to work with a collection of prompt-score pairs $\{(p_i, s_i)\}$ that represent different points in the prompt space. Each pair provides information about the evaluation function $\mathcal{V}(\mathcal{D}(p))$ at different locations.

Instead of just using point-wise differences, we leverage the entire history $\mathcal{H}$ of prompt-score-feedback tuples to approximate a more robust update direction. The LLM, acting as a sophisticated non-parametric estimator, analyzes this historical data to infer the direction in the prompt space that is most likely to increase the objective function.

Conceptually, the prompt update process at time step $t$ can be represented as:
\begin{equation}
p_{t+1} = p_{\text{best}} + \Delta p_{\text{best}}
\end{equation}
where\begin{equation}
\quad \Delta p_{\text{best}} \approx \mathcal{U}_{\text{LLM}}(p_{\text{best}}, \mathcal{H})
\end{equation}
and $p_{\text{best}}$ is the best prompt at time step $t$ based on scores.

Here, $\mathcal{U}_{\text{LLM}}$ serves as both the gradient approximator and the update direction determiner. By analyzing the relationship between previous prompts and their resulting scores, the LLM effectively constructs a local approximation of the prompt-score landscape and generates a new prompt that is likely to achieve a higher score.

A key distinction from standard gradient descent is our use of $p_{\text{best}}$ rather than $p_t$ as the base for updates. In the context of discrete prompt optimization, unlike continuous optimization where small steps can be taken with controlled learning rates, a single prompt update may result in significant performance degradation due to the stochasticity of diffusion models and the discrete nature of language. Therefore, we will need a more conservative strategy of updating the prompt. By always starting from the best-performing prompt, we establish a reliable foundation for exploration.

Moreover, even iterations that do not improve upon $p_{\text{best}}$ contribute valuable information to $\mathcal{H}$, enriching the LLM's understanding of the prompt-performance landscape. This accumulated knowledge enhances the accuracy of subsequent update steps, effectively implementing a form of trust-region optimization that balances exploitation and exploration.

This strategy mitigates the risk of divergence in the optimization process while still allowing for thorough exploration of the prompt space. Although this approach may theoretically limit escape from local optima, the high dimensionality of the prompt space and the relatively small number of iterations ($T=5$) make this a favorable trade-off, prioritizing stable improvement over potentially unstable exploration.
\subsection{Prompt update process}
\label{sec:prompt_update}

In our framework, we combine the gradient approximation and update steps into a single operation performed by an LLM. This design choice is motivated by the observation that separating these steps—first determining how to improve the prompt and then implementing those improvements—introduces unnecessary complexity and potential information loss.

By unifying these steps, we enable the LLM to reason holistically about the optimization process:

\begin{equation}
p_{t+1} = \mathcal{U}_{\text{LLM}}(p_{\text{best}}, \mathcal{H})
\end{equation}

The LLM receives the best-performing prompt so far and the complete history of previous prompt-score-feedback tuples. It then analyzes patterns in this data to identify what aspects of successful prompts contributed to their high scores and what aspects of unsuccessful prompts led to lower scores. Based on this analysis, it generates a new prompt that incorporates successful elements while addressing identified shortcomings.

This approach leverages the LLM's capabilities in several ways:
\begin{itemize}
    \item \textbf{Pattern recognition}: The LLM can identify subtle patterns in the relationship between prompt characteristics and resulting scores.
    \item \textbf{Contextual understanding}: The LLM can interpret feedback in the context of specific prompts and images.
    \item \textbf{Generative capability}: The LLM can produce entirely new prompt formulations rather than being limited to predefined update rules.
\end{itemize}

\subsection{Hyperparameters and example of optimizing ConceptMix}
\label{sec:hyperparams}

\subsubsection{Hyperparameters}
Following the TextGrad approach \cite{yuksekgonul2024textgrad}, we fix the number of iterations to 5 for all experiments, which provides a good balance between computational efficiency and optimization effectiveness. For our implementation of the framework, we use GPT-4o in two critical roles: first, as the VLM ($\mathcal{V}$) that evaluates generated images and provides feedback through concept-specific scoring; and second, as the LLM ($\mathcal{U}_{\text{LLM}}$) that generates improved prompts based on the accumulated history of previous iterations. This dual application of GPT-4o creates a unified optimization framework where both evaluation and improvement processes leverage the same multimodal understanding capabilities.
\begin{figure*}[h]
\centering
\includegraphics[width=\textwidth]{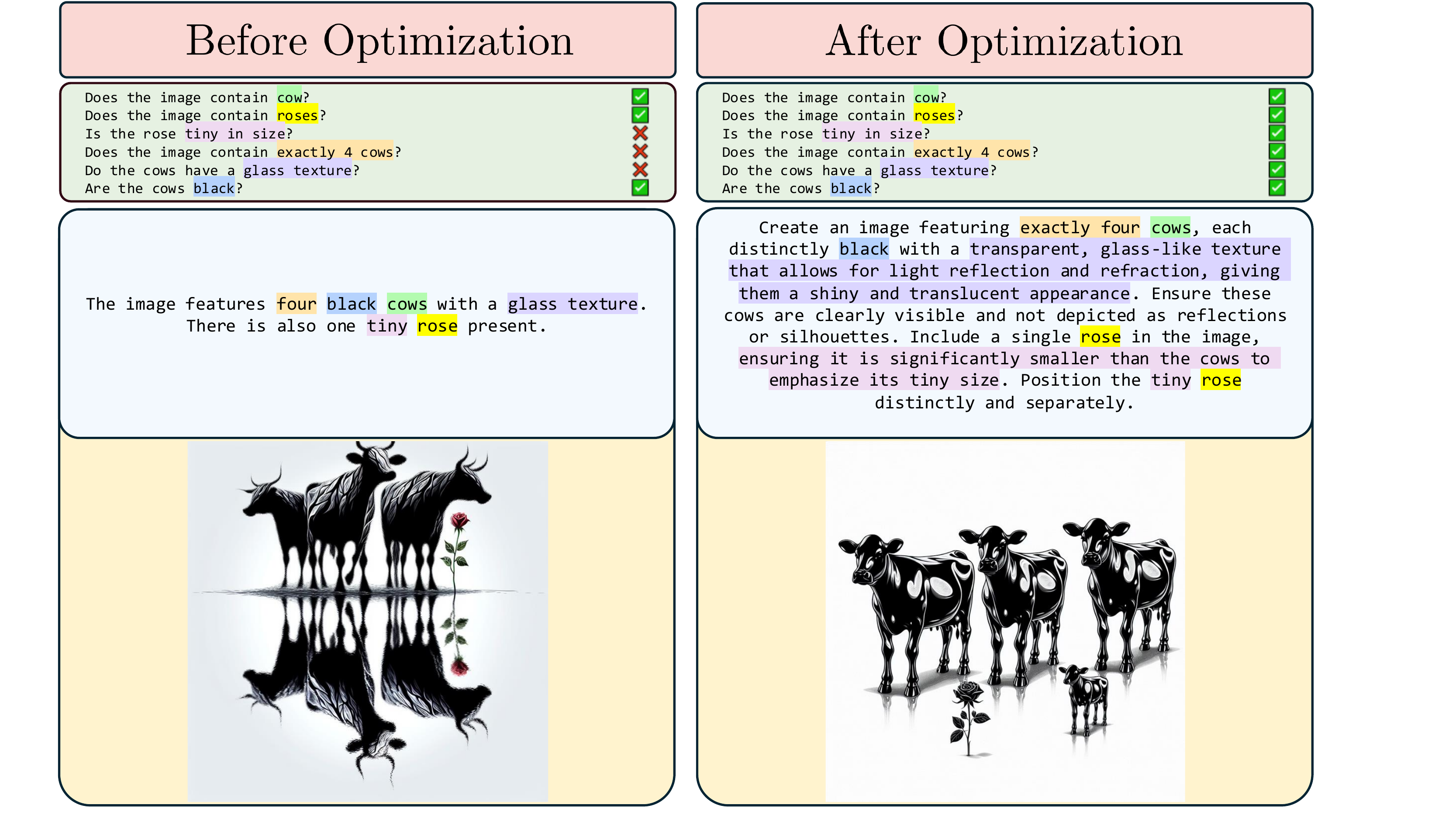}
\caption{Example of the our prompt optimization process. Left: Before optimization, the image shows 3 black cows with reflections and a rose that is not tiny. The corresponding prompt satisfies only 3 out of 6 criteria. Right: After optimization, the image shows exactly 4 black cows with glass texture and a tiny rose, satisfying all 6 criteria.}
\label{fig:example}
\end{figure*}
\subsubsection{Example of optimizing ConceptMix}
\label{sec:example_optimization}

Here we provide a detailed example of the our prompt optimization process applied to a prompt from the ConceptMix dataset. This example illustrates our implementation of Algorithm \ref{alg:t2i_grad} in practice. The visualization is shown in Figure \ref{fig:example}.
\paragraph{Criteria}
We begin with the criteria:
\begin{tcolorbox}[colback=gray!10, colframe=gray!40, title=Criteria $C$]
\begin{itemize}
    \item Does the image contain cow?
    \item Does the image contain roses?
    \item Is the rose tiny in size?
    \item Does the image contain exactly 4 cows?
    \item Do the cows have a glass texture?
    \item Are the cows black?
\end{itemize}
\end{tcolorbox}
\paragraph{Initial Prompt Evaluation}
We then pick the initial prompt $p_0$ from the ConceptMix dataset:
\begin{tcolorbox}[colback=gray!10, colframe=gray!40, title=Initial Prompt $p_0$]
The image features four black cows with a glass texture. There is also one tiny rose present.
\end{tcolorbox}
After generating an image $\mathcal{I}_0 = \mathcal{D}(p_0)$ using our diffusion model, for each criterion $c_i \in C$ we score the image using GPT-4o as our VLM $\mathcal{V}$ with the following prompt:
\begin{tcolorbox}[colback=gray!10, colframe=gray!40, title=VLM Score Prompt]
Does the image contain cow? Respond `Yes' or `No'.
[IMAGE]
\end{tcolorbox}
For each criterion $c_i$ at timestep $t$, we get a score $s_{t,i} = P(\mathcal{V}(c_i+\text{``Respond `Yes' or `No'."}|\mathcal{I}) = \texttt{"Yes"})$ using the probability distribution of the first new token, and feedback $f_{t,i}$. The overall score $s_t$ for the current timestep is calculated as the product of all individual criterion scores:
\begin{equation}
s_t = \prod_{i=1}^{k+1} s_{t,i}
\end{equation}
where $k$ is the number of criteria $-1$ (in this case, $k=5$).
Similarly, for each criterion $c_i \in C$, we get the feedback with the following prompt:
\begin{tcolorbox}[colback=gray!10, colframe=gray!40, title=VLM Feedback Prompt]
Does the image contain cow? If the answer is "No", please explain in one sentence the specific issue that prevents the image from satisfying the question; otherwise, just output that the image satisfies the question.
[IMAGE]
\end{tcolorbox}
The overall feedback $f_t$ is the string produced by concatenating each criterion $c_i$ and its corresponding feedback $f_{t,i}$ for all criteria. Here's an example of combined feedback for the initial prompt ($t=0$):

\begin{tcolorbox}[colback=gray!10, colframe=gray!40, title=VLM Feedback Results]
Does the image contain cow? The image satisfies the question.\\
Does the image contain roses? The image satisfies the question.\\
Is the rose tiny in size? The rose is nearly as tall as the cows, which is not tiny in size.\\
Does the image contain exactly 4 cows? The image does not contain exactly 4 cows because it only shows 3 cows and their reflections.\\
Do the cows have a glass texture? No, the cows do not have a glass texture as they appear as solid, opaque silhouettes without transparency or reflective qualities.\\
Are the cows black? The image satisfies the question.
\end{tcolorbox}

We then add the tuple $(p_0, s_0, f_0)$ to the history table $\mathcal{H}$ and prompt the LLM to generate an improved prompt:

\begin{tcolorbox}[colback=gray!10, colframe=gray!40, title=LLM Optimization Prompt]
I'm trying to generate an image that matches specific requirements. Please create a concise description that will help the model generate an image that satisfies all the requirements and get a high product of scores. Here's the context:

1. Requirements: The image must satisfy these criteria (all should receive 'Yes' answers):
\{criteria\}

2. History: Previous attempts sorted by performance (best to worst):
\{formatted history table\}

Based on the requirements and previous attempts, please provide a new, improved description for the image generation model. The description should:
\begin{itemize}
    \item Be specific to guide the image generation
    \item Address all the required elements from the questions
    \item Learn from previous attempts, especially what worked in higher-scoring versions
\end{itemize}

Now please give a concise description to help the model generate an image that meets all the requirements and gets the highest product of scores.
\end{tcolorbox}

\paragraph{Iterative Optimization}
This prompt optimization process continues for a total of $T=5$ iterations, with each new prompt $p_t$ being evaluated to produce a score $s_t$ and feedback $f_t$. The history table $\mathcal{H}$ is updated at each iteration, and the LLM uses this accumulated information to generate increasingly refined prompts.

\paragraph{Final Result}
After the iterative optimization process, we obtain our final optimized prompt $p^*$:

\begin{tcolorbox}[colback=gray!5, colframe=gray!40, title=Final Optimized Prompt $p^*$]
Create an image featuring exactly four cows, each distinctly black with a transparent, glass-like texture that allows for light reflection and refraction, giving them a shiny and translucent appearance. Ensure these cows are clearly visible and not depicted as reflections or silhouettes. Include a single rose in the image, ensuring it is significantly smaller than the cows to emphasize its tiny size. Position the tiny rose distinctly and separately from the cows.
\end{tcolorbox}

This example demonstrates how our prompt optimization framework systematically improves the initial prompt through targeted feedback and iterative optimization, resulting in increasingly accurate representations of the desired concepts. At each stage, the algorithm leverages both visual evaluation and linguistic refinement capabilities of the LLM to navigate the complex prompt space effectively.